\documentclass[10pt,journal,compsoc]{article}
\usepackage[a4paper, total={7in, 9in}]{geometry}
\usepackage{tikz}
\newtheorem{lemma}{Lemma}
\newtheorem{theorem}{Theorem}
\usepackage{algpseudocode}
\usepackage{float}
\usepackage{amssymb}
\usepackage{amsmath}

\begin{document}
%
%
%
%
%

\author{Carlos~Puerto-Santana, \\
{\footnotesize Universidad Politécnica de Madrid, Madrid, Spain} \protect\\
{\footnotesize E-mail: ce.puerto@alumnos.upm.es} \\
{\footnotesize Aingura IIoT, San Sebastian, Spain} \protect\\
{\footnotesize E-mail: epuerto@ainguraiiot.com} \\
        Pedro~Larra\~naga \\
{\footnotesize Universidad Politécnica de Madrid, Madrid, Spain}\protect\\
{\footnotesize E-mail: pedro.larranaga@fi.upm.es} \\
        Concha~Bielza \\
{\footnotesize Universidad Politécnica de Madrid, Madrid, Spain}\protect\\
{\footnotesize         E-mail: mcbielza@fi.upm.es}
}

%
%

\markboth{}%
{Shell \MakeLowercase{\textit{et al.}}: Bare Advanced Demo of IEEEtran.cls for IEEE Computer Society Journals}
\title{Autoregressive Asymmetric Linear Gaussian Hidden Markov Models}
\maketitle
\begin{abstract}
In a real life process evolving over time, the relationship between its relevant variables may change. Therefore, it is advantageous to have different inference models for each state of the process. Asymmetric hidden Markov models fulfil this dynamical requirement and provide a framework where the trend of the process can be expressed as a latent variable. In this paper, we modify these recent asymmetric hidden Markov models to have an asymmetric autoregressive component, allowing the model to choose the order of autoregression that maximizes its penalized likelihood for a given training set. Additionally, we show how inference, hidden states decoding  and parameter learning must be adapted to fit the proposed model. Finally, we run experiments with synthetic and real data to show the capabilities of this new model.
\end{abstract}





%
\section{Introduction}\label{sec:introduction}


%
%
%
%
Hidden Markov models (HMMs) have been successfully used to analyze dynamic signals, e.g., in speech recognition \cite{Lra90} and tool wearing monitoring \cite{He91} or sequential signals, e.g., in gene prediction \cite{Sta06}. These models assume the existence of a latent or hidden variable that drives an observable set of variables. However, traditional HMMs in the case of continuous data, make the hypothesis that for all the driving dynamical process, a complete dependence probabilistic model involving all the variables is held, which can be untrue. This causes that the models learn a considerable number of unnecessary parameters that may cause data overfitting.

The idea of asymmetric HMMHs is introduced in \cite{Ki04} and \cite{BuM17}. These models imply that depending on the value of certain variables, the distribution of the remaining variables may change. For HMMs, the asymmetric component is expressed with the hidden variable, with which depending on its value, a context-specific Bayesian network \cite{Bo96} encodes the distribution of the emission probabilities. These context-specific Bayesian networks reduce the number of parameters needed.

Autoregressive (AR) processes have been studied for a long time, especially for regression tasks \cite{Bo76}. However, the traditional approaches to AR processes make strong assumptions as to stationariness that do not hold for many real case scenarios. This issue was addressed by \cite{Ha90}, allowing the models to have changing parameters depending on the value of a hidden variable. Nevertheless the order of the AR process  had to be fixed beforehand by trial and error. HMMs and AR processes were combined in \cite{Po82}, where  AR coefficients were added to the emission probabilities. 

In this paper we combine the ideas of asymmetric HMMs with AR processes to overcome the previous shortcomings: determine the AR order of a model for each hidden state and reduce the number of unnecessary parameters. Specifically, our model enables each variable, depending on the hidden state, to determine its parents within the context-specific Bayesian network and the number of lags that its distribution requires to maximize a model fitting score.

The structure of this document is as follows. Section 2 describes related work about asymmetric probabilistic models and  HMMs with AR processes. Section 3 reviews HMMs in general and summarizes the  expectation maximization algorithm (EM), the structural EM and the  Yule-Walker equations  \cite{Bo76} that are relevant tools for our model. Section 4 introduces the proposed autoregressive asymmetric linear Gaussian hidden Markov model (AR-AsLG-HMM). In this section we discuss the adaptation of the forward-backward and Viterbi algorithms \cite{Lra90}. We also describe the parameter and structural learning and show that the EM algorithm iteratively  improves  the log-likelihood of the data for our model. Section 5 presents experiments with synthetic data, real air quality and ball-bearing degradation data. The results obtained using the AR-AsLG-HMM are compared against its non-AR version and other state-of-the-art approaches. The paper is rounded off in Section 6 with conclusions and comments regarding possible future research.

\section{Related work}

In this section we review the related work regarding HMMs with AR behavior and asymmetric probabilistic models. Table~\ref{table:sota} shows the reviewed articles grouped according to their contribution.

\begin{table}
	\centering
	\begin{tabular}{l}
		\hline
		\textbf{Modified emission probabilities in HMMs}: \\
		AR polynomials in emission probabilities \cite{Po82} \\
		AR Mixture of Gaussians HMM (AR-MoG-HMM) \cite{Ju85} \\
		Markov mean-switching AR model (MMSAR) \cite{Ha89} \\
		Vector AR multivariate Gaussian HMM (VAR-MVGHMM) \cite{Ke90} \\
		Linear Markov switching AR model (LMSAR) \cite{Ha90} \\
		Gaussian AR-HMMs with a linear error coefficient \cite{Br15} \\
		AR hidden semi-Markov model (AR-HSMM) \cite{Na15} \\
		Transitional Markov switching autoregressive model (TMSAR) \cite{Chg16} \\
		Vector AR hierarchical HSMM (VAR-HHSMM) \cite{Ma18} \\
		\hline
		\textbf{Modified hidden variables}:\\
		AR-HMM with an additional memoryless hidden variable \cite{As12} \\
		Higher-order AR-HMM (AR-HO-HMM) \cite{Se14} \\
		\hline 
		\textbf{Missing data in HMMs}: \\
		AR-HMM with a missing at random assumption \cite{Sta14} \\
		AR-HMM with missing data as latent variables \cite{Da16} \\
		\hline
		\textbf{Asymmetric models}: \\
		Similarity networks  \cite{He90} \\
		Bayesian multinets \cite{Ge96} \\
		Context-specific Bayesian networks \cite{Bo96}\\
		Buried Markov model (BMM) \cite{Bi03} \\
		Conditional Chow-Liu trees  with HMMs \cite{Ki04} \\
		Chain events graph (CEG) \cite{Sm06} \\
		Stratified graphical model (SGM) \cite{Ny14} \\
		Dynamic chain events graph\cite{Ba15}\\
		Asymmetric HMM with discrete variables (As-HMM)\cite{BuM17} \\
		Asymetric HMM with continuous variables (AsLG-HMM) \cite{Pu18} \\
		\hline
	\end{tabular}
	\caption{Reviewed articles and their contributions to asymmetric HMMs and AR HMMs}
	\label{table:sota}
\end{table}

 \subsection{Modified emission probabilities in HMMs}
One of the first combinations of HMM and AR models attempted to process speech data \cite{Po82}.  Autoregressive polynomials were added to the Gaussian emission probabilities, in which coefficients were determined via the Baum-Welch algorithm \cite{Lra90}.  Later,  \cite{Ju85} proposed mixtures of Gaussian hidden Markov models (AR-MoG-HMMs) where the emission probabilities were modelled as mixtures of Gaussians. These models were used for speech recognition. In \cite{Ke90} a vectorial AR multivariate Gaussian HMM (VAR-MVGHMM) was introduced. This model enables variables to have temporal dependencies with all the other variables. Again, the model was used for speech recognition. 

Some authors \cite{Br15} modified the emission probabilities such that they behave as an AR Gaussian but with an error coefficient given by the linear prediction residuals \cite{It75}. 

Others also considered variations of HMMs such as hidden semi-Markov models (HSMMs), where the time duration of each hidden state can be modified to not always follow a geometric distribution, or hierarchical hidden Markov models (HHMMs) where AR behavior was considered. For instance, \cite{Na15} proposed an AR-HSMM, where AR variables and non-AR variables could be considered in the same model depending on the modeller's decision. \cite{Ma18} proposed a vector AR hierarchical hidden semi-Markov model  (VAR-HHSMM) to classify and determine hand movements. 

Other approximations of HMMs with AR properties can be found in \cite{Ha89}  and \cite{Ha90}. The author proposed an edited log-likelihood function to represent the AR behavior in data. Markov mean-switching AR models (MMSAR) and linear Markov-switching AR model (LMSAR) were studied and their parameters were calculated with the EM algorithm. \cite{Chg16} proposed the transitional Markov switching autoregressive (TMSAR) model as an extension of MMSAR and LMSAR models. In this case, the emission probabilities depend on past values of the hidden process to determine changes in its mean and its weight. The authors used maximum likelihood methods with a Newton-Raphson strategy to estimate the model parameters.

\subsection{Modified hidden variables}
 
 In more recent works, new approaches have been proposed in which the assumptions about the hidden variables that govern the process were modified such as the model given by \cite{As12}, where the authors edited an autoregressive hidden Markov model (AR-HMM) by introducing a memoryless hidden variable. The Markovian hidden states had a probabilistic dependency of this memoryless hidden variable. AR higher-order HMMs (AR-HO-HMMs) were introduced in \cite{Se14}. The authors not only considered an autoregressive property in the observations, but also a fixed order Markov assumption in the hidden states specified by the user. They used mixtures of Gaussians with AR properties for the emission probabilities.
 
 \subsection{Missing data in HMMs}
 
 Other works focused on the missing data. In \cite{Sta14} an AR-HMM with a missing at random assumption was proposed to perform exact inference in such scenarios. In \cite{Da16} the missing data was considered as latent variables. Additionally, the authors proposed a modified forward-backward algorithm and Baum-Welch parameter updating formulas.

 \subsection{Asymmetric models}
 
 As regards asymmetric probabilistic graphical models, the Bayesian multinets introduced in \cite{Ge96} were used to describe different local graphical models depending on the values of certain observed variables; the similarity networks in \cite{He90} allowed the creation of independent influence diagrams\footnote{An influence diagram is a probabilistic graphical model used for decision problems, where random, decision and value nodes are present \cite{Sha88}} for subsets of a given domain. Context-specific independence in Bayesian networks in \cite{Bo96} used tree structured conditional probability distributions with a D-separation-based algorithm to determine statistical dependencies between variables according to contexts given by instantiations of subsets of variables. Following these ideas, more recently in \cite{Ny14}, stratified graphical models (SGM) were proposed, where the concept of stratum was introduced to allow different factorizations for a probability distribution depending on the values of some of the variables. A nonreversible Metropolis-Hastings algorithm to calculate marginal likelihoods and learn decomposable SGMs was given. \cite{Sm06} introduced the chain events graphs (CEG). A CEG consists of a directed colored graph obtained from a staged tree \footnote{A staged tree is a probabilistic graphical model, where the graph is a tree and the nodes are random variables whose non leaf variables are identified with the same color if they have the same  conditional probabilistic relationships  with their children nodes \cite{Sm06}} by successive edge contraction operations. The obtained graphical model can represent conditional independence and causal behavior that traditional Bayesian networks cannot show. Later, a dynamic version was proposed \cite{Ba15}.
 
 Other authors have attempted to combine asymmetric models with HMMs. For example,   in \cite{Bi03} the buried Markov models (BMM) were introduced. In that article, the models of \cite{Ke90} were used, but the temporary dependencies can vary depending on the hidden state. These context-specific dependencies are learned using mutual information strategies. \cite{Ki04} used Chow-Liu trees and conditional Chow-Liu trees coupled with HMMs. The HMMs were used to model the dynamic behavior of a process, and the Chow-Liu tree was used to model the emission probabilities. A Chow-Liu tree or conditional Chow-Liu tree was associated with each value of the hidden variable. The parameters of the model were computed with the EM algorithm; specifically, the tree structure was determined in the maximization step. However, the model was specified only for discrete variables. More recently, asymmetric hidden Markov models (As-HMMs) were proposed in \cite{BuM17}, where a local graphical model was associated with each value of the hidden variable, and the graphical model was not restricted to Chow-Liu trees. However, again only models with discrete observable variables were allowed. In \cite{Pu18}, this issue was addressed with the asymmetric linear Gaussian HMMs (AsLG-HMMs), where the emission probabilities were modeled as conditional linear Gaussian Bayesian networks. The estimation of the model parameters was performed with the EM algorithm. 
 
 In this paper, we extend asymmetric HMMs for continuous variables of \cite{Pu18}, where the model during its learning phase can estimate for each variable the order of the AR process as well as its parameters depending on the context or value of the hidden variable. Thus, we couple for the first time asymmetric linear Gaussian HMMs with AR processes. 
 
\section{Theoretical Framework}

Because the proposed model needs to fit the forward-backward and Viterbi algorithms, we first review these algorithms  and the traditional HMM. The parameter and structure learning of the proposed model will be performed via the EM and SEM algorithms; therefore, we also review these algorithms and their properties. Additionally, because the Yule-Walker equations will be used to determine the order of an AR process, they are briefly examined. Additionally, in Table~\ref{table:symbols3}, a description of relevant symbols used in this article is shown.

\begin{table}
	\centering
	\begin{tabular}{ll}
		Symbol & Meaning \\
		\hline 
		$N$ & Number of hidden states\\
		$M$ & Number of variables \\
		$\boldsymbol{Q}^{0:T}$  & Sequence of hidden states from time $0$ up time $T$ \\
		$\boldsymbol{X}^{0:T}$  & Sequence of observations from time $0$ up time $T$\\
		$ R(\cdot)$ & Range of a random variable\\
		$\textbf{A}$  & Transition matrix \\
		$a_{ij}$ & Transition probability of hidden state $i$ to $j$ \\
		$\boldsymbol{\pi}$ & Initial probability distribution \\
		$\pi_i$ & Probability of starting at hidden state $i$ \\
		$\textbf{B}$  & Emission probabilities \\
		$b_i^{p^*}(\boldsymbol{x}^t)$ & Emission probability for the proposed model  \\
		$\Omega$ & Space of model parameters \\
		$\sigma^2_{im}$ & Variance of the Gaussian of variable $X_m$ at state $i$ \\ 
		$\boldsymbol{\lambda}$  & Model parameters \\
		$\boldsymbol{\lambda}'$ & Prior parameters \\
		$\alpha^t(i)$ & Forward variable at time $t$ for the hidden state $i$\\
		$\alpha_{p^*}^t(i)$ & Forward variable for the proposed model\\
		$\beta^t(i)$  &  Backward variable at time $t$ for hidden state $i$  \\
		$\beta_{p^*}^t(i)$  &  Backward variable for the proposed model  \\
		$\delta^t(i)$ & Most probable sequence of hidden states up to time $t-1$ \\
		$\delta_{p^*}^t(i)$ & Most probable sequence of hidden states for the proposed model\\
		$\psi^{t}(i)$ & Most probable transition from hidden state $i$ at time $t$\\
		$\gamma^t(i)$ & Probability of hidden state $i$ at time $t$ \\
		$\xi^t(i,j)$  & Transition probability from hidden state $i$ to $j$ at time $t$\\
		$\mathcal{Q}$ & Auxiliary optimization function \\
		$\mathcal{Q}^{p^*}$ & Auxiliary optimization function for the proposed model \\
		$\mathcal{B}$ & Probabilistic graphical model \\
		$\mathcal{B}'$& Prior probabilistic graphical model \\
		$\#(\cdot)$ & Number of parameters of the input graphical model \\
		$\phi_{kj}$   & Weight of the $j$ lag when $k$ lags are used \\
		$\rho_k$      & Correlation between $X^t$ and $X^{t-k}$ \\
		$\Phi_k$      & $\rho_k$ but removing intermediary lags effect\\ 
		$E[\cdot]$  & Expectation operator \\
		$p^*$ & Maximum admissible lag \\
		$\textbf{Pa}_i(X_m)$ & The set of parents of $X_m$ for hidden state $i$ \\
		$\textbf{pa}^t_{im}$ & Value at $t$ of $\textbf{Pa}_i(X_m)$ for hidden state $i$ \\
		$U_{imk}$ & $k^{th}$ parent of variable $X_m$ at hidden state $i$\\
		$k_{im}$ & Number of fathers for variable $X_m$ at the hidden state $i$\\
		$\boldsymbol{\beta}_{im}$ & Weights of $\textbf{pa}^t_{im}$  for the linear dependency of $X_m$ \\
		$\textbf{d}^t_{im}$ & Value at $t$ of the AR variables of $X_m$ for hidden state $i$ \\  
		$\boldsymbol{\eta}_{im}$ & Weights of $\textbf{d}^t_{im}$  for the linear dependency of $X_m$ \\
		$f^t_{im}$ & $\boldsymbol{\beta}_{im}\cdot\textbf{pa}^t_{im} + \boldsymbol{\eta}_{im}\cdot\boldsymbol{d}_{im}^t$ \\
		$g(i)$ & Numeric label to hidden state $i$ \\
		$\boldsymbol{\kappa}$ & Standard values for labelling \\
		$\textbf{v}$& Scaling vector for labelling\\
	\end{tabular}
	\caption{Symbols used in Section 3 and Section 4}
	\label{table:symbols3}
\end{table}

\subsection{Hidden Markov Models }\label{sec:HMM}

An HMM can be seen as a double chain stochastic model, where a chain is observed, namely $\boldsymbol{X}^{0:T}= (\boldsymbol{X}^0,...,\boldsymbol{X}^T)$, where $\boldsymbol{X}^t = (X^t_1,...,X^t_M) \in \mathbb{R}^M$ and the other chain is hidden, namely $\boldsymbol{Q}^{0:T} = (Q^0,...,Q^T)$. Here, $T$ is the length of the data. The usual approach for HMMs \cite{Lra90} is to assume that the hidden process has the first-order Markovian property, that is, $P(Q^t|\boldsymbol{Q}^{0:t-1}) = P(Q^t|Q^{t-1})$. Furthermore, it is assumed that the observable process depends on the hidden process, more specifically $P(\boldsymbol{X}^t|\boldsymbol{X}^{0:t-1},\boldsymbol{Q}^{0:t})=P(\boldsymbol{X}^t|Q^t)$. Additionally it is assumed that the range $R$ of the hidden variable is finite, i.e., $R(Q^t) = \{1,2,...,N\}$ for $t =0,1,...,T$. Moreover $\boldsymbol{R}(\boldsymbol{Q}^{0:T}) = \{1,2,...,N\}^{T+1}$.

All the previous HMM specifications can be summarized with the parameter $\boldsymbol{\lambda}=(\textbf{A},\textbf{B},\boldsymbol{\pi})\in \Omega$, where  $\Omega$ denotes the space of all possible parameters, $\textbf{A}= [a_{ij}]_{i,j=1}^N$ is a matrix representing the transition probabilities between  hidden states $i ,j\in R(Q^t)$ over time, i.e., $a_{ij}=P(Q^{t+1}=j|Q^t=i, \boldsymbol{\lambda})$; $\textbf{B}$ is a vector representing the emission probability of the observations given the hidden state, $\textbf{B} = [b_i(\mathbf{x}^t)]_{i=1}^N$, where $b_i(\boldsymbol{x}^t)= P(\boldsymbol{X}^t=\boldsymbol{x}^t|Q^t=i, \boldsymbol{\lambda})$ is a probability density function; $\boldsymbol{\pi}$ is the initial probability distribution of the hidden states, $\boldsymbol{\pi}=[\pi_j]_{j=1}^N$, where $\pi_j = P(Q^0= j|\boldsymbol{\lambda})$.

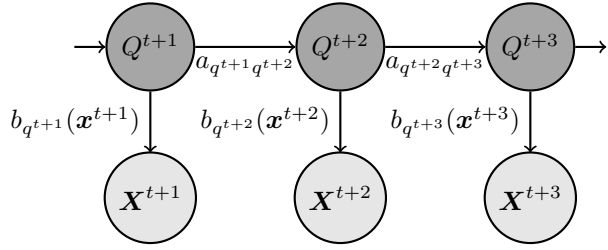
\begin{figure}[h]
	\centering
		\begin{tikzpicture} 
		[->,thick, scale=1,auto=center]
		
		\node (q1)[style={circle,fill=black!35,draw=black!100}] at (0,2)  {$Q^{t+1}$};
		\node (q2)[style={circle,fill=black!35,draw=black!100}] at (2.5,2)  {$Q^{t+2}$};
		\node (q3)[style={circle,fill=black!35,draw=black!100}] at (5,2)  {$Q^{t+3}$};
		
		\node (x1)[style={circle,fill=gray!20,draw=black!100}] at (0,0)  {$\boldsymbol{X}^{t+1}$};
		\node (x2)[style={circle,fill=gray!20,draw=black!100}] at (2.5,0)  {$\boldsymbol{X}^{t+2}$};
		\node (x3)[style={circle,fill=gray!20,draw=black!100}] at (5,0)  {$\boldsymbol{X}^{t+3}$};
		
		\draw (q1) -> node[below] {$a_{q^{t+1}q^{t+2}}$}(q2);
		\draw (q2) -> node[below] {$a_{q^{t+2}q^{t+3}}$}(q3);
		\draw (-1,2) -> (q1);
		\draw (q3) -> (6,2);
		
		\draw (q1) -> node[left] {$b_{q^{t+1}}(\boldsymbol{x}^{t+1})$}(0,0.6);
		\draw (q2) -> node[left] {$b_{q^{t+2}}(\boldsymbol{x}^{t+2})$}(2.5,0.6);
		\draw (q3) -> node[left] {$b_{q^{t+3}}(\boldsymbol{x}^{t+3})$}(5,0.6);

		\end{tikzpicture}
		\caption{An HMM as a probabilistic graphical model}
		\label{fig:hmmgraph}
\end{figure}

Additionally, an HMM can be seen as a probabilistic graphical model \cite{Mu02} (Fig.~\ref{fig:hmmgraph}), where the nodes of the graph represent random variables and the arcs represent direct probabilistic dependencies. 

Three main tasks can be performed in the context of HMMs. First, compute the likelihood of an observation $\boldsymbol{x}^{0:T}$ given a model $\boldsymbol{\lambda}$, i.e., $P(\textbf{x}^{0:T}|\boldsymbol{\lambda})$, which can be performed using the forward-backward algorithm. Second, compute the most likely sequence of hidden states for a sequence of observations, i.e., find the value of $\delta^t(i)=\max_{\boldsymbol{q}^{0:t-1}}\{P(\boldsymbol{x}^{0:t},\boldsymbol{q}^{0:t-1},Q^t=i|\boldsymbol{\lambda})\}$, $t=0,...,T$, $i=1,...,N$, which can be solved using the Viterbi algorithm. Third, learn the parameter $\boldsymbol{\lambda}$, which is estimated with the EM algorithm. A theoretical tutorial for understanding these algorithms can be found in \cite{Lra90}. We briefly review them below.
\subsection{The forward-backward algorithm}To execute the forward-backward algorithm, we first must define the forward and backward variables: $\alpha^t(i) = P(Q^t=i,\boldsymbol{x}^{0:t}|\boldsymbol{\lambda})$, $\beta^t(i) = P(\boldsymbol{x}^{t+1:T}|Q^{t}=i,\boldsymbol{\lambda})$, respectively,  $i =1,...,N$, $ t=0,...,T$. The forward and backward variables can be written recursively:

\begin{equation*}\label{eq:forward}
\begin{aligned}
\alpha^{t+1}(i)&= \sum_{j=1}^{N}b_i(\boldsymbol{x}^{t+1})a_{ji}\alpha^t(j)\\
\beta^{t}(i) &= \sum_{j=1}^Nb_j(\boldsymbol{x}^{t+1})a_{ij}\beta^{t+1}(j)
\end{aligned}
\end{equation*}
Their initial values are $\alpha^0(i) = \pi_ib_i(\boldsymbol{x}^0)$  and $\beta^T(i)=  1$. The forward variable can help us compute the likelihood of $\boldsymbol{x}^{0:T}$ since: 
\begin{equation*}\label{eq:likeli}
P(\boldsymbol{x}^{0:T}|\boldsymbol{\lambda})= \sum_{i=1}^NP(\boldsymbol{x}^{0:T},Q^T=i|\boldsymbol{\lambda}) =\sum_{i=1}^N\alpha^T(i).
\end{equation*}

\subsection{The Viterbi algorithm}
Variable $\delta^t(i)$ for time $t=0,...,T$ and hidden state $i=1,...,N$ can be written as:
\begin{equation*}\label{eq:viterbi}
\delta^t(i) =\max_{j=1,...,N}\{a_{ji}\delta^{t-1}(j)\}b_i(\boldsymbol{x}^{t}).
\end{equation*}
Its initial value is $\delta^0(i) = \pi_ib_i(\boldsymbol{x}^0)$. However, to find the most likely sequence of states $\boldsymbol{q}^{0:T}$, it is necessary to iteratively calculate an auxiliary variable $\psi^t(i) = \arg\max_{j=1,...,N}\{\delta^{t-1}(j)a_{ji}\}$, $i=1,...,N$, $t=0,...,T$,  which records the most likely transitions between states. Then, a backtracking process must be performed to recover $\boldsymbol{q}^{0:T}$, taking $q^T= \arg\max_{i=1,...,N}\{\delta^T(i)\}$ and $q^t = \psi^{t+1}(q^{t+1})$ for $t=T-1,...,0$.

\subsection{The EM algorithm}
To learn the parameter $\boldsymbol{\lambda} = (\bf{A},\bf{B},\boldsymbol{\pi})$ given a dataset $\boldsymbol{x}^{0:T}$ and a priori $\boldsymbol{\lambda}'$, the traditional EM approach \cite{DeA77} is used. In the EM algorithm, two steps called the expectation step (E-step) and maximization step (M-step) are iterated until convergence is met. 

For  the E step, we will need  only to calculate the probabilities $\gamma^t(i) := P(Q^t=i|\boldsymbol{x}^{0:T},\boldsymbol{\lambda}')$ and $\xi^t(i,j) := P(Q^t=i,Q^{t+1}=j|\boldsymbol{x}^{0:T}, \boldsymbol{\lambda}') $ $i,j=1,...,N$, $t=0,...,T$, which are related in the following manner: $\sum_{j=1}^N\xi^t(i,j)= \gamma^t(i)$.

For the M step, we must derive the updating formulas for parameter $\boldsymbol{\lambda}$. For $\pi_i$ and $a_{ij}$ for the hidden states $i,j=1,...,N$, are:
\begin{equation*}\label{eq:uppi}
\begin{aligned}
\pi_i^* &= \gamma^0(i), \\
a_{ij}^* &= \frac{\sum_{t=0}^{T-1}\xi^t(i,j)}{\sum_{t=0}^T\gamma^t(i)}.
\end{aligned}
\end{equation*}  
The updating formula for parameter $\textbf{B}$  relies on the assumptions made over the transition and emission probabilities. For example, in \cite{Lra90}, the updating formulas are calculated when the emission probabilities are assumed to be discrete, a mixture of Gaussians (MoG) or a mixture of AR Gaussians (AR-MoG). If the hypotheses about the transition probabilities or the initial distribution change, the formulas given above are no longer valid.

\subsection{The SEM algorithm} 

When we deal with an unknown a priori  probabilistic graphical model $\mathcal{B}$, it is desirable to find the structure that maximizes the likelihood of the data. However, as many parameters are used in dense networks, the likelihood improves but it can be due to data overfitting. Therefore, penalized likelihood-based scores such as the Bayesian information criterion (BIC) or Akaike information criterion are used in structure optimization algorithms to prevent this issue. In \cite{Fri98}, the structural EM (SEM) algorithm is introduced with its convergence and optimality properties. SEM finds both the desired model and the parameters. SEM tries to maximize the function $\mathcal{Q}(\mathcal{B}, \boldsymbol{\lambda}|\mathcal{B}',\boldsymbol{\lambda}')$, where $\mathcal{B}'$ is a previous or a prior graphical model:
\begin{equation}\label{eq:quijSEM}
\mathcal{Q}(\mathcal{B},\boldsymbol{\lambda}|\mathcal{B}',\boldsymbol{\lambda}') = E_{P(\boldsymbol{q}^{0:T}|\boldsymbol{x}^{0:T},\mathcal{B}',\boldsymbol{\lambda}')}[\ln P(\boldsymbol{x}^{0:T},\boldsymbol{q}^{0:T}|\mathcal{B},\boldsymbol{\lambda})]-0.5\#(\mathcal{B})\ln(T).
\end{equation}
Eq.~(\ref{eq:quijSEM}) considers changes in the structure of the probabilistic graphical model and its parameters, and to prevent overfitting, it is penalized by the number of parameters in the model $\#(\mathcal{B})$ and the logarithm of the length of the data $T$. The SEM algorithm consists of using the EM algorithm with a prior model, $\mathcal{B}'$, and a prior parameter $\boldsymbol{\lambda}'$ to obtain the parameters $\boldsymbol{\lambda}''$; then, using the latent probabilities $P(\boldsymbol{q}^{0:T}|\boldsymbol{x}^{0:T},\mathcal{B}',\boldsymbol{\lambda}'')$ and the parameters $\boldsymbol{\lambda}''$ finding a new structure $\mathcal{B}''$ by solving $\max_\mathcal{B}\mathcal{Q}(\mathcal{B},\boldsymbol{\lambda}''|\mathcal{B}',\boldsymbol{\lambda}'')$. Finally, the EM is again applied to the new structure. This process is iterated until convergence is met.

\subsection{ The Yule-Walker equations}

The Yule-Walker equations \cite{Bo76}, will be a key issue in constructing the proposed model. A linear AR process with $k$ time lag coefficients for a one-dimensional variable $Y^t$ can be described as:
\begin{equation}\label{eq:arpro}
Y^t = \phi_{k1}Y^{t-1}+\cdots+\phi_{kk}Y^{t-k}+\epsilon^t
\end{equation}
where $\epsilon^t\sim\mathcal{N}(0,\sigma^2)$ is an error term following a Gaussian distribution with mean zero and variance $\sigma^2$. Correlogram function  $\rho_k$ returns the correlation between $Y^t$ and  $Y^{t-k}$. We define $\bar{Y}^t := Y^t-\mu_Y$ where $\mu_Y$ is the mean of $Y^t$ and $\zeta_k := E[\bar{Y}^t\bar{Y}^{t-k}]$, which is the expected value of the product of both shifted variables. The correlogram function is computed as:

\begin{equation*}
\rho_k. := \frac{\zeta_k}{\zeta_0}
\end{equation*}

The partial correlogram function  $\Phi(k)$ encodes the correlation between variables $Y^t$ and $Y^{t-k}$ once the effect from intermediary lags has been removed. To determine these partial correlations, observe that for $l\in\{1,...,k\}$:

\begin{equation}\label{eq:yulededuction}
\begin{aligned}
\bar{Y}^t = &\phi_{k1}\bar{Y}^{t-1}+\cdots+\phi_{kk}\bar{Y}^{t-k}+\epsilon^t \\
\bar{Y}^t\bar{Y}^{t-l}  = &\phi_{k1}\bar{Y}^{t-1}\bar{Y}^{t-l}+\cdots+\phi_{kk}\bar{Y}^{t-k}\bar{Y}^{t-l}+\epsilon^t\bar{Y}^{t-l}\\
\rho_l =& \phi_{k1}\rho_{l-1}+\cdots+\phi_{kk}\rho_{k-l}
\end{aligned}
\end{equation}

In the last line of Eq.~(\ref{eq:yulededuction}), we applied the expectation operator and divided by $\zeta_0$. We assumed that $E[\bar{Y}^{t-l}\epsilon^t]=0$ for all $t$, which implies that $Y^t$ is not correlated with the error term; a plausible hypothesis in real situations. Additionally, $\rho(0)=1$. Moreover, notice that if these equations are computed for $l=1,...,k$, we obtain a system of linear equations, which corresponds to the Yule-Walker equations:

	 \begin{equation*}\label{eq:lsyul}
\begin{bmatrix}
\rho_1 \\
\rho_2 \\
\vdots \\
\rho_k \\
\end{bmatrix} 
=
\begin{bmatrix}
1 & \rho_1 & \rho_2 & \cdots & \rho_{k-1}\\
\rho_1& 1 & \rho_1 & \cdots & \rho_{k-2} \\
\vdots & \vdots & \vdots & \ddots & \vdots \\
\rho_{k-1} & \rho_{k-2} &\rho_{k-3} & \cdots & 1 \\
\end{bmatrix}
\begin{bmatrix}
\phi_{k1} \\
\phi_{k2}\\
\vdots \\
\phi_{kk}\\
\end{bmatrix}.
\end{equation*}

The partial correlogram function returns $\Phi(k) := \hat{\phi}_{kk}$. Note that if we wish to evaluate up to $k$ lags for the partial correlogram function, we must construct and solve $k$ linear systems. 

Assume that the sample is white noise. Then the parameter $\hat{\phi}_{kk}$ is distributed approximately as $\mathcal{N}(0,1/T)$. With this information, it is possible to perform hypothesis tests to determine the relevancy of each lag parameter. If $\Phi(p^*)$ is the higher time lag coefficient that is significantly different from zero, then $p^*$ is considered the AR order of the model \cite{Bo76}. 

It is worth mentioning that the previous lag estimation is only useful when the observed data are stationary; in other words, the parameters do not change over time. This particular assumption is violated for the problems in which we want to use HMMs because identifying  changes in the model parameters according to changes in the data distribution is sought. With our proposed model,  however, we will see that the order of the AR process within the HMM will be able to change dynamically  and self adapt depending on the state of the hidden variable.

\section{Proposed model}

The proposed model uses context-specific linear Gaussian Bayesian networks to factorize the emission probabilities. The context is given by the hidden variable. Also, an AR component is added to each variable. The AR order of each variable for each possible context is determined by the SEM algorithm and the Yule-Walker equations when a score (to be specified later on) is optimized. Furthermore, for the proposed model, the likelihood function is modified; therefore, the forward-backward, Viterbi and EM algorithms have to be adapted.

\subsection{Autoregressive asymmetric linear Gaussian hidden Markov models}

Let $p^*_m$ be the AR order (time lag) determined by the Yule-Walker equations and the individual relevancy  hypothesis tests for each variable $X_m^t$, $m=1,...,M$. Set $p^* = \max_{m}{p^*_m}$. For our proposed model we work with the following log-likelihood function which ensures that during the SEM algorithm, the updated structures and AR orders are comparable: 
\begin{equation}\label{eq:logl}
\begin{split}
LL(\boldsymbol{\lambda}) &=\ln P(\boldsymbol{x}^{p^*:T}|\boldsymbol{x}^{0:p^*-1},\boldsymbol{\lambda}) \\&= \ln\sum_{\boldsymbol{q}^{0:T}\in\boldsymbol{R}(\boldsymbol{Q}^{p^*:T})}P(\boldsymbol{q}^{p^*:T},\boldsymbol{x}^{p^*:T}|\boldsymbol{x}^{0:p^*-1},\boldsymbol{\lambda}).
\end{split}
\end{equation}

For this proposed HMM model which is, as explained below, asymmetric autoregressive with linear Gaussian emission probabilities (AR-AsLG-HMM), we modify the emission probabilities $\{b_i(\boldsymbol{x}^t)\}_{i=1}^N$ such that they can be factorized into linear Gaussian Bayesian networks \cite{Sc89} with an asymmetric component \cite{BuM17}, i.e., each variable $X_m$ for each state $i\in R(Q)$ is associated with a set of parents $\textbf{Pa}_i(X_m) = \{U_{im1},...,U_{imk_{im}}\}\subset \{X_1,...,X_M\}$ of size $k_{im}$ (apart from $Q$) which influences its mean in a linear form. Additionally, the emission probabilities are now conditional probabilities given $p_{im}\leq p^*$ past values of the variables $X^t_m$, $m=1,...,M$ (AR terms) for each state $i\in R(Q)$. More specifically, we define: 

\begin{equation}\label{eq:emisionnew2}
\begin{aligned}
b_i^{p^*}(\boldsymbol{x}^t) &= P(\boldsymbol{x}^t|Q^t=i,\boldsymbol{x}^{t-p^*:t-1},\boldsymbol{\lambda}) \\
& = \prod_{m=1}^MP(x^t_m|Q^t=i,x_m^{t-p_{im}:t-1},\text{\bf{Pa}}_i(X_m),\boldsymbol{\lambda})  \\
& = \prod_{m=1}^M\mathcal{N}(x^t_m|\boldsymbol{\beta}_{im}\cdot\textbf{pa}^t_{im} + \boldsymbol{\eta}_{im}\cdot\boldsymbol{d}_{im}^t, \sigma_{im}^2)
\end{aligned}
\end{equation}

\begin{figure}[h]
	\centering
		\begin{tikzpicture}
		[->,thick, scale=1,auto=center]
		\node (q1)[style={circle,fill=black!35,draw=black!100}] at (0,0)  {$Q^{t+1}$};
		\node (q2)[style={circle,fill=black!35,draw=black!100}] at (2.5,0)  {$Q^{t+2}$};
		\node (q3)[style={circle,fill=black!35,draw=black!100}] at (5,0)  {$Q^{t+3}$};
		
		\draw (0,3) ellipse (0.9cm and 2cm)  node[left,rotate=90,xshift=1.cm,yshift=1.2cm] {$b_{1}^{p^*}(\boldsymbol{x}^{t+1})$};
		\node (x11)[style={circle,fill=gray!20,draw=black!100}] at (0,2)  {$X_1^{t+1}$};
		\node (x12)[style={circle,fill=gray!20,draw=black!100}] at (0,4)  {$X_2^{t+1}$};
		\draw (0,-3) ellipse (0.9cm and 2cm) node[left,rotate=90,xshift=1.cm,yshift=1.2cm] {$b_{2}^{p^*}(\boldsymbol{x}^{t+1})$};
		\node (x13)[style={circle,fill=gray!20,draw=black!100}] at (0,-2)  {$X_1^{t+1}$};
		\node (x14)[style={circle,fill=gray!20,draw=black!100}] at (0,-4)  {$X_2^{t+1}$};
		
		\draw (2.5,3) ellipse (0.9cm and 2cm)node[left,rotate=90,xshift=1.cm,yshift=1.2cm] {$b_{1}^{p^*}(\boldsymbol{x}^{t+2})$};
		\node (x21)[style={circle,fill=gray!20,draw=black!100}] at (2.5,2)  {$X_1^{t+2}$};
		\node (x22)[style={circle,fill=gray!20,draw=black!100}] at (2.5,4)  {$X_2^{t+2}$};
		\draw (2.5,-3) ellipse (0.9cm and 2cm)node[left,rotate=90,xshift=1.cm,yshift=1.2cm] {$b_{2}^{p^*}(\boldsymbol{x}^{t+2})$};
		\node (x23)[style={circle,fill=gray!20,draw=black!100}] at (2.5,-2)  {$X_1^{t+2}$};
		\node (x24)[style={circle,fill=gray!20,draw=black!100}] at (2.5,-4)  {$X_2^{t+2}$};
		
		\draw (5,3) ellipse (0.9cm and 2cm)node[left,rotate=90,xshift=1.cm,yshift=1.2cm] {$b_{1}^{p^*}(\boldsymbol{x}^{t+3})$};
		\node (x31)[style={circle,fill=gray!20,draw=black!100}] at (5,2)  {$X_1^{t+3}$};
		\node (x32)[style={circle,fill=gray!20,draw=black!100}] at (5,4)  {$X_2^{t+3}$};
		\draw (5,-3) ellipse (0.9cm and 2cm)node[left,rotate=90,xshift=1.cm,yshift=1.2cm] {$b_{2}^{p^*}(\boldsymbol{x}^{t+3})$};
		\node (x33)[style={circle,fill=gray!20,draw=black!100}] at (5,-2)  {$X_1^{t+3}$};
		\node (x34)[style={circle,fill=gray!20,draw=black!100}] at (5,-4)  {$X_2^{t+3}$};

		\draw (-1.5,0) -> (q1);
		\draw (q1) -> node[below] {$a_{q^{t+1}q^{t+2}}$}(q2);
		\draw (q2) -> node[below] {$a_{q^{t+2}q^{t+3}}$}(q3);
		\draw (q3) -> (6.5,0);

		\draw (q1) -> node[left] {$Q^{t+1}=1$}(0,1);
		\draw (q1) -> node[left] {$Q^{t+1}=2$}(0,-1);
		\draw (q2) -> node[left] {$Q^{t+2}=1$} (2.5,1);
		\draw (q2) -> node[left] {$Q^{t+2}=2$} (2.5,-1);
		\draw (q3) -> node[left] {$Q^{t+3}=1$}(5,1);
		\draw (q3) -> node[left] {$Q^{t+3}=2$}(5,-1);

		\draw (x11) -> (x12);
		\draw (x21) -> (x22);
		\draw (x31) -> (x32);		
		
		\draw[->] (-1.5,2) to[out=-0,in=180] (x11);
		\draw[->] (x11) to[out=-0,in=180] (x21);
		\draw[->] (x21) to[out=-0,in=180] (x31);
		\draw[->] (x31) to[out=-0,in=180] (6.5,2);

		\draw[->] (x14) to[out=0,in=-180] (x24);
		\draw[->] (-1.5,-4) to[out=-0,in=180] (x14);
		\draw[->] (x24) to[out=0,in=-180] (x34);
		\draw[->] (x34) to[out=0,in=-180] (6.5,-4);
		
		\draw[->] (-1.5,4) to[out=0,in=-180] (x12);
		\draw[->] (x12) to[out=0,in=-180] (x22);
		\draw[->] (-1.5,5) to[out=60,in=120,distance=1cm] (x22);
		\draw[->] (-1.5,4.5) to[out=60,in=120,distance=0.5cm] (x12);
		\draw[->] (x22) to[out=0,in=-180] (x32);
		\draw[->] (x12) to[out=60,in=120,distance=1.5cm] (x32);
		\draw[->] (x32) to[out=0,in=-180] (6.5,4);
		\draw[->] (x22) to[out=60,in=120] (6.5,5);
		\draw[->] (x32) to[out=60,in=120,distance=0.5cm] (6.5,4.5);
		\end{tikzpicture}
		\caption{Graphical representation of an AR-AsLG-HMM model}
		\label{fig:arhmmgraph}
\end{figure}
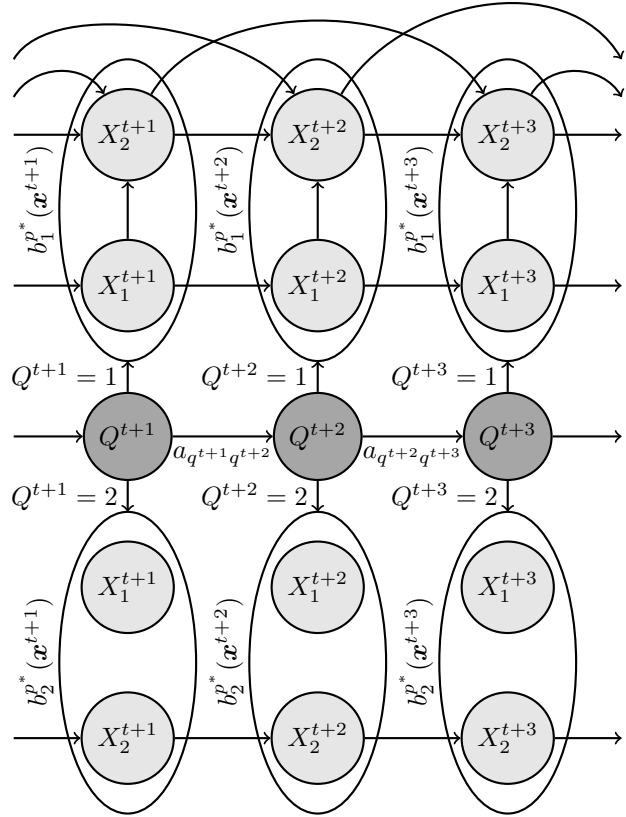

In Eq.~(\ref{eq:emisionnew2}), we have  $\boldsymbol{\beta}_{im} = (\beta_{im0},...,\beta_{imk_{im}})$, $\textbf{pa}^t_{im} = (1,u^t_{im1},...,u^t_{imk_{im}})$,  $\boldsymbol{\eta}_{im} = (\eta_{im1},...,\eta_{imp_{im}})$ and $\boldsymbol{d}_{im}^t =(x_m^{t-1},...,x_m^{t-p_{im}})$. Fig.~\ref{fig:arhmmgraph} shows an example of  an AR-AsLG-HMM. In this example, when $Q^t=1$ (top), variable $X_2^t$ is dependent on $Q^t$, $X_1^t$, $X_2^{t-1}$ and $X_2^{t-2}$, but $X_1^t$ depends only on $Q^t$ and $X_1^{t-1}$. However, when $Q^t=2$ (bottom), $X_1^t$ depends only on $Q^t$ and $X_2^t$ is dependent on $X_2^{t-1}$ and $Q^t$. In terms of the model, this can be expressed as  $p_{11}=1$, $p_{12}=2$ AR terms, $k_{11}=0$ and $k_{12}=1$ when $Q=1$, and $p_{21}= 0$, $p_{22}= 1$ AR terms, $k_{21} = 0$ and $k_{22}=0$ when $Q=2$. From the model we can see that $p^*\geq 2$, because $p_{im}\leq2$ for $i=1,2$ and $m=1,2$.  

Some comments on Eq.~(\ref{eq:emisionnew2}) follow. The set of parents  $\textbf{Pa}_i(X_m)$ of each variable $X_m$ for each state $i \in R(Q)$ is related to a context-specific Bayesian network $\mathcal{B}_i$. Furthermore, depending on that hidden state, each variable  $X_m$  may have a different AR order, namely $p_{im}$, which is upper bounded by $p^*$. This model must estimate the new parameters $\{\beta_{im0},...,\beta_{imk_{im} },\eta_{im1},...,\eta_{imp_{im}},\sigma_{im}^2\}_{m=1,i=1}^{M,N}$. Additionally,  because the first $p^*$ observations are used as conditionals in Eq.~(\ref{eq:logl}), the $\boldsymbol{\pi}$ parameter is shifted to predict the initial distribution of the $Q^{p^*}$ hidden variable, i.e., $\{\pi_{i}\}_{i=1}^N = \{P(Q^{p^*}=i|\boldsymbol{\lambda})\}_{i=1}^N$. Observe that the complete information probability of an instance $\boldsymbol{x}^{p^*:T}$ of $\boldsymbol{X}^{p^*:T}$ and an instance $\boldsymbol{q}^{p^*:T}$ of $\boldsymbol{Q}^{p^*:T}$ can be expressed as:
\begin{equation*}\label{eq:fullinfonew}
P(\boldsymbol{q}^{p^*:T},\boldsymbol{x}^{p^*:T}|\boldsymbol{x}^{0:p^*-1},\lambda)= \pi_{q^{p^*}}\prod_{t=p^*}^{T-1}a_{q^tq^{t+1}}\prod_{t=p^*}^Tb_{q^t}^{p^*}(\boldsymbol{x}^t).
\end{equation*}

\subsection{Feasibility of the EM algorithm in AR-AsLG-HMMs}
To perform the parameter learning, the EM algorithm can be applied. However, we must define an auxiliary function $\mathcal{Q}$ for the log-likelihood defined in  Eq.~(\ref{eq:logl}). We propose $\mathcal{Q}^{p^*}(\boldsymbol{\lambda}|\boldsymbol{\lambda}')$ as the auxiliary function for the EM algorithm, defined as:
\begin{equation}\label{eq:Qnew}
\begin{split}
&\mathcal{Q}^{p^*}(\boldsymbol{\lambda}|\boldsymbol{\lambda}') \\
& = \sum_{\boldsymbol{R}(\boldsymbol{Q}^{p^*:T})} P(\boldsymbol{q}^{p^*:T}|\boldsymbol{x}^{0:T}, \boldsymbol{\lambda}')\ln P(\boldsymbol{q}^{p^*:T},\boldsymbol{x}^{p^*:T}|\boldsymbol{x}^{0:p^*-1},\boldsymbol{\lambda}).
\end{split}
\end{equation}
Moreover, $\mathcal{Q}^{p^*}(\boldsymbol{\lambda}|\boldsymbol{\lambda}')$ can be decomposed as:
\begin{equation}\label{eq:decomposed}
\begin{aligned}
&\mathcal{Q}^{p^*}(\boldsymbol{\lambda}|\boldsymbol{\lambda}')= \sum_{\boldsymbol{R}(\boldsymbol{Q}^{p^*:T})} P(\boldsymbol{q}^{p^*:T}|\boldsymbol{x}^{0:T}, \boldsymbol{\lambda}')\ln P(\boldsymbol{q}^{p^*:T}|\boldsymbol{x}^{0:T},\boldsymbol{\lambda})\\
&\qquad \quad  +\ln P(\boldsymbol{x}^{p^*:T}|\boldsymbol{x}^{0:p^*-1},\boldsymbol{\lambda})\sum_{\boldsymbol{R}(\boldsymbol{Q}^{p^*:T})} P(\boldsymbol{q}^{p^*:T}|\boldsymbol{x}^{0:T}, \boldsymbol{\lambda}')\\
&= \sum_{\boldsymbol{R}(\boldsymbol{Q}^{p^*:T})} P(\boldsymbol{q}^{p^*:T}|\boldsymbol{x}^{0:T}, \boldsymbol{\lambda}')\ln P(\boldsymbol{q}^{p^*:T}|\boldsymbol{x}^{0:T},\boldsymbol{\lambda}) +LL(\boldsymbol{\lambda})\\
\end{aligned}
\end{equation}
If we define $\mathcal{H}^{p^*}(\boldsymbol{\lambda}|\boldsymbol{\lambda}')$ as the first summand of Eq.~(\ref{eq:decomposed}), i.e.,
\begin{equation*}
\mathcal{H}^{p^*}(\boldsymbol{\lambda}|\boldsymbol{\lambda}'):= \sum_{\boldsymbol{R}(\boldsymbol{Q}^{p^*:T})} P(\boldsymbol{q}^{p^*:T}|\boldsymbol{x}^{0:T}, \boldsymbol{\lambda}')\ln P(\boldsymbol{q}^{p^*:T}|\boldsymbol{x}^{0:T},\boldsymbol{\lambda})
\end{equation*}
therefore we have that $\mathcal{Q}^{p^*}(\boldsymbol{\lambda}|\boldsymbol{\lambda}') =  \mathcal{H}^{p^*}(\boldsymbol{\lambda}|\boldsymbol{\lambda}')+ LL(\boldsymbol{\lambda})$. We now show that if we apply the EM algorithm with $\mathcal{Q}^{p^*}(\boldsymbol{\lambda}|\boldsymbol{\lambda}')$, each iteration does not decrease the log-likelihood as required. 

\begin{lemma}\label{lem:Q} Let $\boldsymbol{\lambda}^{(s)}$ be the parameters  at iteration $s$ of the EM and $\boldsymbol{\lambda}^{(s+1)}$ be the resulting parameters after the next iteration of the EM. We have that $\mathcal{Q}^{p^*}(\boldsymbol{\lambda}^{(s+1)}|\boldsymbol{\lambda}^{(s)})\geq\mathcal{Q}^{p^*}(\boldsymbol{\lambda}^{(s)}|\boldsymbol{\lambda}^{(s)})$. \begin{flushright}
$\blacksquare$
\end{flushright}
\end{lemma}

\begin{lemma}\label{lem:H}
	Given two arbitrary models with respective parameters $\boldsymbol{\lambda}$ and $\boldsymbol{\lambda}'$, we have that $\mathcal{H}^{p^*}(\boldsymbol{\lambda}|\boldsymbol{\lambda}')\leq \mathcal{H}^{p^*}(\boldsymbol{\lambda}'|\boldsymbol{\lambda}')$, and the equality holds when $ P(\boldsymbol{q}^{p^*:T}|\boldsymbol{x}^{0:T},\boldsymbol{\lambda})=  P(\boldsymbol{q}^{p^*:T}|\boldsymbol{x}^{0:T},\boldsymbol{\lambda}')$.\begin{flushright}
		$\blacksquare$
	\end{flushright}
\end{lemma}

\begin{theorem}\label{theo:likeli}
	Let $\boldsymbol{\lambda}^{(s)}$ be the parameters  at an iteration $s$ of the EM and $\boldsymbol{\lambda}^{(s+1)}$ be the resulting parameters after the next iteration of the EM. We have that
	\begin{itemize}
		\item[(a)] $LL(\boldsymbol{\lambda}^{(s+1)})\geq LL(\boldsymbol{\lambda}^{(s)})$. In other words, the log-likelihood of the model cannot worsen after an EM iteration. 
		\item[(b)] The sequence $\{LL(\boldsymbol{\lambda}^{(s)})\}_{s}$ converges.
	\end{itemize}
\begin{flushright}
	$\blacksquare$
\end{flushright}
\end{theorem}

\subsection{The forward-backward algorithm in AR-AsLG-HMMs}
As the likelihood function of Eq.~(\ref{eq:logl}) and the emission probabilities given by Eq.~(\ref{eq:emisionnew2}) have changed, the forward-backward algorithm must be adapted. In the E step, we compute the probabilities $\gamma^t(i)=P(Q^t=i|\boldsymbol{x}^{0:T},\boldsymbol{\lambda})$ for $t=0,...,T$ and $i=1,...,N$ as the initial point to fit the forward-backward algorithm. Note that $\gamma^t(i)$ can be expressed as:
\begin{equation}\label{eq:newgam}
\begin{aligned}
&\gamma^t(i) = \frac{P(Q^t=i,\boldsymbol{x}^{p^*:T}|\boldsymbol{x}^{0:p^*-1},\boldsymbol{\lambda})}{P(\boldsymbol{x}^{p^*:T}|\boldsymbol{x}^{0:p^*-1},\boldsymbol{\lambda})} \\
& = \frac{P(Q^t=i,\boldsymbol{x}^{p^*:t},\boldsymbol{x}^{t+1:T}|\boldsymbol{x}^{0:p^*-1},\boldsymbol{\lambda})}{P(\boldsymbol{x}^{p^*:T}|\boldsymbol{x}^{0:p^*-1},\boldsymbol{\lambda})} \\
& = \frac{P(\boldsymbol{x}^{t+1:T}|Q^t=i,\boldsymbol{x}^{0:t},\boldsymbol{\lambda})P(Q^t=i,\boldsymbol{x}^{p^*:t}|\boldsymbol{x}^{0:p^*-1},\boldsymbol{\lambda})}{P(\boldsymbol{x}^{p^*:T}|\boldsymbol{x}^{0:p^*-1},\boldsymbol{\lambda})} \\
& = \frac{\beta^{t}_{p^*}(i)\alpha_{p^*}^t(i)}{\sum_{j=1}^N\beta_{p^*}^{t}(j)\alpha_{p^*}^t(j)}
\end{aligned}
\end{equation}

From Eq.~(\ref{eq:newgam}), the forward variable is $\alpha_{p^*}^t(i):= P(Q^t=i,\boldsymbol{x}^{p^*:t}|\boldsymbol{x}^{0:p^*-1},\boldsymbol{\lambda})$ and the backward variable is $\beta_{p^*}^t(i) := P(\boldsymbol{x}^{t+1:T}|Q^t=i,\boldsymbol{x}^{0:t},\boldsymbol{\lambda})$. Observe that these equations only make sense when $t\geq p^*$. The next lemma shows that we can easily adapt the forward-backward algorithm to compute the $\alpha_{p^*}$ and $\beta_{p^*}$ parameters of an AR-AsLG-HMM.

\begin{lemma}\label{lem:forward-backward}
	$\alpha^t_{p^*}(i)$ and $\beta^t_{p^*}(i)$ can be computed as:
	\begin{equation*}
	\begin{aligned}
	\alpha^t_{p^*}(i)& =  \sum_{j=1}^N b_i^{p^*}(\boldsymbol{x}^t)a_{ji}\alpha^{t-1}(j) \\
	\beta^t_{p^*}(i)   & =  \sum_{j=1}^{N} \beta^{t+1}(j)b_j^{p^*}(\boldsymbol{x}^{t+1})a_{ij}
	\end{aligned}
	\end{equation*}
	for $t=p^*,...,T$ and $i=1,...,N$, with initial values $\alpha^{p^*}_{p^*}(i) = \pi_{i}b_i^{p^*}(\boldsymbol{x}^{p^*})$ and $\beta^T_{p^*}(i)= 1$, $i=1,...,N$.
	\begin{flushright}
		$\blacksquare$
	\end{flushright}
\end{lemma}

\subsection{Parameter learning in AR-AsLG-HMMs}
To execute the EM algorithm, we must iterate the E step and the M step. For the E step we can use the adapted forward-backward algorithm of Section 4.3 to compute $\gamma^t(i)$ and $\xi^t(i,j)$ for $i,j=1,...,N$ and $t=0,...,T$:
\begin{equation}\label{eq:newestep}
\begin{aligned}
\gamma^t(i) &= \frac{\beta_{p^*}^{t}(i)\alpha_ {p^*}^t(i)}{\sum_{j=1}^N\beta_{p^*}^{t}(j)\alpha_{p^*}^t(j)}\\ 
\xi^t(i,j) &= \frac{\alpha_{p^*}^t(i)a_{ij}b^{p^*}_j(\boldsymbol{x}^{t+1})\beta_ {p^*}^{t+1}(j)}{\sum_ {u,v=1}^N\alpha_ {p^*}^t(u)a_{uv}b^{p^*}_v(\boldsymbol{x}^{t+1})\beta_{p^*}^{t+1}(v)}
\end{aligned}
\end{equation}
Computing these quantities is enough for the E step because $\mathcal{Q}^{p^*}(\boldsymbol{\lambda}|\boldsymbol{\lambda}')$ can be expressed as:
\begin{equation}\label{eq:likelihoodar}
\mathcal{Q}^{p^*}(\boldsymbol{\lambda}|\boldsymbol{\lambda}') = \sum_{i=1}^{N}\gamma^{p^*}(i)\ln\pi_i+\sum_{t=p^*}^{T-1}\sum_{i=1}^{N}\sum_{j=1}^N\xi^t(i,j)\ln a_{ij}
+ \sum_{t=p^*}^T\sum_{i=1}^N\gamma^t(i)\ln b_i^{p^*}(\boldsymbol{x}^t).
\end{equation}

Now, for the M step, we must find the updating formulas for the parameters $(\textbf{A},\textbf{B},\boldsymbol{\pi})$, where $\textbf{B}$ includes the parameters $\eta_{imr}$, $\beta_{imk}$ and $\sigma^2_{im}$ of the Gaussian. In the following theorem, we provide the updating formulas for the proposed model.

\begin{theorem}\label{theo:update_formulas}The M-step for an AR-AsLG-HMM model can be performed using the following updating formulas:
parameter $\boldsymbol{\pi}=\{\pi_i\}_{i=0}^{N}$ is updated as:
	\begin{equation}\label{eq:uppinewtheo}
	\pi_i^* = \gamma^{p^*}(i).
	\end{equation}  
The parameter $\textbf{A} = \{a_{ij}\}_{i,j=1}^{N} $ is updated as:
	\begin{equation}\label{eq:updateaijtheo}
	a_{ij}^* = \frac{\sum_{t=p^*}^{T-1}\xi^t(i,j)}{\sum_{t=p^*}^{T-1}\gamma^t(i)}.
	\end{equation} 

If we set $f^t_{im} := \boldsymbol{\beta}_{im}\cdot\textbf{pa}^t_{im} + \boldsymbol{\eta}_{im}\cdot\boldsymbol{d}_{im}^t$, the parameters $\{\eta_{imr}\}_{r=1}^{p_{im}}$, $\{\beta_{imk}\}_{k=0}^{k_{im}}$ can be updated jointly, solving the following linear system:
	\begin{equation}\label{eq:sislintheo}
\begin{cases}
\sum_{t=p^*}^{T}\gamma^{t}(i)x^t_m= \sum_{t=p^*}^{T}\gamma^{t}(i)f^t_{im}\\
\sum_{t=p^*}^{T}\gamma^{t}(i)x^t_mu^t_{im1} = \sum_{t=p^*}^{T}\gamma^{t}(i)u^t_{im1}f^t_{im}\\
\qquad \vdots  \qquad\qquad\qquad   \vdots \qquad\qquad\qquad   \vdots \\
\sum_{t=p^*}^{T}\gamma^{t}(i)x^t_mu^t_{imk_{im}} = \sum_{t=p^*}^{T}\gamma^{t}(i)u^t_{imk_{im}}f^t_{im} \\
\sum_{t=p^*}^{T}\gamma^{t}(i)x^t_mx^{t-1}_{m} = \sum_{t=p^*}^{T}\gamma^{t}(i)x^{t-1}_{m}f^t_{im} \\
\qquad \vdots  \qquad\qquad\qquad   \vdots \qquad\qquad\qquad   \vdots  \\
\sum_{t=p^*}^{T}\gamma^{t}(i)x^t_mx^{t-p_{im}}_{m} = \sum_{t=p^*}^{T}\gamma^{t}(i)x^{t-p_{im}}_{m}f^t_{im} \\
\end{cases}
\end{equation}

If we set  $\hat{f}_{im}^t:=\beta^*_{im0}+ \beta^*_{im1} u_{im1}^{t}+\cdots+\beta^*_{imk_{im}}u_{imk_{im}}^{t}+\eta^*_{im1}x^{t-1}_m+\cdots+\eta^*_{imp_{im}}x^{t-p_{im}}_m$, then, $\sigma_{im}^2$ can be updated as:
\begin{equation}\label{eq:vartheo}
(\sigma_{im}^2)^*  = \frac{\sum_{t=p^*}^T\gamma^t(i)(x_m^t-\hat{f}_{im}^t)^2}{\sum_{t=p^*}^T\gamma^t(i)}.
\end{equation}
This update must be done for every variable $m=1,...,M$ and hidden state $i=1,...,N$.
\begin{flushright}
	$\blacksquare$
\end{flushright}
\end{theorem}

Eq.~(\ref{eq:sislintheo}) forms a linear system of $k_{im}+p_{im}+1$ unknowns with $k_{im}+p_{im}+1$ equations. If the resulting  context-specific Bayesian model for every hidden state is a naïve Bayesian network and $p_{im}=0$ for $i=1,...,N$ and $m=1,...,M$, then we only require to update the parameters $\{\beta_{im0}\}_{i=1,m=1}^{N,M}$. Its updating formula is:
\begin{equation*}
\beta_{im0}^*= \frac{\sum_{t=p^*}^{T}\gamma^{t}(i)x^t_m}{\sum_{t=p^*}^{T}\gamma^{t}(i)}
\end{equation*}

Otherwise, we must solve the linear system which can be done using exact or iterative methods. If we use for example the Gauss-Jordan reduction algorithm to solve the linear system, an additional computational cost of $O((k_{im}+p_{im}+1)^3)$ must be assumed. Therefore, simpler structures are recommended in order to not slow down the learning process. This requirement is taken into account during the SEM algorithm as mentioned in Section 3.5.
 
A pseudocode of the adapted EM algorithm can be found in Fig.~\ref{alg:EM}.

\begin{figure}
\begin{algorithmic}[1]
	\Require{ A prior parameter $\boldsymbol{\lambda}^{(0)}$, a dataset $\boldsymbol{X}^{0:T}$ }
	\Ensure{ A learned parameter $\boldsymbol{\lambda}^*$ }
	\For{$s=0,1,...$ until convergence in likelihood is met:}
	\State{Use Eq.~(\ref{eq:forar}) and Eq.~(\ref{eq:backar}) to perform the adapted forward-backward algorithm with $\boldsymbol{\lambda}^{(s)}$}
	\State{Estimate $\gamma^t(i)$ and $\xi^t(i,j)$ with Eq.~(\ref{eq:newestep})}
	\State{Use Eq.~(\ref{eq:uppinewtheo}) and Eq.~(\ref{eq:updateaijtheo}) to obtain: $\boldsymbol{\pi}^*$ and $\textbf{A}^*$}
	\For{$i=1,...,N$ and $m=1,...,M$ }
	\State{Use Eq.~(\ref{eq:sislintheo}) to obtain:\\ \qquad \qquad  $\{\beta^*_{im0}, \beta^*_{im1},..., \beta^*_{imk_{im}},\eta^*_{im1},...,\eta^*_{imp_{im}}\}$}
	\State{Use Eq.~(\ref{eq:vartheo}) to obtain: $(\sigma_{im}^2)^*$}
	\EndFor
	\EndFor
	\State{\bf{return} $\boldsymbol{\lambda^*}$}
\end{algorithmic}
\caption{Pseudocode for the adapted EM algorithm}
\label{alg:EM}
\end{figure}

\subsection{The Viterbi algorithm in AR-AsLG-HMMs}

In the following lemma we show that the traditional Viterbi algorithm can be adapted to determine the most probable sequence of hidden states in AR-AsLG-HMMs. 
\begin{lemma}\label{lem:vit}
	If $\delta^t_{p^*}(i) = \max_{\boldsymbol{q}^{p^*:t-1}}\{P(\boldsymbol{x}^{p^*:t},\boldsymbol{q}^{p^*:t-1},Q^t=i|\boldsymbol{x}^{0:p^*-1},\boldsymbol{\lambda})\}$ represents the most probable sequence of hidden states up to time $t-1$ for state $i$ at time $t$, then $\delta^t_{p^*}(i)$ can be computed recursively.
	\begin{equation*}
	\delta^t_{p^*}(i) = \max_{j=1,...,N}\{\delta_{p^*}^{t-1}(j)a_{ji}\}b^{p^*}_i(\boldsymbol{x^t})
	\end{equation*}
	The Viterbi algorithm is initialized with $\delta^{p^*}_{p^*}(i) = \pi_ib^{p^*}_i(\boldsymbol{x}^{p^*})$.
	\begin{flushright}
		$\blacksquare$
	\end{flushright}
\end{lemma}

\subsection{The SEM algorithm in AR-AsLG-HMMs}

Regarding the structural optimization process, the SEM algorithm for AR-AsLG-HMM must  also be modified. The proposed auxiliary function is:
\begin{equation}\label{eq:quijSEMAR}
\mathcal{Q}^{p^*}(\mathcal{B},\boldsymbol{\lambda}|\mathcal{B}',\boldsymbol{\lambda}') = E_{P(\boldsymbol{q}^{p^*:T}|\boldsymbol{x}^{0:T},\mathcal{B}',\boldsymbol{\lambda}')}[\ln P(\boldsymbol{x}^{p^*:T},\boldsymbol{Q}^{p^*:T}|\mathcal{B},\boldsymbol{x}^{0:p^*-1},\boldsymbol{\lambda})]-0.5\#(\mathcal{B})\ln(T).
\end{equation}
The steps for the adapted SEM algorithm are the same as in the general SEM. However, we must consider that given a time slice $t$, the algorithm must not only look for the best instantaneous structure at time $t$ or the best structure with variables $(X^t_1,...,X^t_M)$ but also look for the best  transition structure at time $t$  or the relationships between $(X^t_1,...,X^t_M)$ variables and their AR versions, i.e., $(X^{t-1}_1,X^{t-2}_1,...,X_M^{t-p^*-1},X^{t-p^*}_M)$, which implies that the search space dimension increases. More specifically, we have to search not only in the space of directed acyclic graphs (DAGs) for the best instantaneous structures, but also in the space $S_{p^*} = \{0,1,...,p^*\}^{N}\times\{0,1,...,p^*\}^{M}$, for the best transition structure. A component $p_{im}$ of a vector $\boldsymbol{p}\in S_{p^*}$ indicates the number of lags for variable $X_m$ in the hidden state $i\in R(Q)$. For instance, if $p_{im}=2$, $X_m^t$ has incoming arcs from the variables $X_m^{t-2}$ and $X_m^{t-1}$ when $Q^t=i$. A pseudocode of the adapted SEM is given in Fig.~\ref{alg:SEM}.

It is pertinent to mention that in the SEM algorithm in the step of finding $\mathcal{B}^{(s)}=\arg\max_{\mathcal{B}}\mathcal{Q}^{p^*}(\mathcal{B},\boldsymbol{\lambda}^{(s)}|\mathcal{B}^{(s-1)},\boldsymbol{\lambda}^{(s)})$ it is not necessary to use  Eq.~(\ref{eq:quijSEMAR}), since the initial distribution and the transition matrix are being unchanged. We can take advantage of the linearity of Eq.~(\ref{eq:quijSEMAR}) to compare structures, i.e. if a dependency of $X_m$ has been added or deleted (AR or parent parameter) at the hidden state $i$, it is reasonable to use the following score:
\begin{equation}\label{eq:local_score}
\text{score}_{im} = \sum_{t=p^*}^{T}\ln(\mathcal{N}(x_m^t|f_{im}^t,\sigma^2_{im})^{\gamma^t(i)})
\end{equation}

If changes have been done to many variables in many hidden states, it is better to use the following score:
\begin{equation}\label{eq:score}
\text{score} = \sum_{i=1}^N\sum_{m=1}^M\text{score}_{im}
\end{equation}

To perform the structural optimization step, we must search in the space of structures. In this article we use a heuristic forward greedy algorithm to perform the structure optimization. In this approach, we initialize all the structures in a naïve form with no AR parameters. During the optimization, we visit each variable for each hidden state and add AR or parent dependencies as long as Eq.~(\ref{eq:local_score}) improves. Its pseudocode can be seen in Fig.~\ref{alg:greedy}. 

\begin{figure}
	\begin{algorithmic}[1]
		\Require{ A prior parameter $\boldsymbol{\lambda}^{(0)}$, a dataset $\boldsymbol{X}^{0:T}$, a prior structure $\mathcal{B}^{(0)}$}
		\Ensure{  A learned parameter $\boldsymbol{\lambda}^*$ and structure $\mathcal{B}^*$  }
		\State{Solve $\boldsymbol{\lambda}^{(1)}=\arg\max_{\boldsymbol{\lambda}}\mathcal{Q}^{p^*}(\mathcal{B}^{(0)},\boldsymbol{\lambda}|\mathcal{B}^{(0)},\boldsymbol{\lambda}^{(0)})$ with the EM algorithm }
		\For{ $s=1,...$ until convergence in the penalized  log-likelihood is met}
		\State{\quad Solve $\mathcal{B}^{(s)}=\arg\max_{\mathcal{B}}\mathcal{Q}^{p^*}(\mathcal{B},\boldsymbol{\lambda}^{(s)}|\mathcal{B}^{(s-1)},\boldsymbol{\lambda}^{(s)})$ with any meta-heuristic method}  
		\Comment{Here we obtain a DAG for the instantaneous time structure, and a vector $\boldsymbol{p}$ for the transition time structure}
		\State{\quad Solve $\boldsymbol{\lambda}^{(s+1)}=\arg\max_{\boldsymbol{\lambda}}\mathcal{Q}^{p^*}(\mathcal{B}^{(s)},\boldsymbol{\lambda}|\mathcal{B}^{(s-1)},\boldsymbol{\lambda}^{(s)})$ with the EM algorithm}
		\EndFor
		\State{\textbf{return} $\boldsymbol{\lambda}^*$ and $\mathcal{B}^*$ }
	\end{algorithmic}
	\caption{Pseudo-code for the adapted SEM algorithm}
	\label{alg:SEM}
\end{figure}

\begin{figure}
	\begin{algorithmic}[1]
		\Require{ A parameter $\boldsymbol{\lambda}^{(s)}$, a prior structure $\mathcal{B}^{(s-1)}$}
		\Ensure{ A structure $\mathcal{B}^{(s)}$  }
		\For{ $i=1,...,N$ } \Comment{Optimization of AR structures}
		\quad \For{$m=1,...,M$}
		\quad \State{Compute $\text{score}_{im}$ with $\mathcal{B}^{(s-1)}$  and $\boldsymbol{\lambda}^{(s)}$ }
		\quad \While{$p_{ik}\leq p^*$}
		\quad \quad \State{Define $\widehat{\mathcal{B}}$ with $\hat{p}_{ik} := p_{ik}+1$ and estimate its parameters $\widehat{\boldsymbol{\lambda}} = \arg\max_{\boldsymbol{\lambda}}\mathcal{Q}^{p^*}(\widehat{\mathcal{B}},\boldsymbol{\lambda}|\mathcal{B}^{(s-1)},\boldsymbol{\lambda}^{(s)})$  }
		\quad \quad \State{Compute $\widehat{\text{score}_{im}}$ with $\widehat{\mathcal{B}}$  and $\widehat{\boldsymbol{\lambda}}$ }
		\quad \quad \If{ $\widehat{\text{score}_{im}}>\text{score}_{im}$}
		\quad \quad \quad \State{Update $\mathcal{B}^{(s-1)}$, $\text{score}_{im}$ and $\boldsymbol{\lambda}^{(s)}$}
		\quad \quad \Else{ \textbf{Break}}
		\quad \quad \EndIf
		\quad \EndWhile
		\quad \EndFor
		\EndFor
		\For{ $i=1,...,N$}\Comment{Optimization of non-AR structures}
		\quad \For{$m=1,...,M$}
		\quad \State{Compute $\boldsymbol{\mathcal{B}_{im}}$:= all the possible DAG graphs resulting by adding one arc to $\mathcal{B}^{(s-1)}$ in its $i$ context-specific Bayesian network, where $X_m$ is a new children of any variable}
		\quad \If{$\boldsymbol{\mathcal{B}_{im}}$ is non-empty}
		\quad \quad \State{Compute $\text{score}_{im}$ with $\mathcal{B}^{(s-1)}$ and $\boldsymbol{\lambda}^{(s)}$}
		\quad \quad \For{ $\widehat{\mathcal{B}}$ in $\boldsymbol{\mathcal{B}_{im}}$ }
		\quad \quad \quad \State{Estimate the parameters $\widehat{\boldsymbol{\lambda}}= \arg\max_{\boldsymbol{\lambda}}\mathcal{Q}^{p^*}(\widehat{\mathcal{B}},\boldsymbol{\lambda}|\mathcal{B}^{(s-1)},\boldsymbol{\lambda}^{(s)})$ }
		\quad \quad \quad \State{Compute $\widehat{\text{score}_{im}}$ with $\hat{\mathcal{B}}$ and $\widehat{\boldsymbol{\lambda}}$}
		\quad \quad \quad  \If{ $\widehat{\text{score}_{im}}>\text{score}_{im}$}
		\quad \quad \quad \State{Update $\mathcal{B}^{(s-1)}$, $\text{score}_{im}$ and $\boldsymbol{\lambda}^{(s)}$}
		\quad \quad \quad \EndIf
		\quad \quad \EndFor
		\quad \EndIf
		\quad \EndFor
		\EndFor
		\State{Set $\mathcal{B}^{(s)} := \mathcal{B}^{(s-1)}$  }	
	\end{algorithmic}
	\caption{Pseudo-code for the forward greedy algorithm}
	\label{alg:greedy}
\end{figure}

\subsection{Hidden states labelling}

In practice, when HMMs are used, categorical labels are given to the hidden states for interpretation purposes. However, only after  training the model, the model parameters are manually checked to determine which categorical label corresponds with each trained hidden state. Here, we propose an automatic numerical labelling for trained models, where a numerical function is used to label a trained hidden state. Let $g:R(Q)\rightarrow \mathcal{R}$ be a function that maps each hidden state into a real number depending on the models parameters. This function $g$ not only helps us determine whether a change in hidden states occurs but also the magnitude of the change. For example, if deviations from known standards or desired values $\boldsymbol{\kappa} = \{\kappa_1,...,\kappa_m\}$ of $\boldsymbol{X}$ imply changes in state, the following  $g$ functions described in Eq.~(\ref{eq:label}) and Eq.~(\ref{eq:label2})  can be used to help in hidden states labelling in AR-AsLG-HMMs.
\begin{equation}\label{eq:label}
g_1(i) = \sum_{m=1}^M v_m(\nu_{im}-\kappa_{m})
\end{equation}
\begin{equation}\label{eq:label2}
g_2(i) = \max_{m=1,...,M} \{v_m(\nu_{im}-\kappa_{m})\}
\end{equation}
Where
\begin{equation}\label{eq:nu}
\nu_{im} =  \frac{\beta_{im0}+ \beta_{im1}\nu_{iu_{im1}}+\cdots+\beta_{imk_{im}}\nu_{iu_{imk_{im}}}}{1-(\eta_{im1}+\cdots+\eta_{imp_{im}})}. 
\end{equation}
Observe that in Eq.~(\ref{eq:nu}), the value of $\nu_{im}$ depends on the $\nu$ value of the parents of variable $X_m$ in the context-specific graph related to the $i$ state, so Eq.~(\ref{eq:label}) and Eq.~(\ref{eq:label2}) must be calculated recursively. The recursion begins with those variables that fulfil the following condition: $\textbf{Pa}_i(X_m) = \emptyset$. Next, the recursion is computed for their descendants in the context-specific graph, until no variables are left. In general, $\nu_{im}$ can be interpreted as the mean of variable $X_m$ at the hidden state $i$. Additionally, the vector $\textbf{v}=(v_1,...,v_M)$ for $m=1,..,M$ can be considered a feature relevance constant vector or a scaling constant vector that can be tuned according to the nature of the problem. 

Eq.~(\ref{eq:label}) can be used in cases where the addition of errors determines the driven process. For example, in the case of a country economy where the aggregation of economic variables can determine if there is economic growth or not. Or in the case of bearings degradation, where the aggregation of the amplitude of desired frequencies represents the presence of failure. On the other hand, Eq.~(\ref{eq:label2}) can be used when high deviations from a single variable is enough to determine the dynamical process. For example, consider a patient with a chronic disease with many sensors that measure different biological variables. For each variable there is a desirable value that determine good health. If only one variable drifts from the desirable value, the health of the patient can be in danger. In conclusion, the experiment and context of the problem may require a different $g$ function to describe the hidden states.

\section{Experiments} 

In this section, we will compare our model (AR-AsLG-HMMs) with  AsLG-HMMs,  LMSAR, AR-MoG-HMMs, MoG-HMMs, BMMs and a simple AR-AsLG-HMM with Naïve Bayes context-specific Bayesian networks that we will call naïve-HMMs (this kind of models have been used in \cite{Ma08}). In the case of LMSAR \cite{Ha90}\footnote{See page 57}  and AR-MoG-HMM  \cite{Ju85}, it was defined only for one variable. However, in these experiments, we assume that for these models every variable is independent. In particular, LMSAR is a special case of an AR-AsLG-HMM, where only AR parameters are used in the mean, but the number of them do not change with the hidden state and the variance of the model does not depend on the hidden state.  In the case of AR-MoG-HMM, the model assumes that the variances are unitary and do not depend on the hidden state; in spite of that it cannot be expressed directly as an AR-AsLG-HMM. The aim is to show the capabilities of our model to change the number of AR parameters and the context-specific Bayesian networks when they are needed.

Experiments with synthetic data are performed. The data are generated such that over time, the AR process changes. Two dynamic processes are used, one with three variables and another with six. The models are learned using only one time series, where three possible hidden states are present and appear in time blocks. We aim to determine for new data the most likely sequence of the hidden states. The sequence of hidden states tells us the current probabilistic distribution of the data and therefore which probabilistic relationships are relevant. Also, the number of parameters and BIC score play an important role to identify which model is better.

Air quality data and real bearing degradation data are also used. The $p^*$ values are computed using the Yule-Walker equations. These are used as well for  AR-MoG-HMM, BMM  and LMSAR to determine the maximum lag during the learning task. For the mixture models, one, two and three mixture components were used, and the models with two mixture components had the highest BIC and log-likelihood. Then, just two mixture components were used.

For both synthetic and real data, the models are initialized with a uniform transition matrix $\textbf{A}$, specifically $a_{ij} = 1/N$; for $i,j=1,...,N$; the same for the initial  distribution $\boldsymbol{\pi}$, specifically, $\pi_i= 1/N$ for $i=1,...,N$. In the case of our AR-AsLG-HMM, we evaluate the partial correlation function up to five AR values to prevent high computational times and set $p_{im} = 0$ for $i=1,...,N$ and $m=1,...,M$; this means that no AR relationships are assumed a priori in the models. For both AR-AsLG-HMM and AsLG-HMM, all the context-specific Bayesian networks are initialized as naïve Bayes networks. The emission probabilities for the AR-AsLG-HMM and AsLG-HMM are initialized with $\beta_{im0} = i(\max_t{x_m^t}-\min_t{x_m^t})/(N+1)+\min_t{x_m^t}$ and $\sigma_{im}^2 = 2(\max_t{x^t_m}-\min_t{x_m^t})$ for $i=1,..N$ and $m=1,...,M$. The purpose of this selection for $\beta_{im0}$ is to initialize the mean of each variable for each hidden state in an equally separated different point in the possible range of values given by the training dataset. The selection of  $\sigma_{im}^2$, is to avoid infinite or nan values in the forward-backward algorithm. For the AR-MoG-HMM, the AR coefficients are initialized as zero, the distribution of the mixture coefficients is uniform, the mean coefficients for the mixtures models are initialized using a k-means algorithm and the covariance matrices where initialized with the covariance matrix coming from the data.

\subsection{Synthetic data}

We consider two scenarios with three known hidden states which follow AR-AsLG emission probabilities. We generate blocks of data for each hidden state. We mix these blocks with no particular order to create a signal and train every model with this unique signal. We try to simulate data as in real life applications where hidden states do not have a particular order of appearance i.e., any possible transition between hidden states is possible. We will evaluate the learned models with four different types of sequences of hidden states. Each one of these four sequences are generated fifty times to be evaluated in the testing phase. From the fifty evaluations we report the  mean BIC, mean log-likelihood (LL), the standard deviation of the log-likelihoods (Std) and the number of parameters of the model ($\#$).

\begin{figure}
	\begin{center}
		\begin{tabular}{cc}
		\includegraphics[scale=0.2]{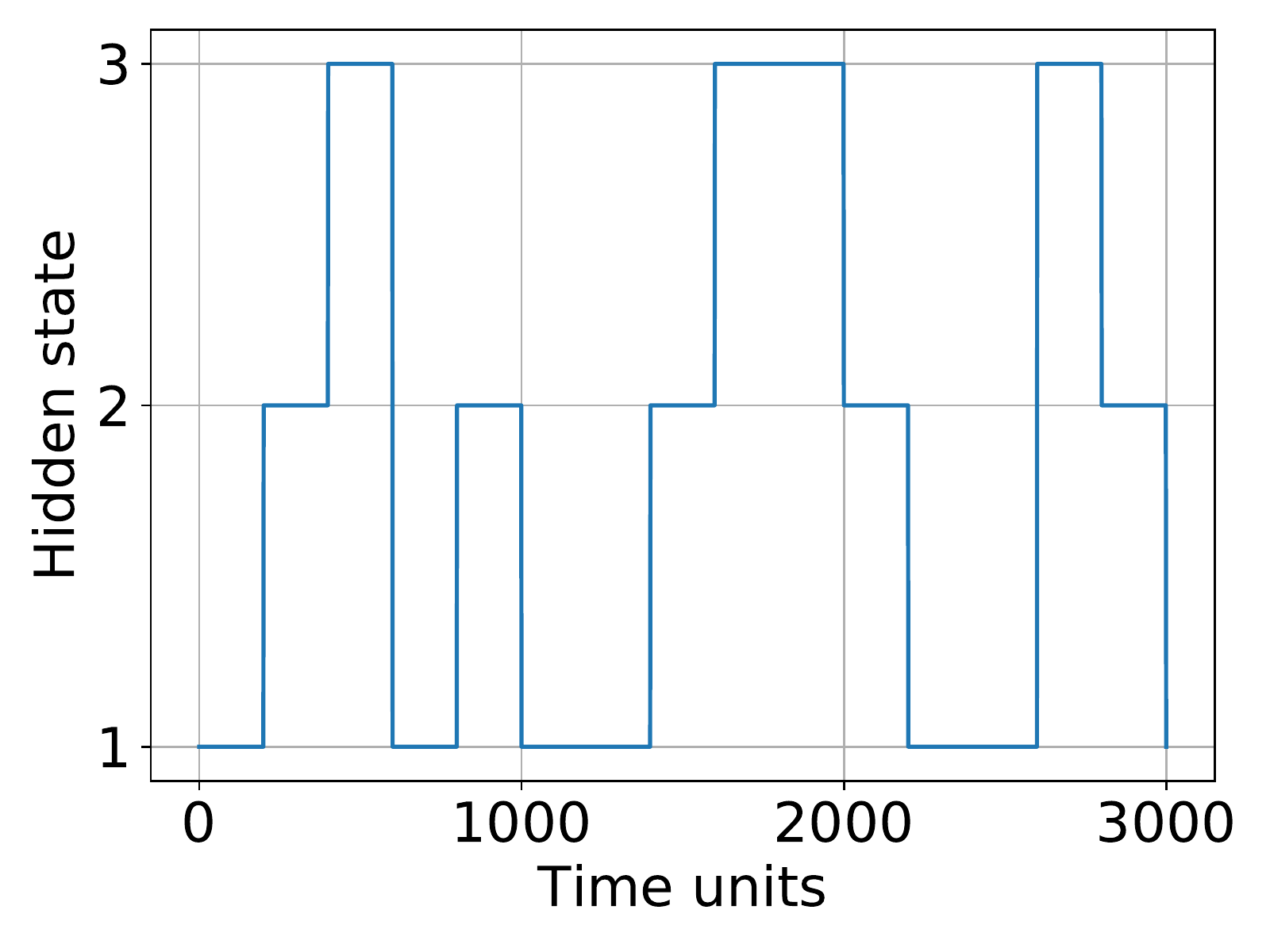}  &
		 \includegraphics[scale=0.2]{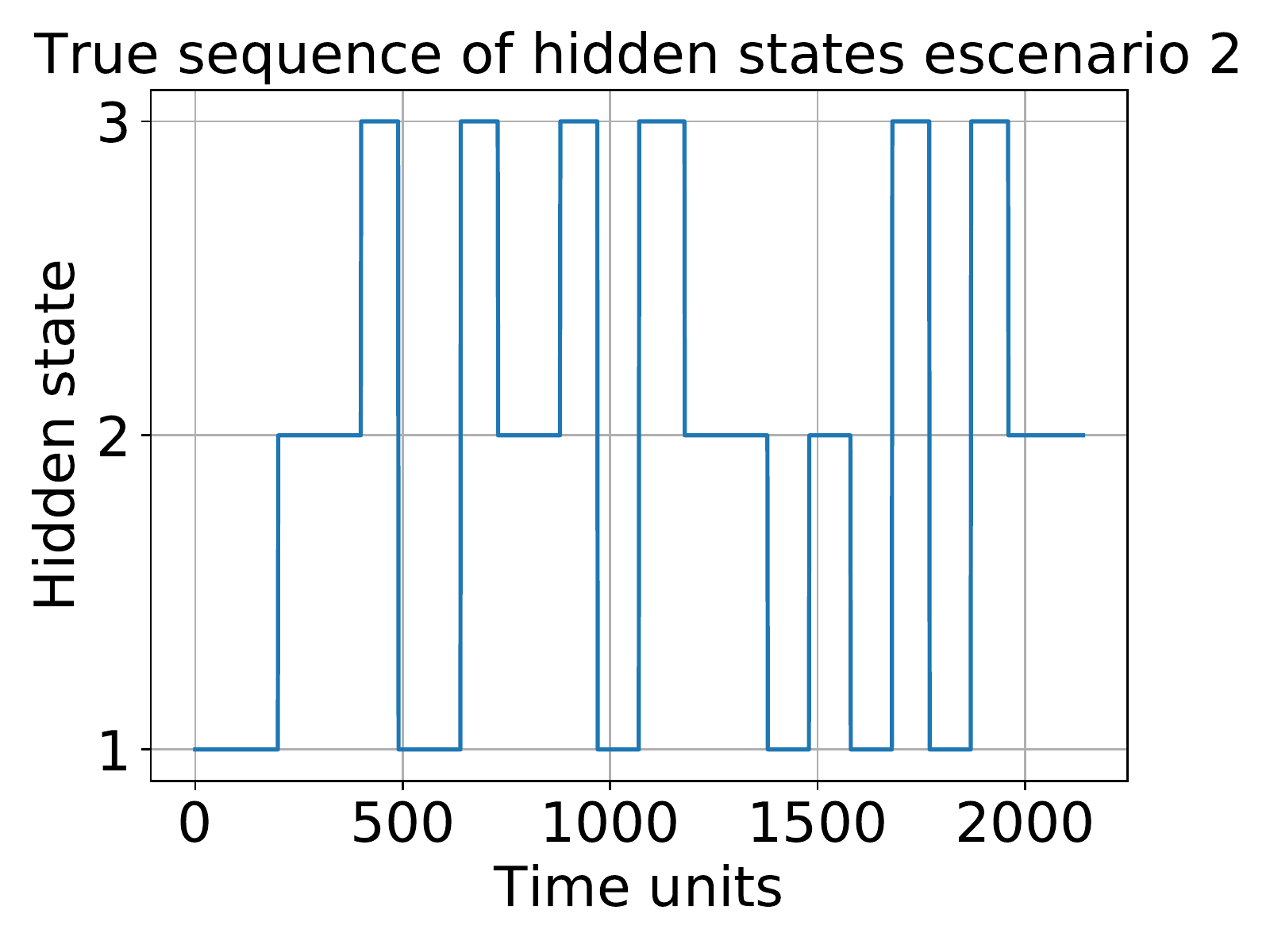} \\
		 (a) &(b) 
		\end{tabular}
	\end{center}
	\caption{Sequences of hidden states used to construct the training signals for scenario 1 (a) and  for scenario 2 (b)}
	\label{fig:sequencetrain}	
\end{figure}

\begin{figure}[h]
 	
 	\begin{center}
 		\begin{tabular}{cccc}
 			\includegraphics[scale=0.2]{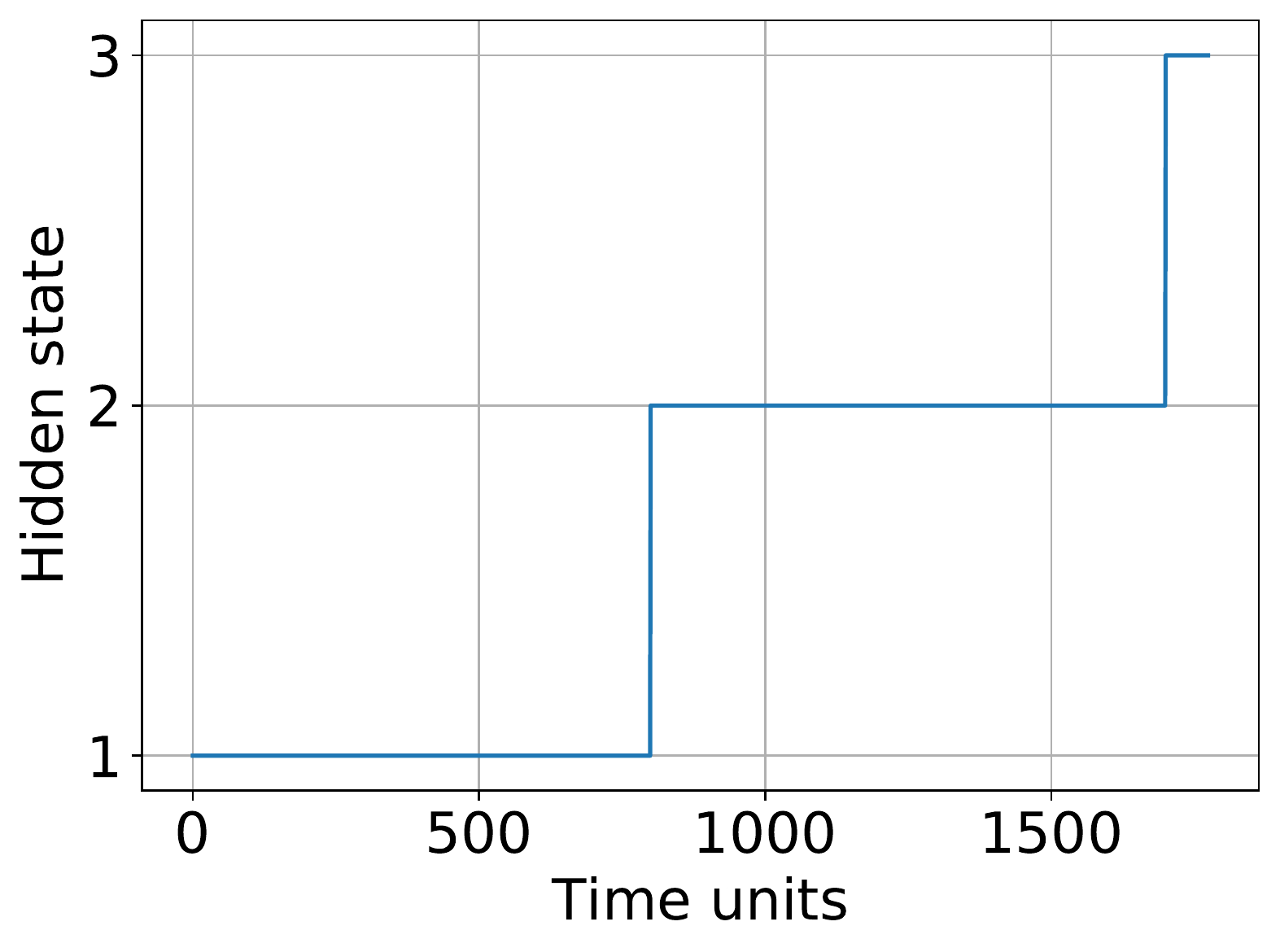} &\includegraphics[scale=0.2]{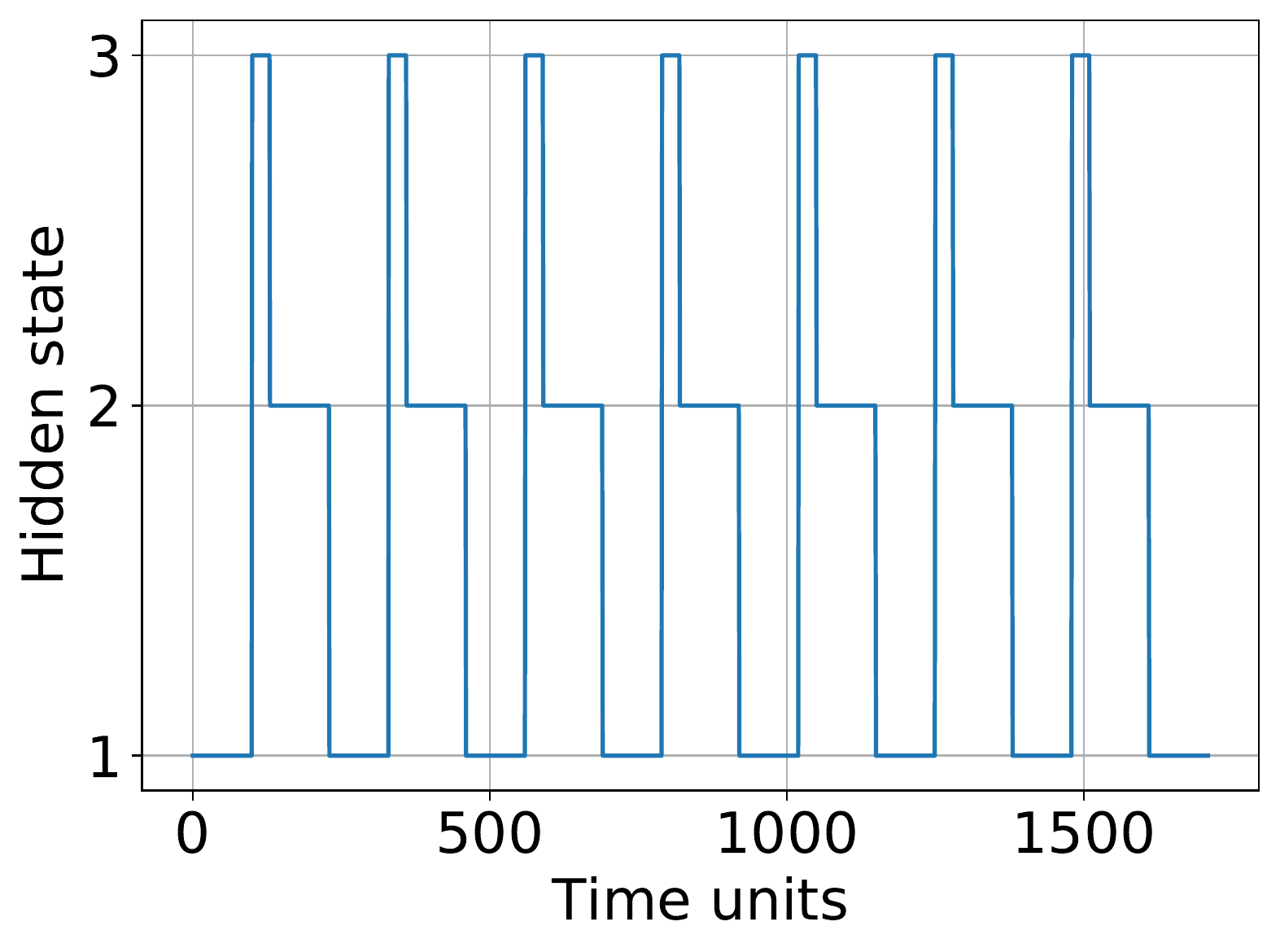} &
 			\includegraphics[scale=0.2]{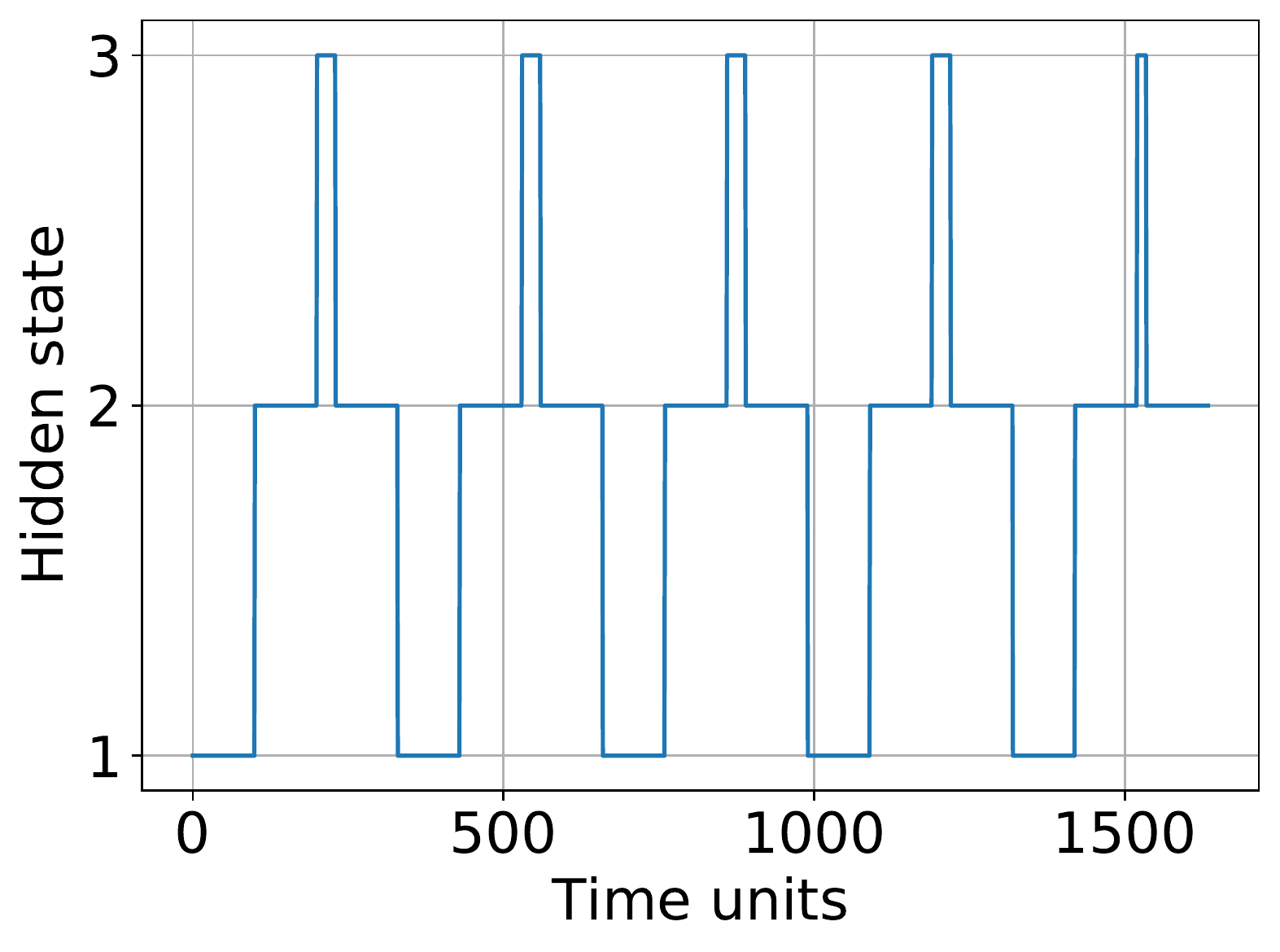}  &\includegraphics[scale=0.2]{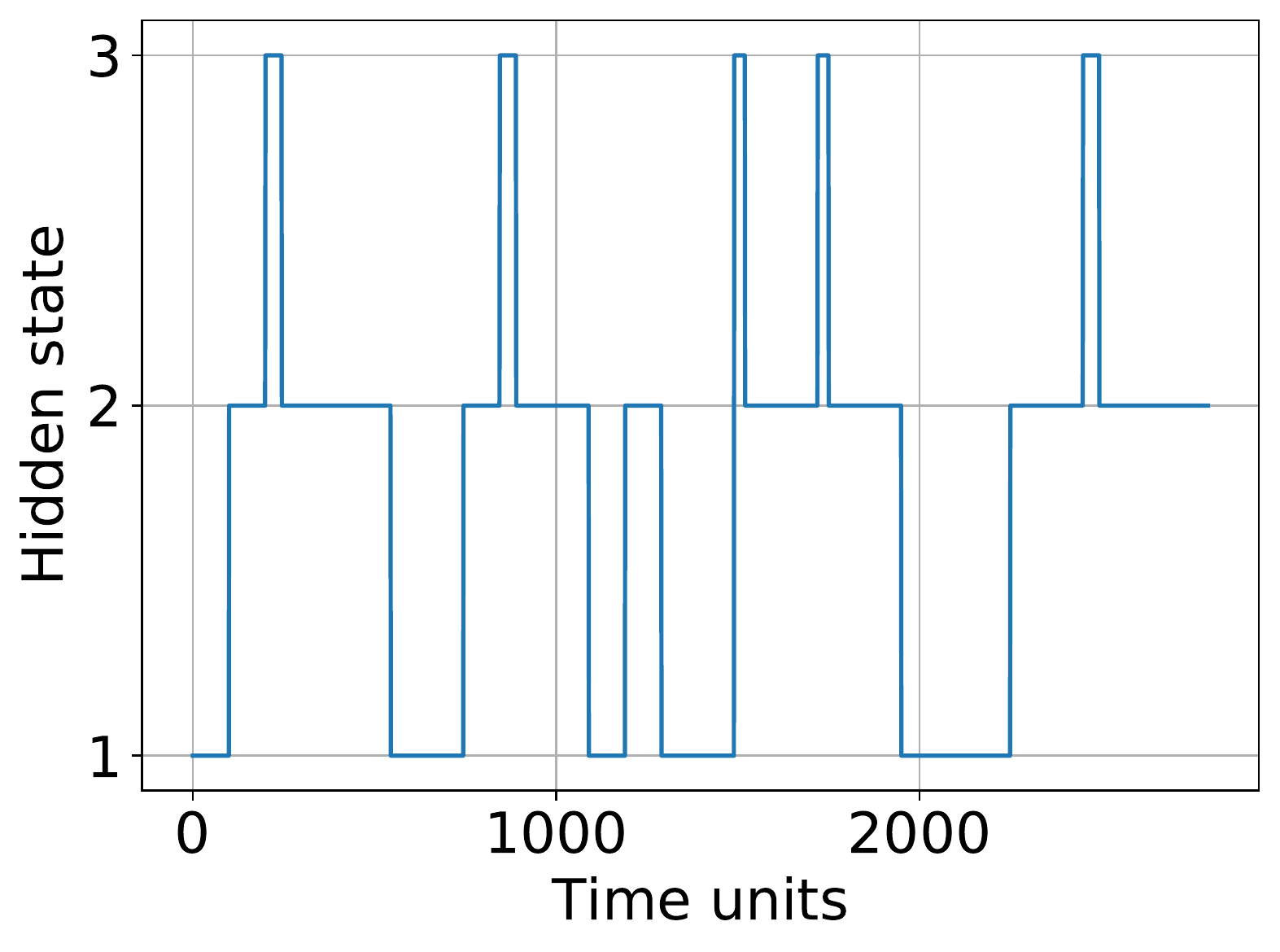} \\
 			(a) Sequence 1 & (b) Sequence 2 & (c) Sequence 3 &(d) Sequence 4 \\
 		\end{tabular}
 	\end{center}
 	\caption{Sequences of hidden states used to construct  the test signals (a) sequence 1, (b) sequence 2, (c) sequence 3, (d) sequence 4, are used for both scenarios}
 	\label{fig:sequencetest}	
 \end{figure}

The first scenario uses three variables and the second scenario uses six variables.  We aim to compare results in the performance of the models when different number of variables are used. From the parameters, in both scenarios we have a hidden state with no structural complexity (no-AR and no-parent relationships in $f_{im}^t$), a second one with some structural complexity and the last one with a complex structure (many AR and parent relationships in $f_{im}^t$)). We also edit the parameters in such a way that the more complex  the context-specific Bayesian networks is, the greater amplitudes for the $g_1$ (Eq.~(\ref{eq:label})) function are. The parameters used for the hidden states can be found in the appendix. The $g_1$ function used for all the experiments has $v_m=1$ for $m=1,...,M$ and $\kappa_{m}=0$, for $m=1,...,M$. The sequence of hidden states used to construct the training signal for both scenarios  can be seen in Fig.~\ref{fig:sequencetrain}. The four sequences of hidden states for testing are illustrated in Fig.~\ref{fig:sequencetest}.

\begin{table}[H]
	\centering
	\begin{tabular}{llrrrr}
		  &Model &  mean LL & mean BIC & Std & $\#$ \\
		\hline
1&AR-AsLG-HMM        &-11481.91 &23225.69 &59.36 &35\\
&AsLG-HMM    		 &-11692.59 &23609.64 &59.09 &\textbf{30}\\
&LMSAR 				 &-13610.09 &27669.10 &90.64 &60\\
&AR-MoG-HMM 		 &\textbf{-11028.45 }&\textbf{22820.08} &44.02 &102\\
&MoG-HMM 			 &-11675.42 &23844.67 &60.55 &66\\
&Naïve-HMM   		 &-11692.59 &23609.64 &59.09 &\textbf{30}\\
&BMM   		         &-19959.85 &40413.51 &\textbf{6.16} &66\\
		\hline
2&AR-AsLG-HMM        &-10766.49 &21793.54 &48.58 &35\\
&AsLG-HMM    		 &-11696.91 &23617.17 &60.48 &\textbf{30}\\
&LMSAR 				 &-13475.42 &27397.52 &84.66 &60\\
&AR-MoG-HMM 		 &\textbf{-10495.98} &\textbf{21751.33} &30.31 &102\\
&MoG-HMM 			 &-11508.64 &23508.65 &56.1 &66\\
&Naïve-HMM   		 &-11696.92 &23617.19 &60.48 &\textbf{30}\\
&BMM   		         &-19215.87 &38923.10 &\textbf{4.60} &66\\
		\hline
3&AR-AsLG-HMM        &-11487.53 &23234.06 &55.74 &35\\
&AsLG-HMM    		 &-12140.67 &24503.33 &62.52 &\textbf{30}\\
&LMSAR 				 &-13540.81 &27525.63 &91.08 &60\\
&AR-MoG-HMM 		 &\textbf{-10525.10} &\textbf{21805.01} &44.8 &102\\
&MoG-HMM 			 &-11999.15 &24486.71 &59.84 &66\\
&Naïve-HMM   		 &-12140.67 &24503.35 &62.52 &\textbf{30}\\
&BMM   		         &-18416.46 &37321.32 &\textbf{6.27} &66\\
		\hline
4&AR-AsLG-HMM        &-19492.59 &39262.94 &62.40 &35\\
&AsLG-HMM    		 &-20247.84 &40733.75 &80.23 &30\\
&LMSAR 				 &-22843.96 &46164.08 &115.15 &60\\
&AR-MoG-HMM 		 &\textbf{-17929.36} &\textbf{36668.19} &58.0 &102\\
&MoG-HMM 			 &-20105.77 &40735.30 &78.30 &66\\
&Naïve-HMM   		 &-20247.85 &40733.77 &80.23 &\textbf{30}\\
&BMM   		         &-31504.97 &63533.71 &\textbf{8.73} &66\\
\end{tabular}
\caption{Scores for each testing sequence of Scenario 1}
\label{table:Scenario 1}
\end{table}

In Table~\ref{table:Scenario 1}, we observe the results for scenario 1 for all the sequences. We observe that AR-AsLG-HMM obtained the second best results in LL and BIC score followed by AR-MoG-HMM. The  MoG-HMM, AsLG-HMM, LMSAR and naïve-HMM obtained fair results. However, BMM obtained the worst results for all the  evaluated sequences. In the case of BIC score, the penalization for mixture models was higher since a greater number of parameters were needed. It is remarkable that MoG-HMM in sequences 1 and 4 obtained better LL than AsLG-HMM, however due to the penalization AsLG-HMM obtained slightly better mean BIC scores. In terms of standard deviation, the BMM obtained the best results in spite of having the worst LL  and BIC scores. AR-AsLG-HMM was the third model with best Std score after BMM and AR-MoG-HMM. LMSAR in this case obtained the worst results in terms of standard deviation. In terms of number of parameters, the naïve-HMM had the best results since it has the most simple structure. In general, asymmetric models used fewer parameters than mixture models which used up to three times the number of parameters of the Naïve-HMM.

\begin{table}[H]
	\centering
	\begin{tabular}{llrrrr}
		&Model &  mean LL & mean BIC & Std & $\#$ \\
		\hline
1&AR-AsLG-HMM        &\textbf{-25882.99 }&\textbf{52244.88} &82.33 &64\\
&AsLG-HMM    		 &-32831.37 &66074.28 &214.78 &55\\
&LMSAR 				 &-34963.39 &70734.92 &70.32 &108\\
&AR-MoG-HMM 		 &-28953.60 &59343.88 &\textbf{27.46} &192\\
&MoG-HMM 			 &-32466.47 &66234.92 &678.72 &174\\
&Naïve-HMM   		 &-33270.22 &66899.60 &220.51 &\textbf{48}\\
&BMM   		         &nan &nan &nan &174\\
		\hline
2&AR-AsLG-HMM        &\textbf{-25426.30} &\textbf{51329.10} &55.27 &64\\
&AsLG-HMM    		 &-31548.50 &63506.50 &86.94 &55\\
&LMSAR 				 &-34961.10 &70726.30 &99.39 &108\\
&AR-MoG-HMM 		 &-28182.75 &57795.02 &\textbf{31.21} &192\\
&MoG-HMM 			 &-39861.48 &81018.46 &1076.60 &174\\
&Naïve-HMM   		 &-31811.46 &63980.3 &63.2 &\textbf{48}\\
&BMM   		         &nan &nan &nan &174\\
		\hline 
3&AR-AsLG-HMM        &\textbf{-24394.30} &\textbf{49262.24} &64.59 &64\\
&AsLG-HMM    		 &-32648.50 &65704.03 &205.68 &55\\
&LMSAR 				 &-33374.33 &67547.94 &100.21 &108\\
&AR-MoG-HMM 		 &-26674.30 &54769.52 &\textbf{22.37} &192\\
&MoG-HMM 			 &-36246.21 &73780.13 &977.36 &174\\
&Naïve-HMM   		 &-32223.11 &64801.46 &116.51 &\textbf{48}\\
&BMM   		         &nan &nan &nan &174\\
		\hline
4&AR-AsLG-HMM        &\textbf{-41481.71} &\textbf{83471.34 }&103.08 &64\\
&AsLG-HMM    		 &-54150.03 &108736.55 &172.95 &55\\
&LMSAR 				 &-56626.35 &114109.82 &135.52 &108\\
&AR-MoG-HMM 		 &-45546.89 &92617.54 &\textbf{32.60} &192\\
&MoG-HMM 			 &-59192.10 &119765.11 &994.71 &174\\
&Naïve-HMM   		 &-54372.90 &109126.75 &146.41 &\textbf{48}\\
&BMM   		         &nan       &nan       &nan    &174\\
\end{tabular}
\caption{Results for each testing sequence of scenario 2}
\label{table:Scenario 2}
\end{table}

Table~\ref{table:Scenario 2} shows the results for scenario 2 for all the sequences. In this case we observe that AR-AsLG-HMM obtained the best results in LL and BIC score. The naïve-HMM, AsLG-HMM, AR-MoG-HMM obtained fair results. The remaining models: MoG-HMM, LMSAR and BMM obtained poor results in LL and BIC score. In paticular, BMM could not iterate its EM algorithm after the graph optimization. In terms of standard deviation AR-MoG-HMM obtained the best results in Std followed by AR-AsLG-HMM, whereas MoG-HMM obtained the worst variance results. In terms of number of parameters, again Naïve-HMM uses the least amount of parameters due to its simple structure and again we found that mixture models use up to three times the number of parameters used by Naïve-HMM.

\begin{figure}[H]
	\centering
	\begin{tabular}{cccc}
		\includegraphics[scale=0.22]{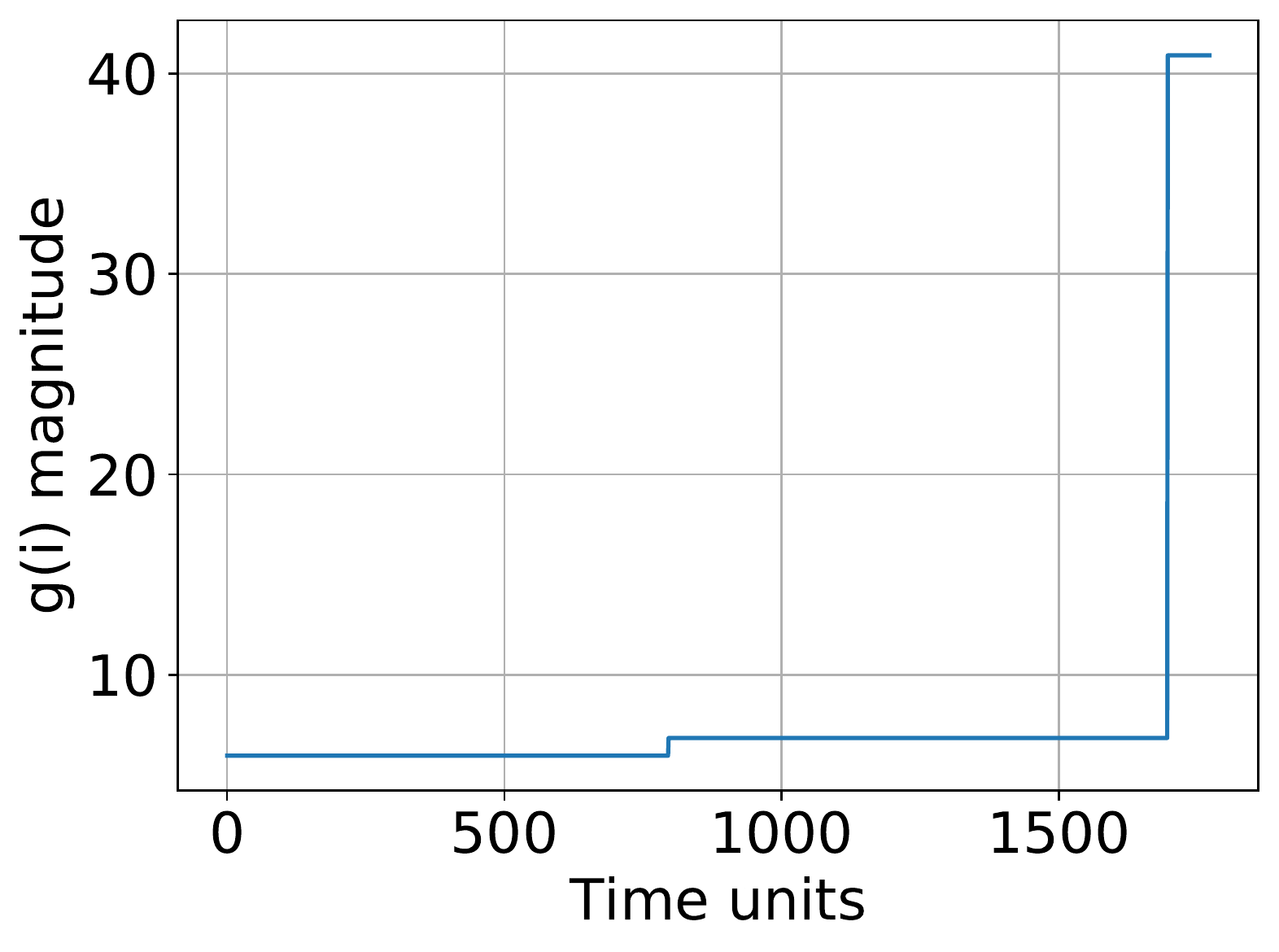} & 		
		\includegraphics[scale=0.22]{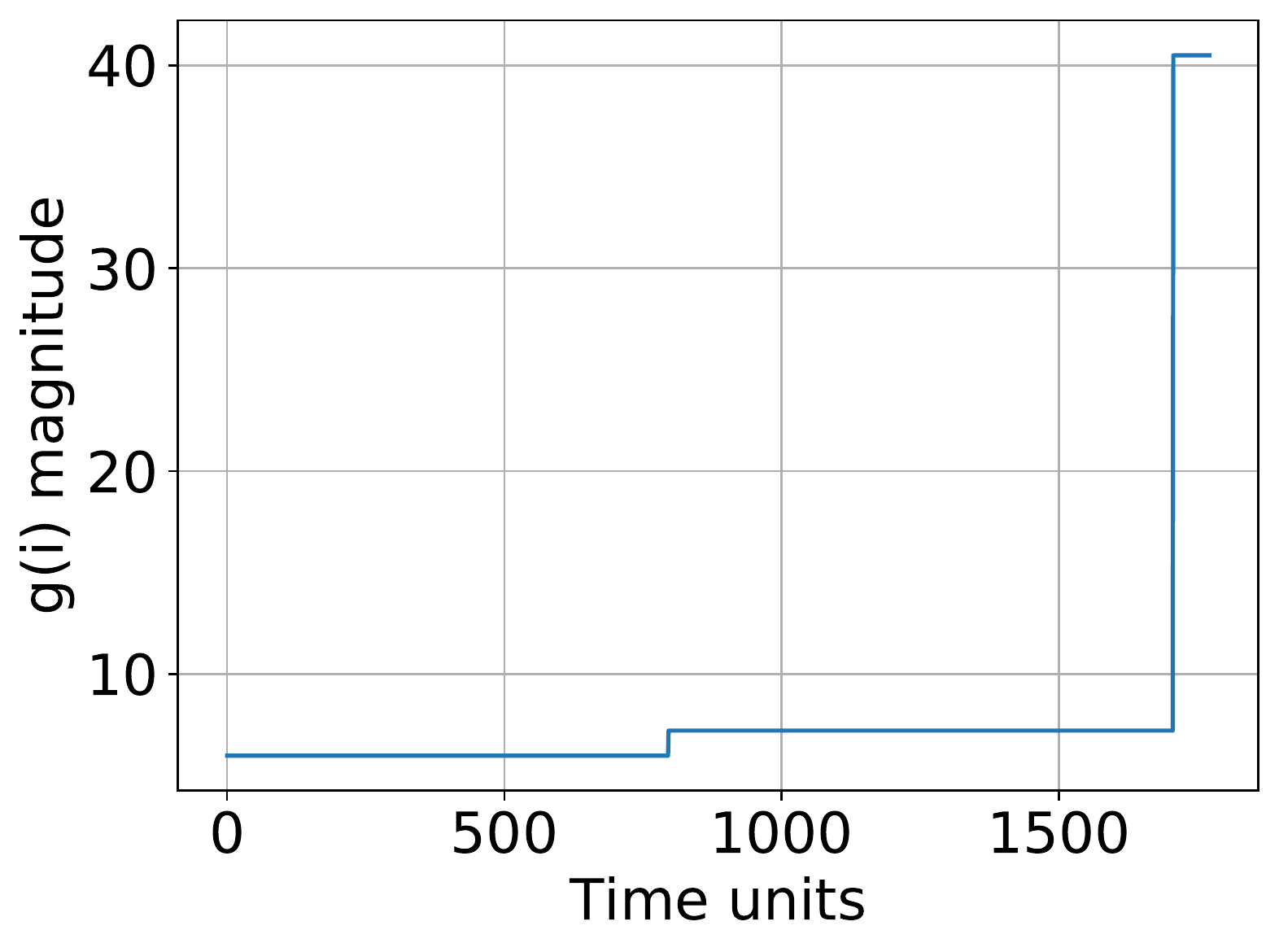}&
		\includegraphics[scale=0.22]{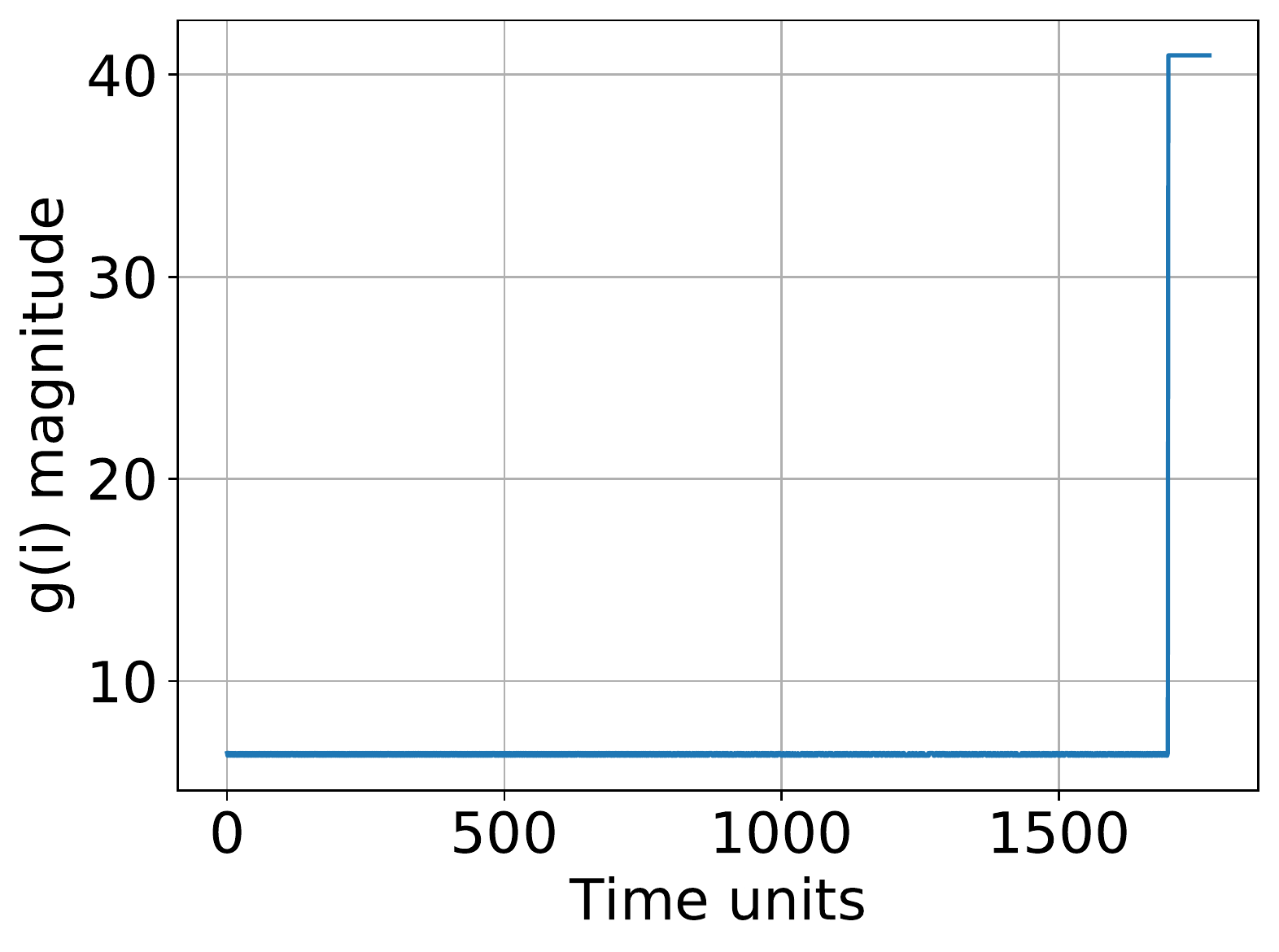}&
		\includegraphics[scale=0.22]{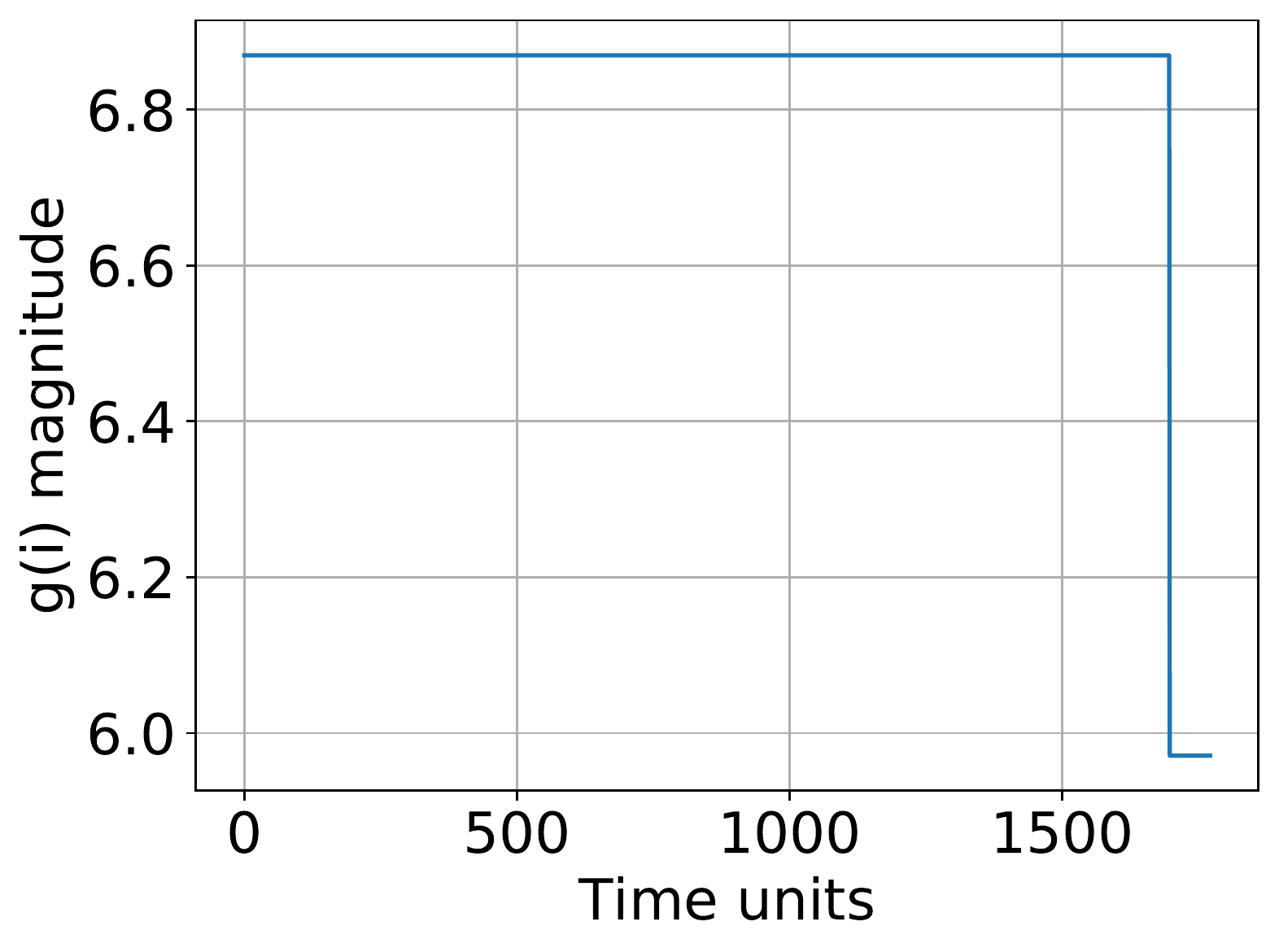}
		\\
		(a) 	AR-AsLG-HMM & (b)  AsLG-HMM&(c) LMSAR &(d) AR-MoG-HMM\\
		\includegraphics[scale=0.22]{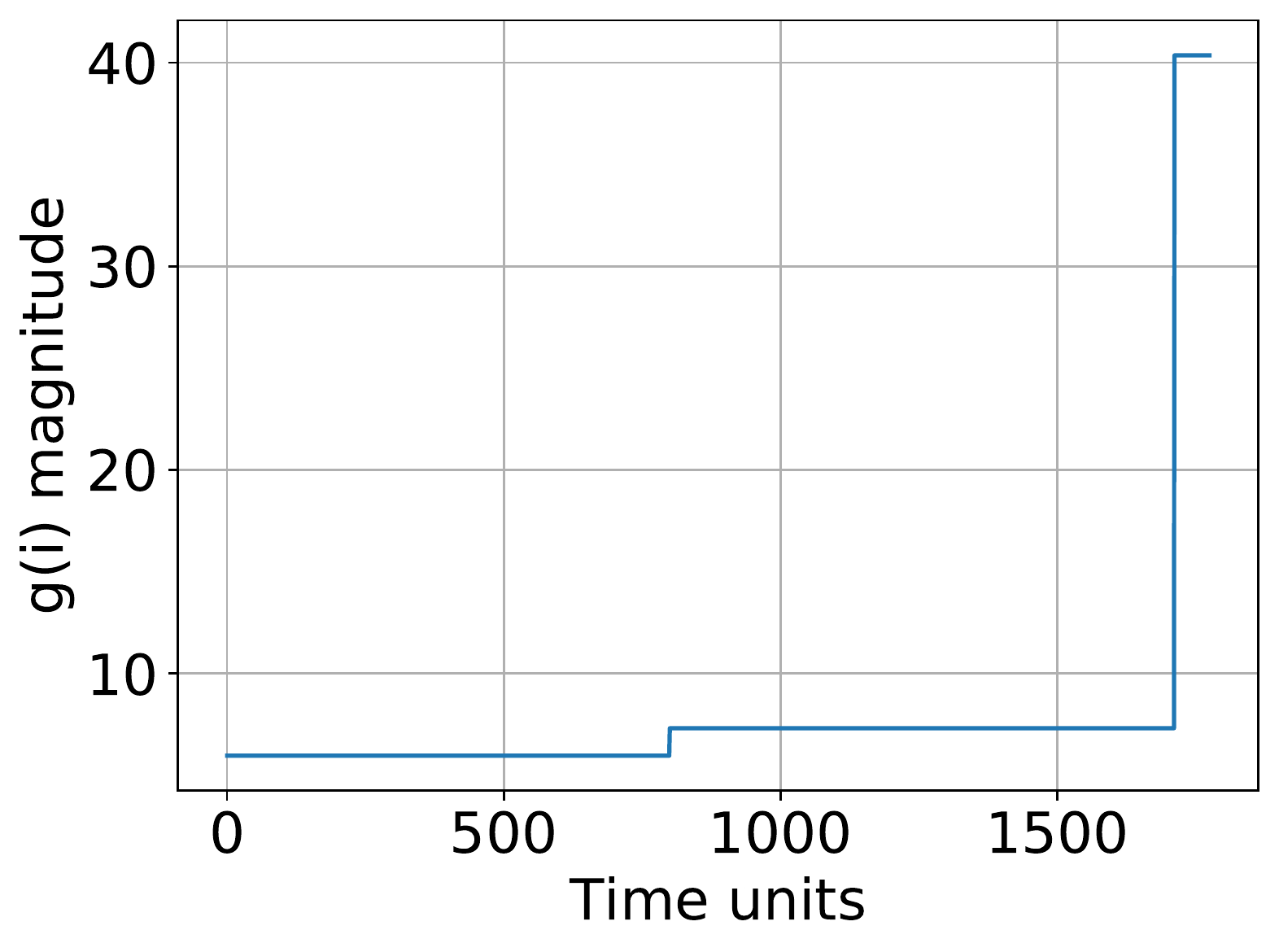}& 
		\includegraphics[scale=0.22]{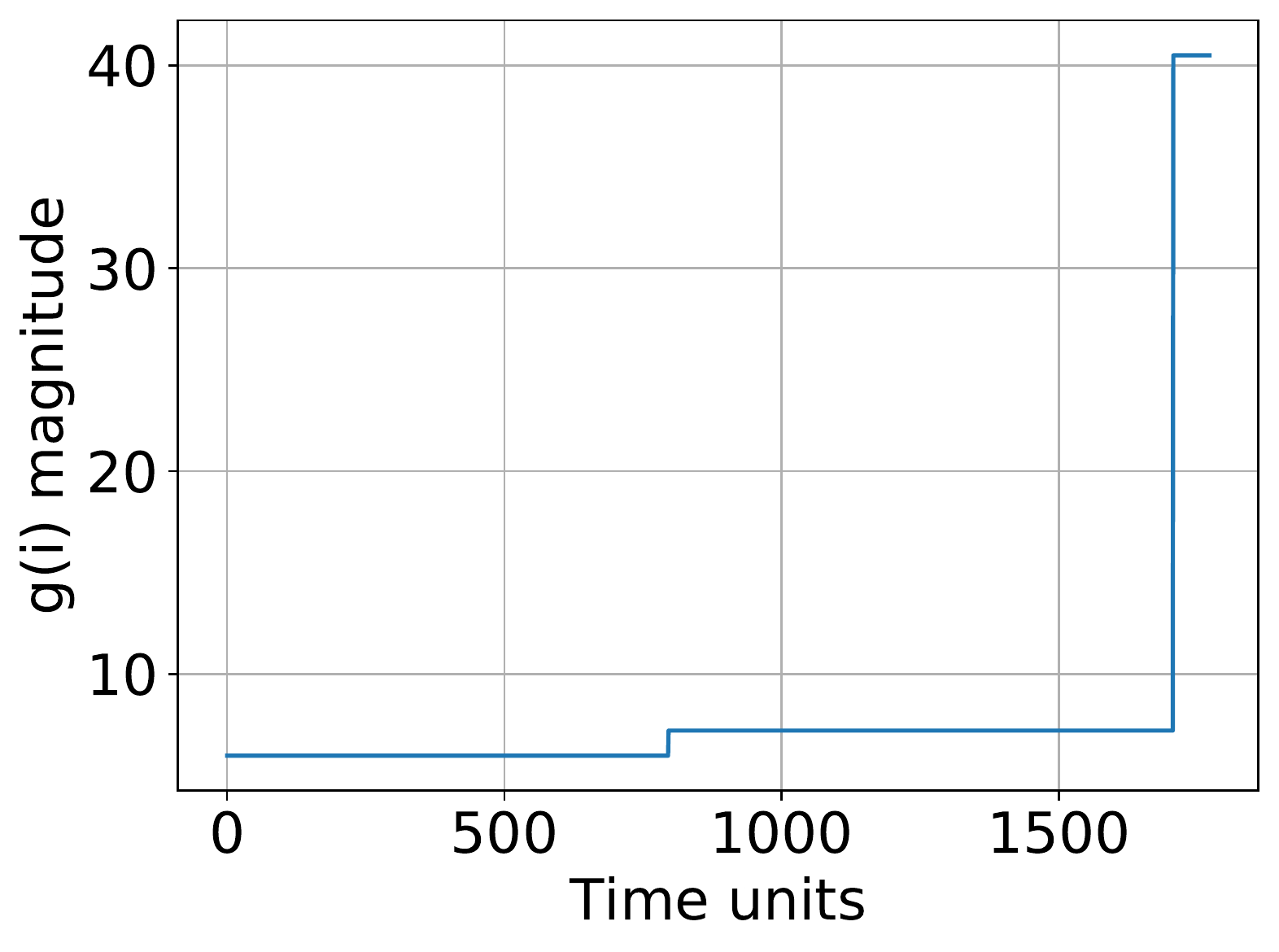}&
		\includegraphics[scale=0.22]{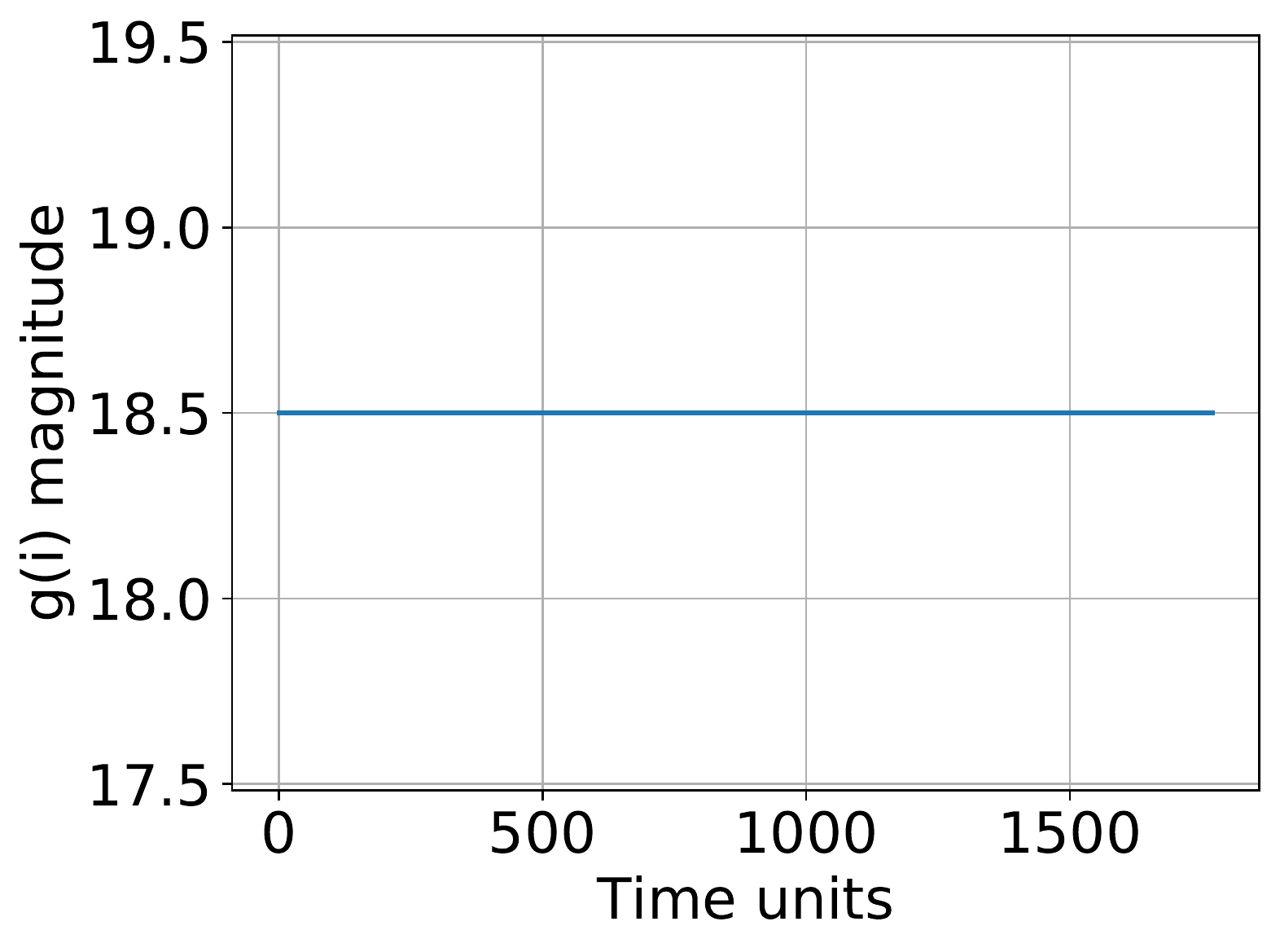}&\\
		(e) MoG-HMM & (f) Naïve-HMM	& (g) BMM & \\
	\end{tabular}
	\caption{Viterbi paths for scenario 1 and sequence 1}
	\label{fig:s1_s1}	
\end{figure}

\begin{figure}[H]
	\centering
	\begin{tabular}{cccc}
		\includegraphics[scale=0.22]{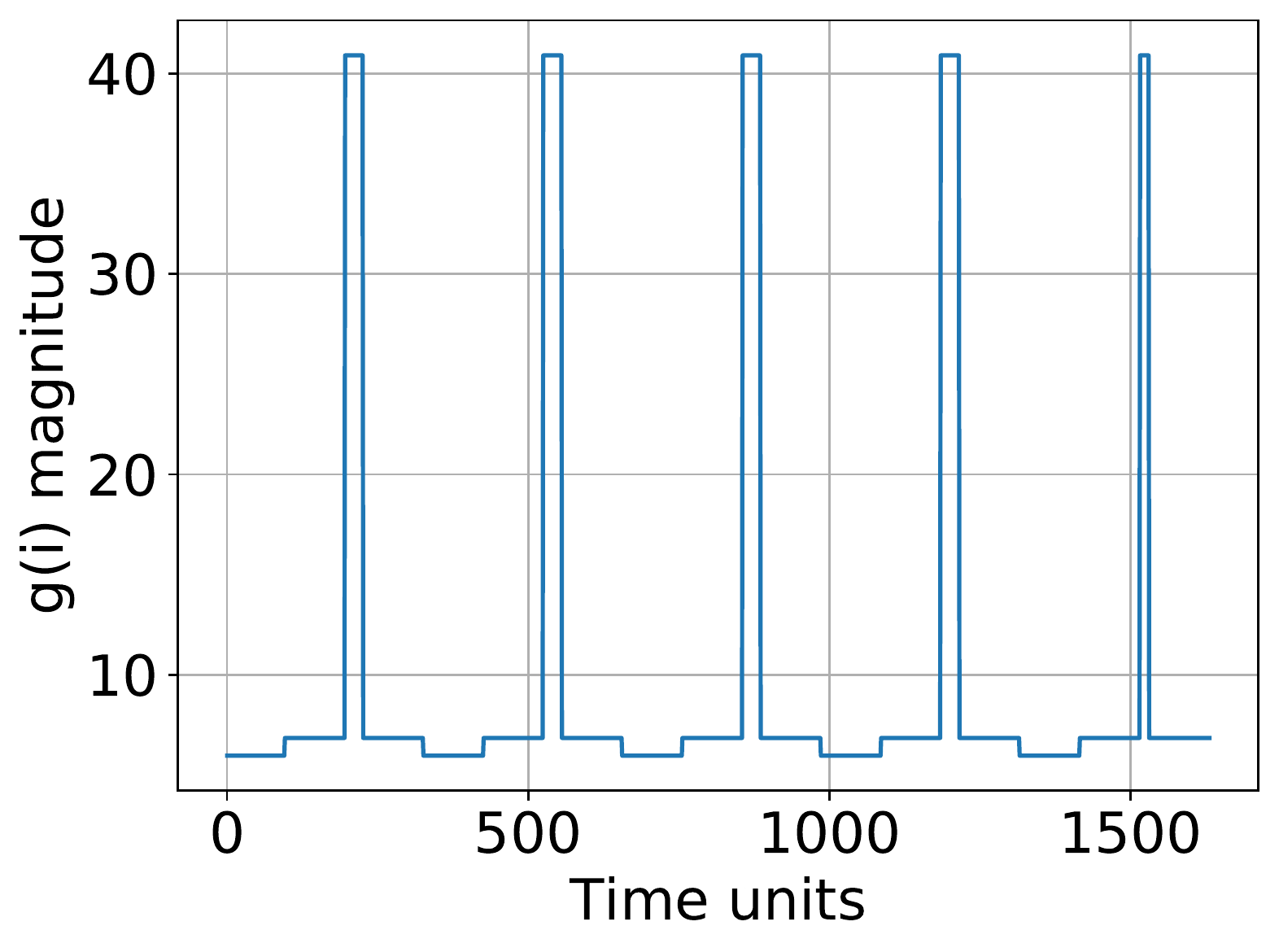} & 		
		\includegraphics[scale=0.22]{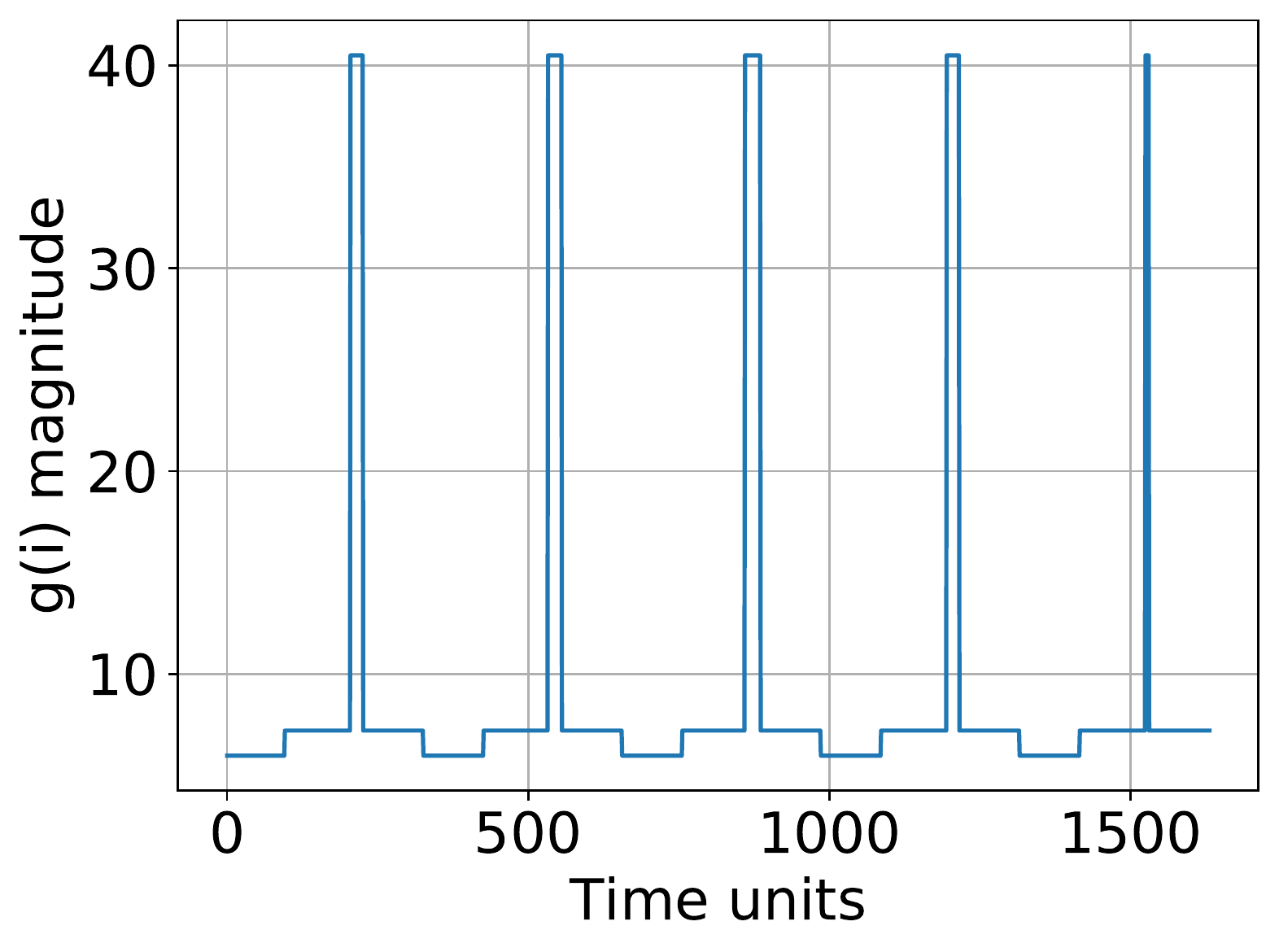} &
		\includegraphics[scale=0.22]{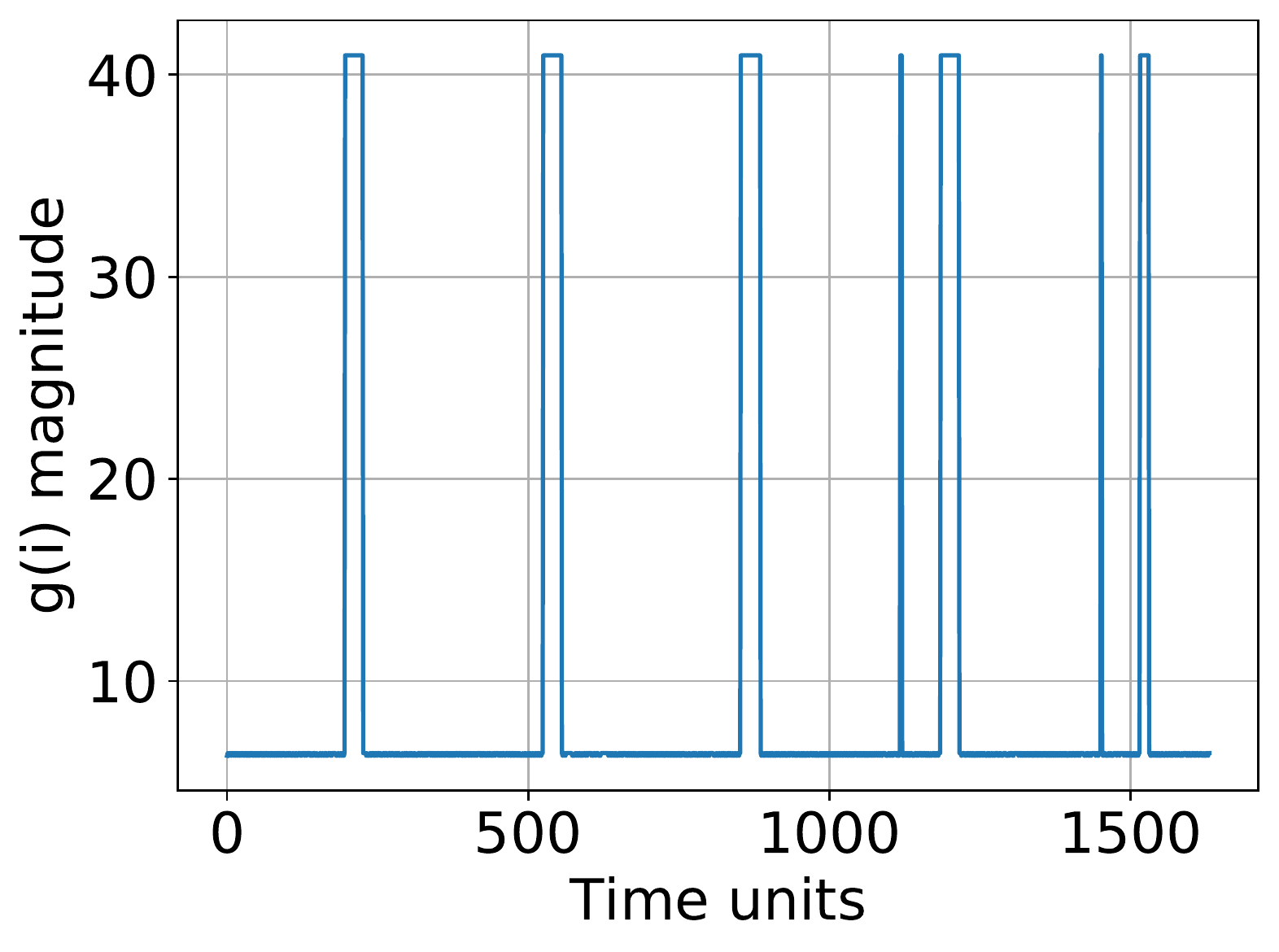}&
		\includegraphics[scale=0.22]{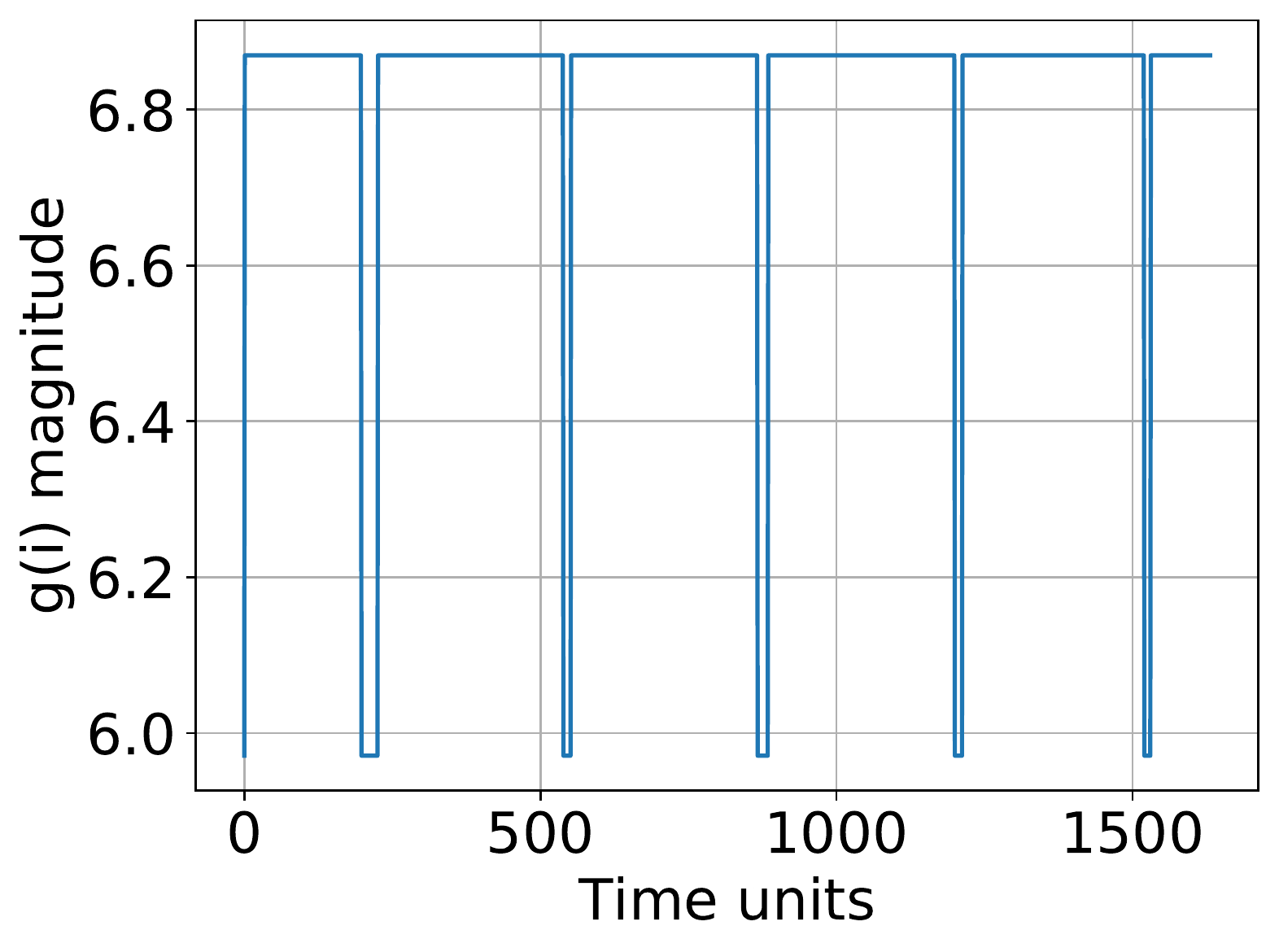}
		\\
		(a) 	AR-AsLG-HMM & (b)  AsLG-HMM &(c) LMSAR & (d) AR-MoG-HMM\\
		\includegraphics[scale=0.22]{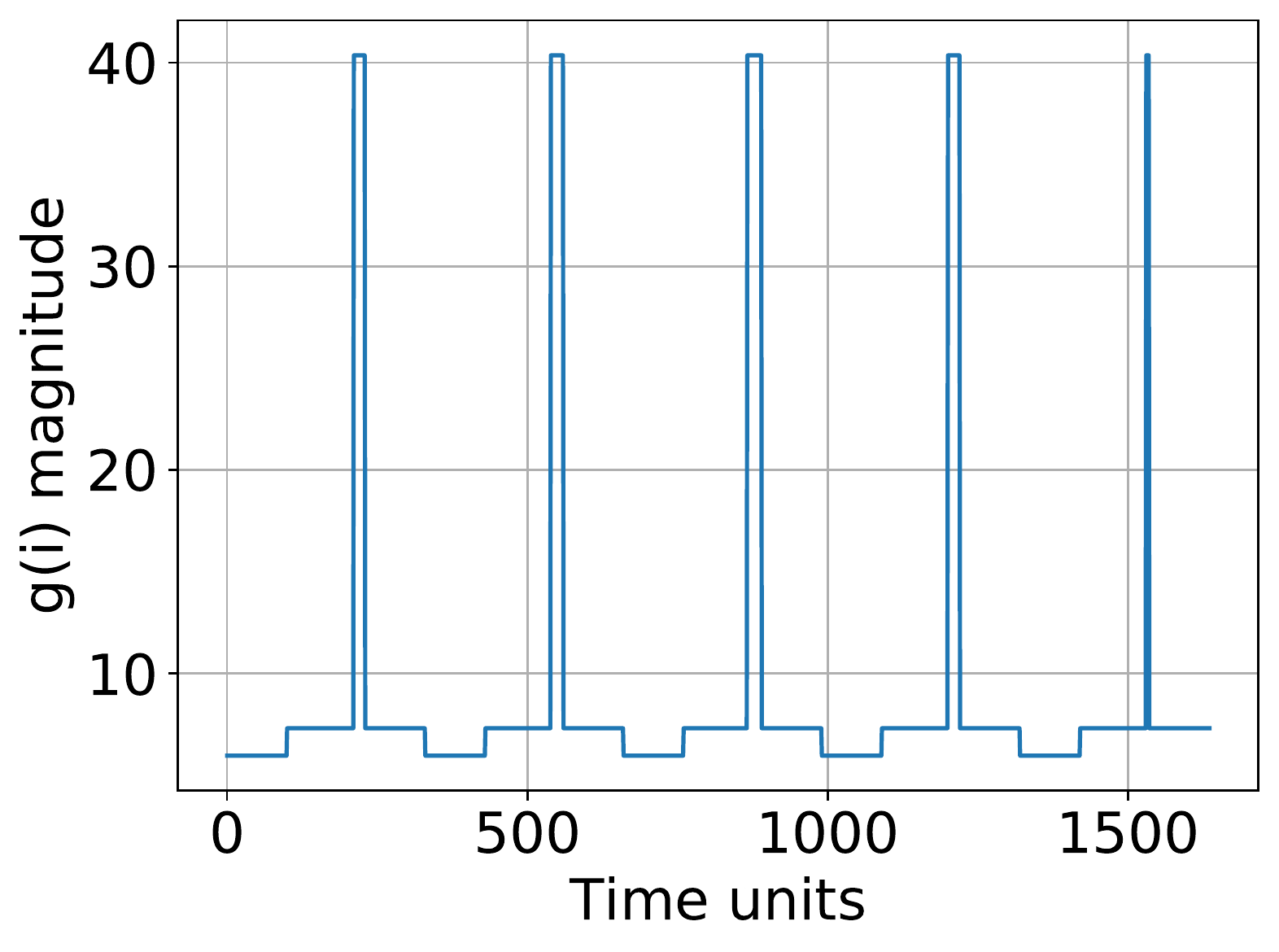}& 
		\includegraphics[scale=0.22]{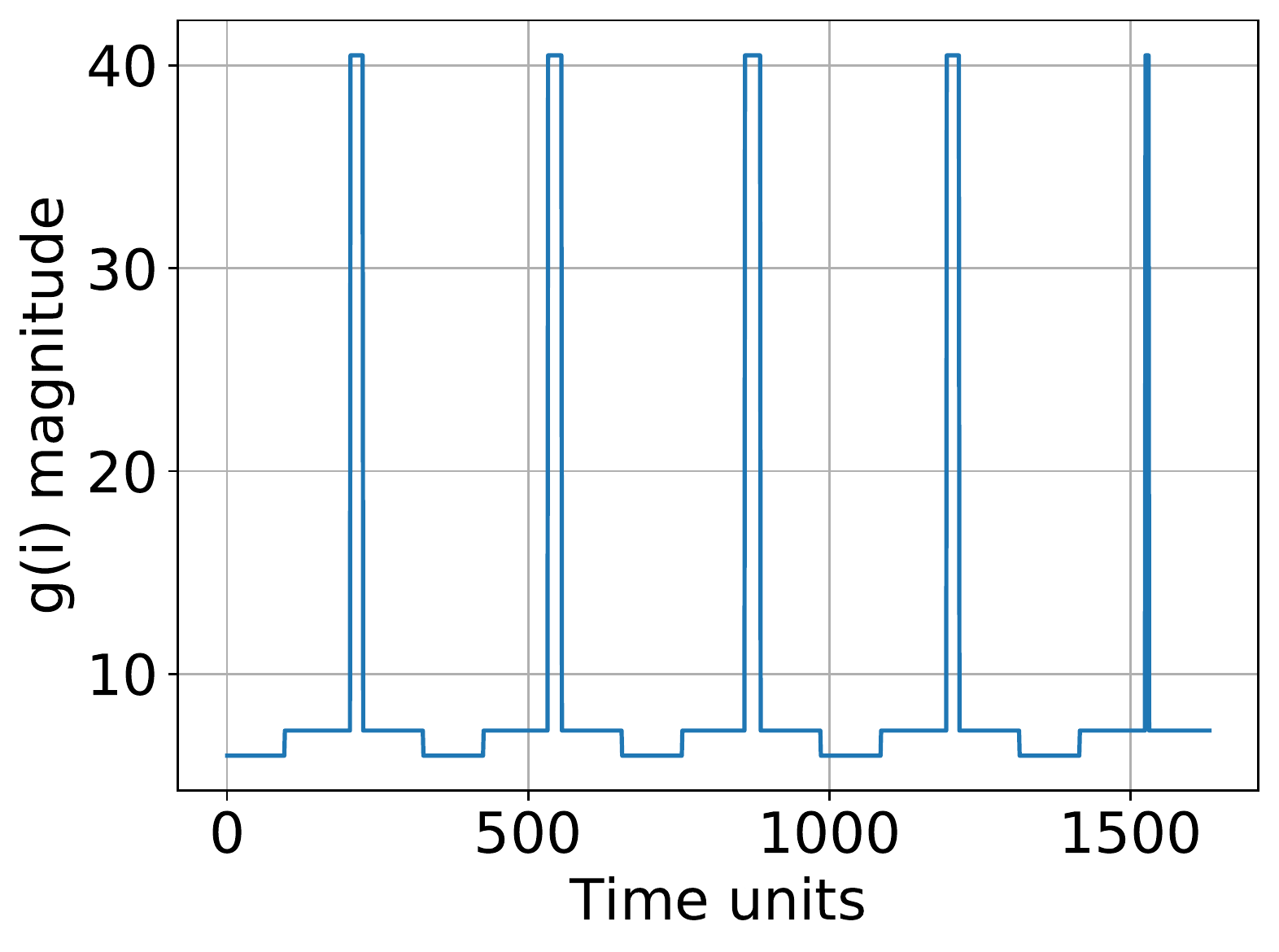}&
		\includegraphics[scale=0.22]{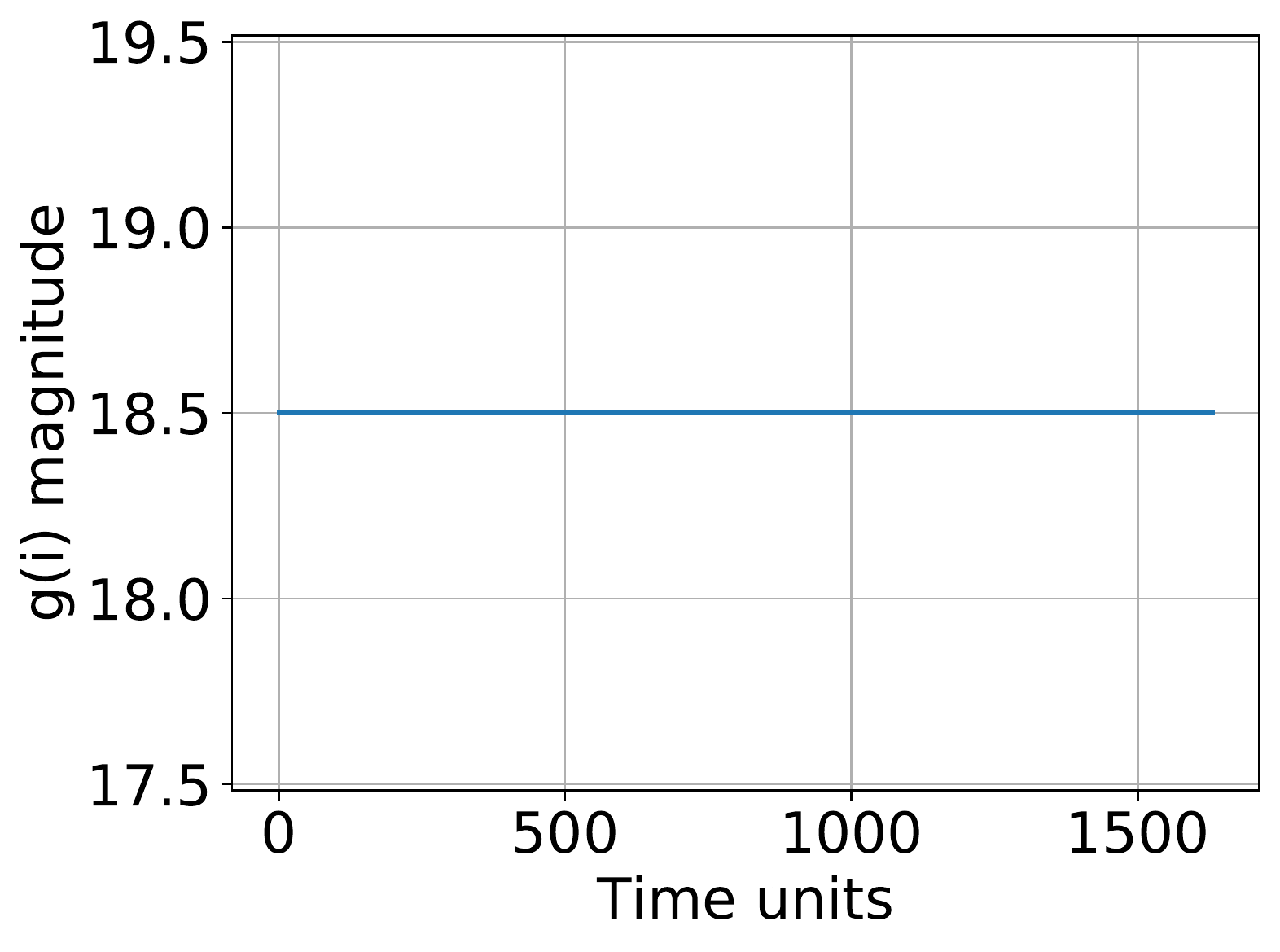}&
		\\
		(e) MoG-HMM & (f) Naïve-HMM& (g) BMM & \\
	\end{tabular}
	\caption{Viterbi paths for scenario 1 and sequence 3}
	\label{fig:s1_s3}	
\end{figure}

\begin{figure}[H]
	\centering
	\begin{tabular}{cccc}
		\includegraphics[scale=0.22]{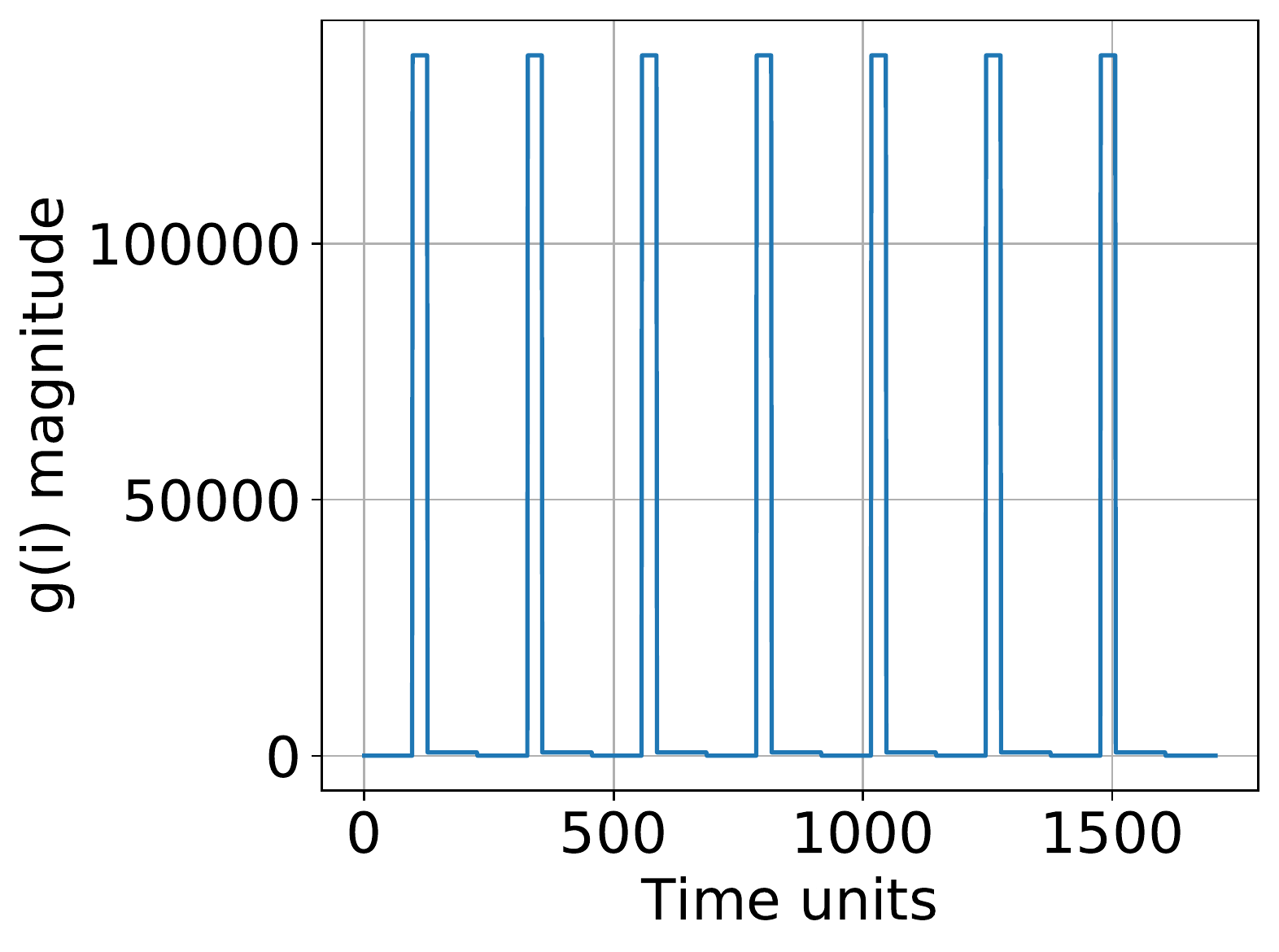} & 		
		\includegraphics[scale=0.22]{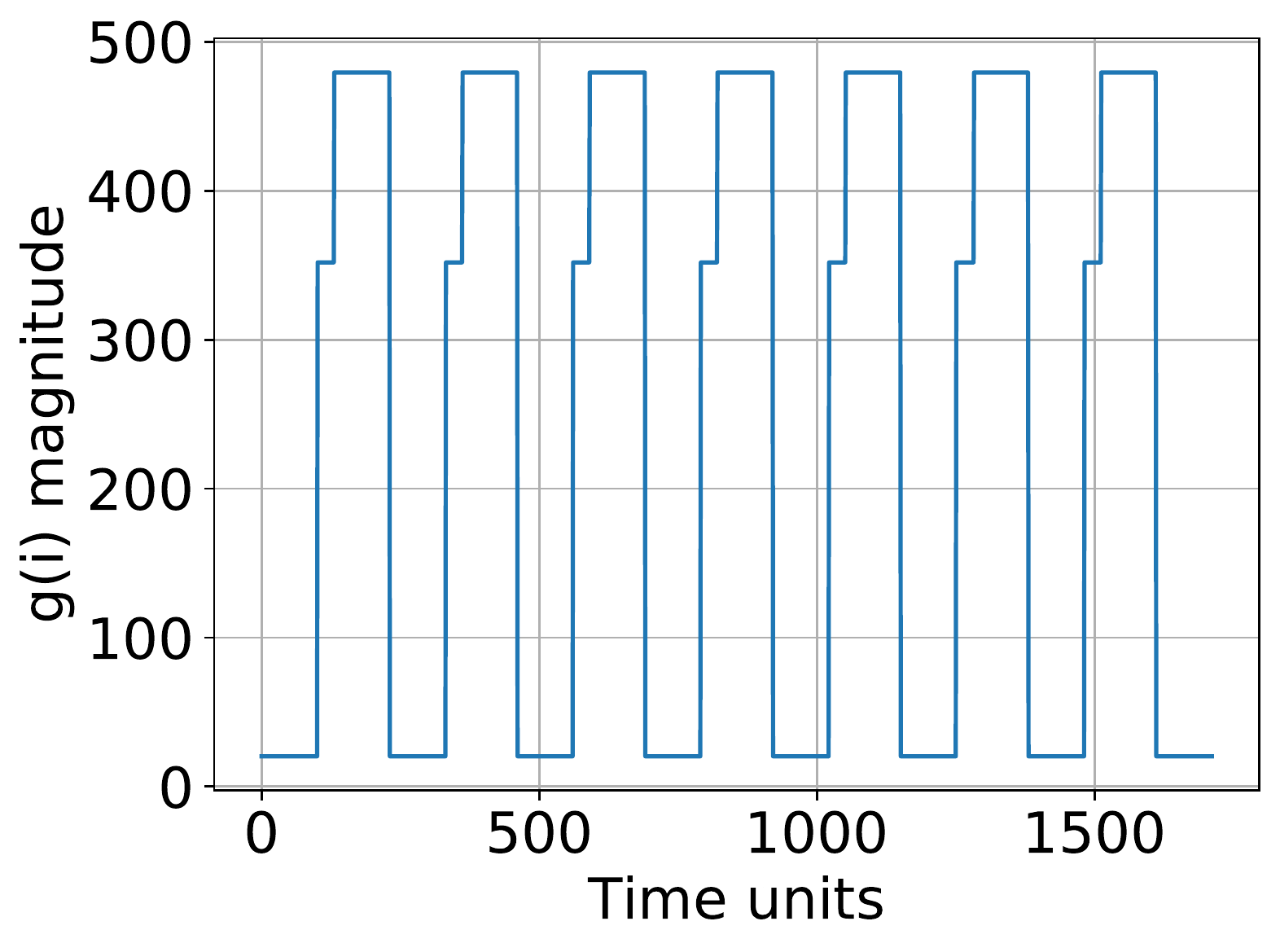}&
		\includegraphics[scale=0.22]{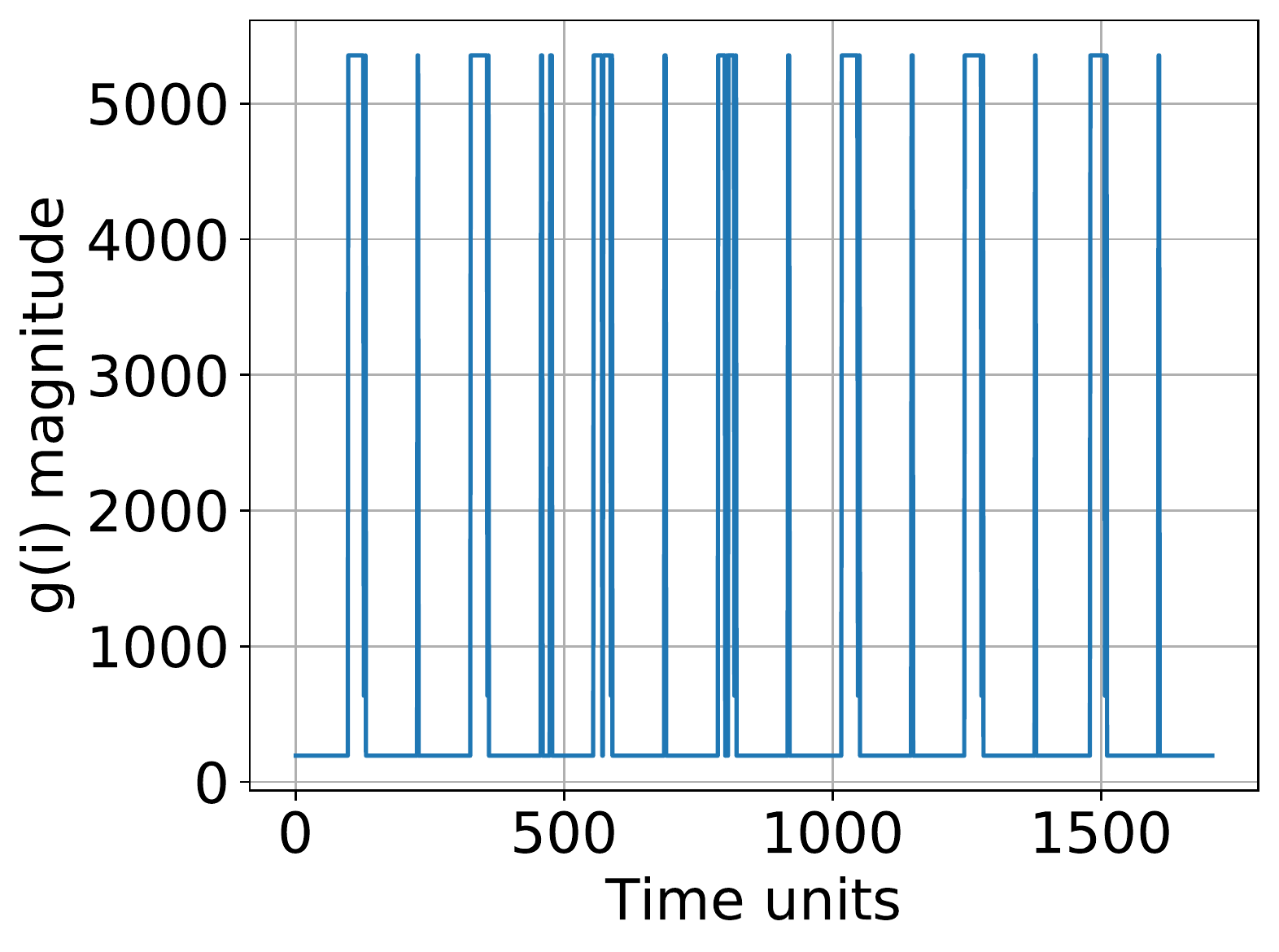}&
		\includegraphics[scale=0.22]{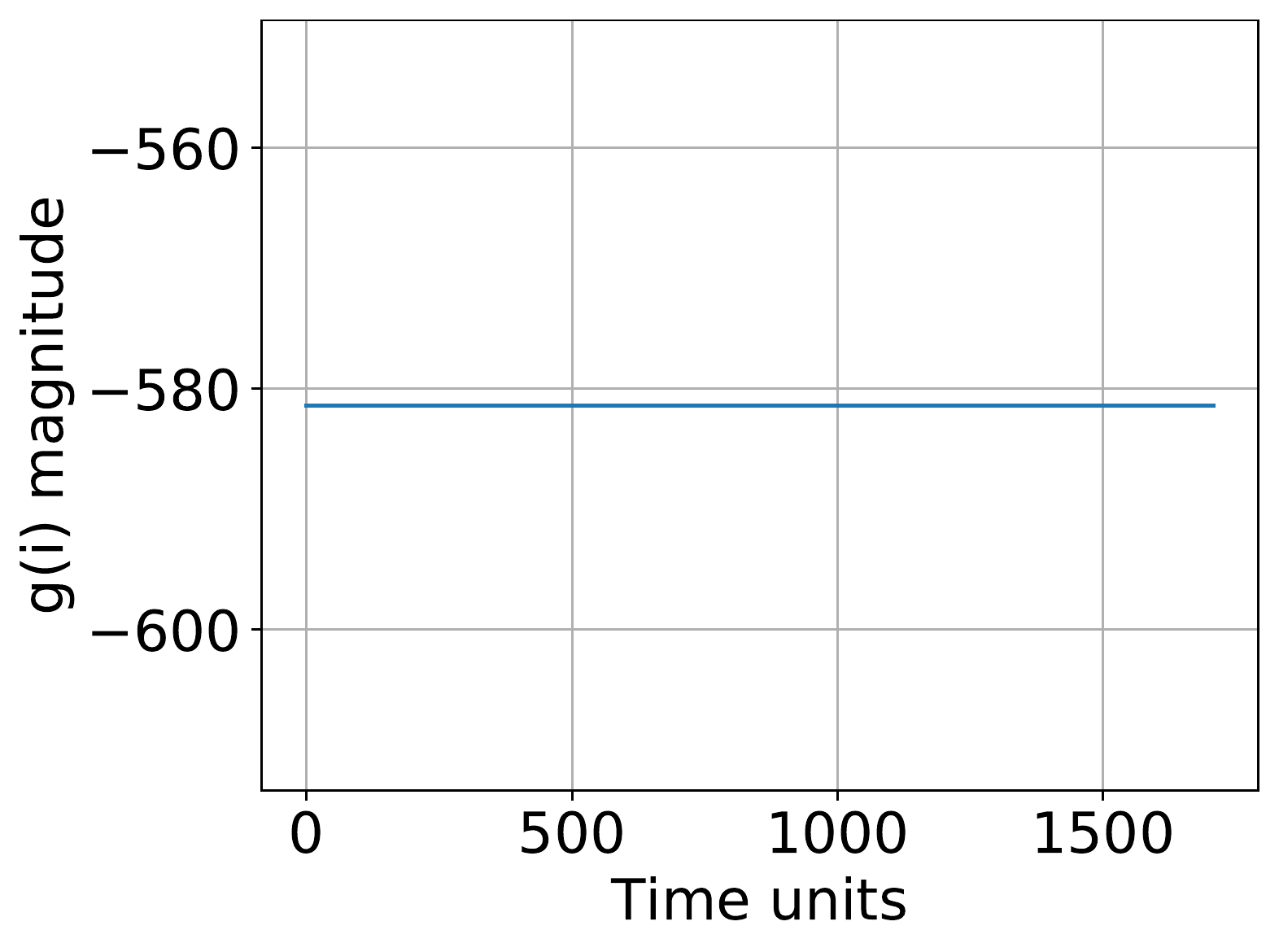}\\
		(a) 	AR-AsLG-HMM & (b)  AsLG-HMM &  (c) LMSAR & (d) AR-MoG-HMM\\
		\includegraphics[scale=0.22]{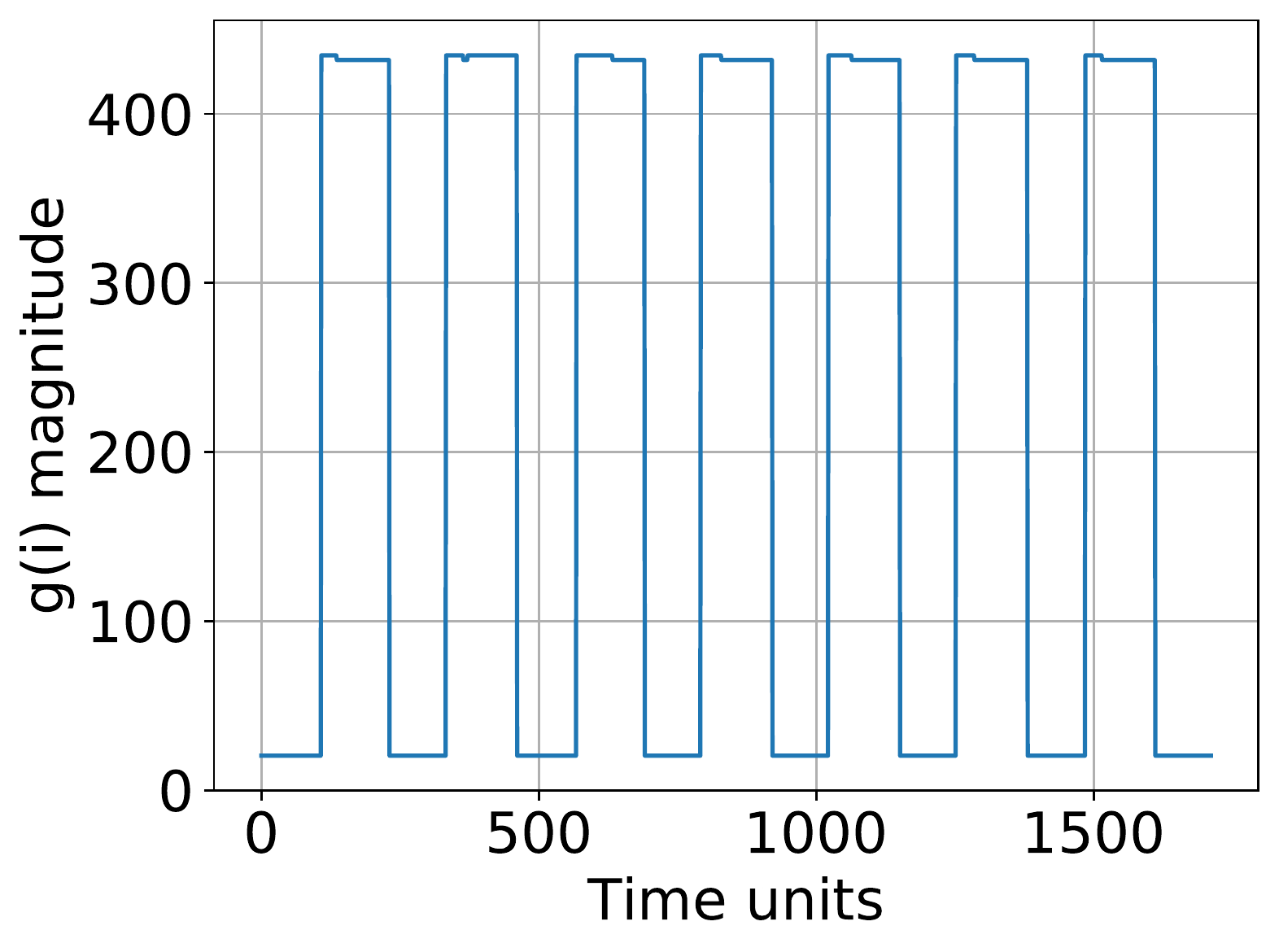}& 
		\includegraphics[scale=0.22]{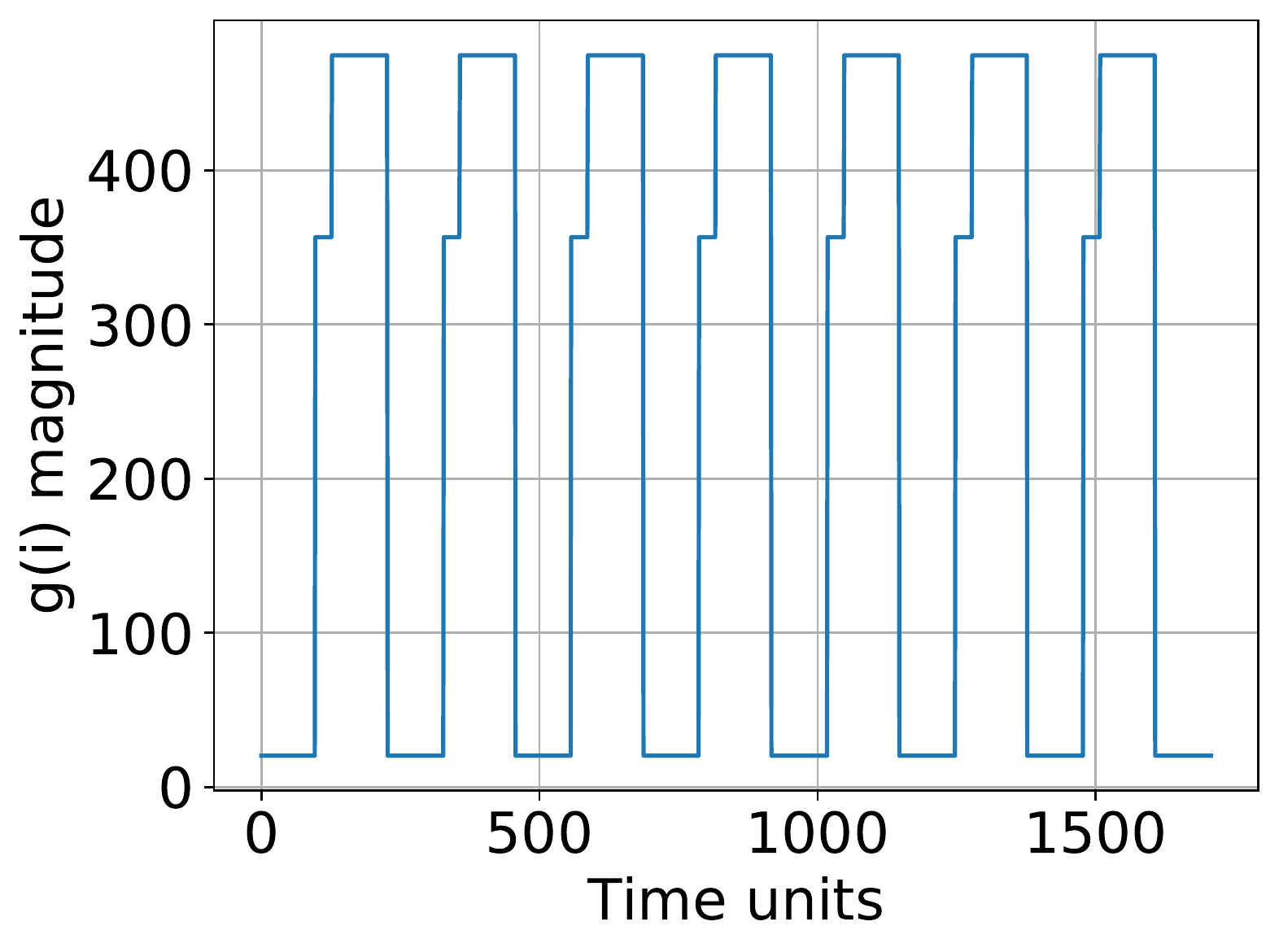}&
		\includegraphics[scale=0.22]{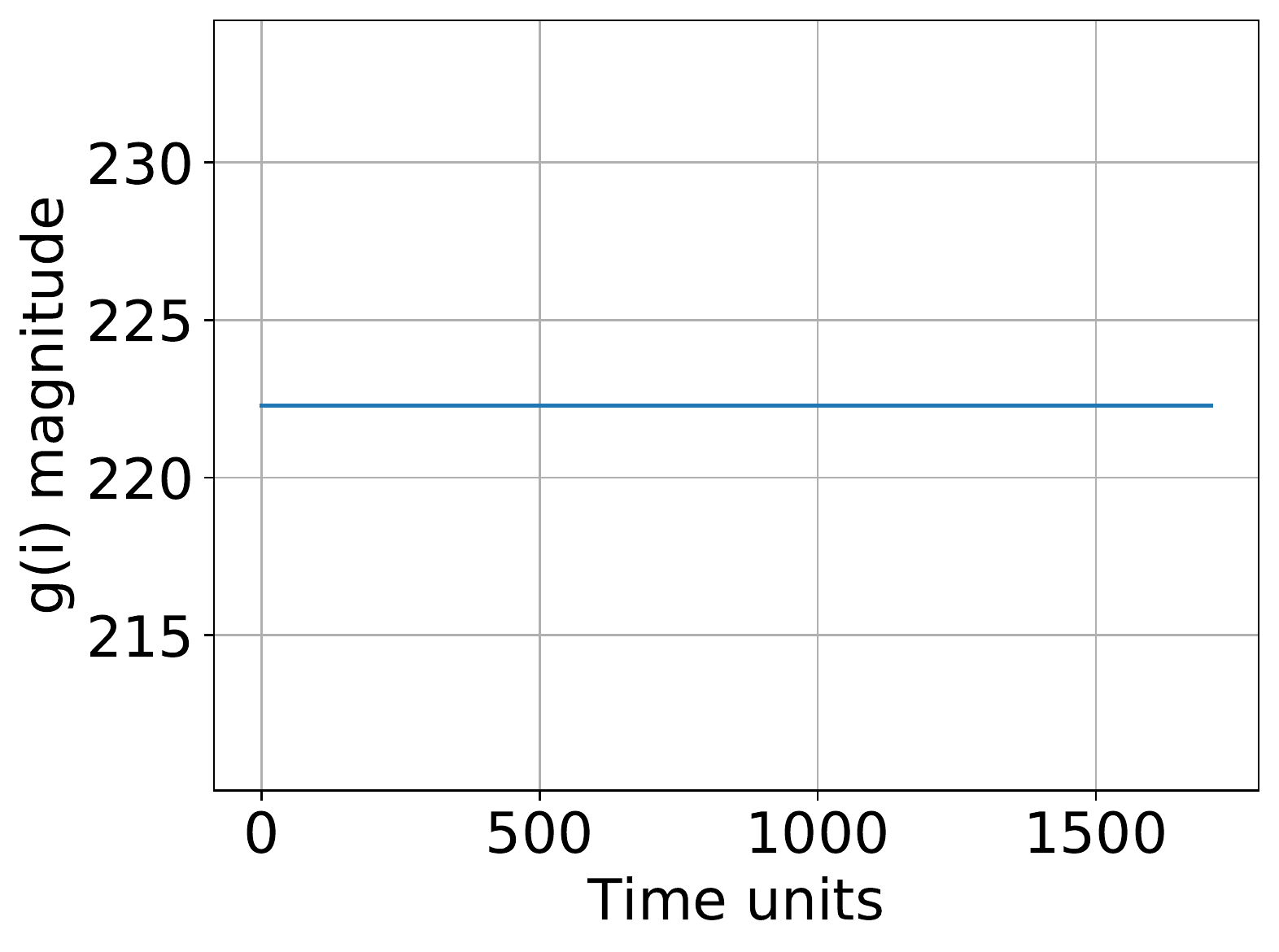}&
		\\
		(e) MoG-HMM & (f) Naïve-HMM & (g) BMM & \\
	\end{tabular}
	\caption{Viterbi paths for scenario 2 and sequence 2}
	\label{fig:s2_s2}	
\end{figure}

\begin{figure}[h]
	\centering
	\begin{tabular}{cccc}
		\includegraphics[scale=0.22]{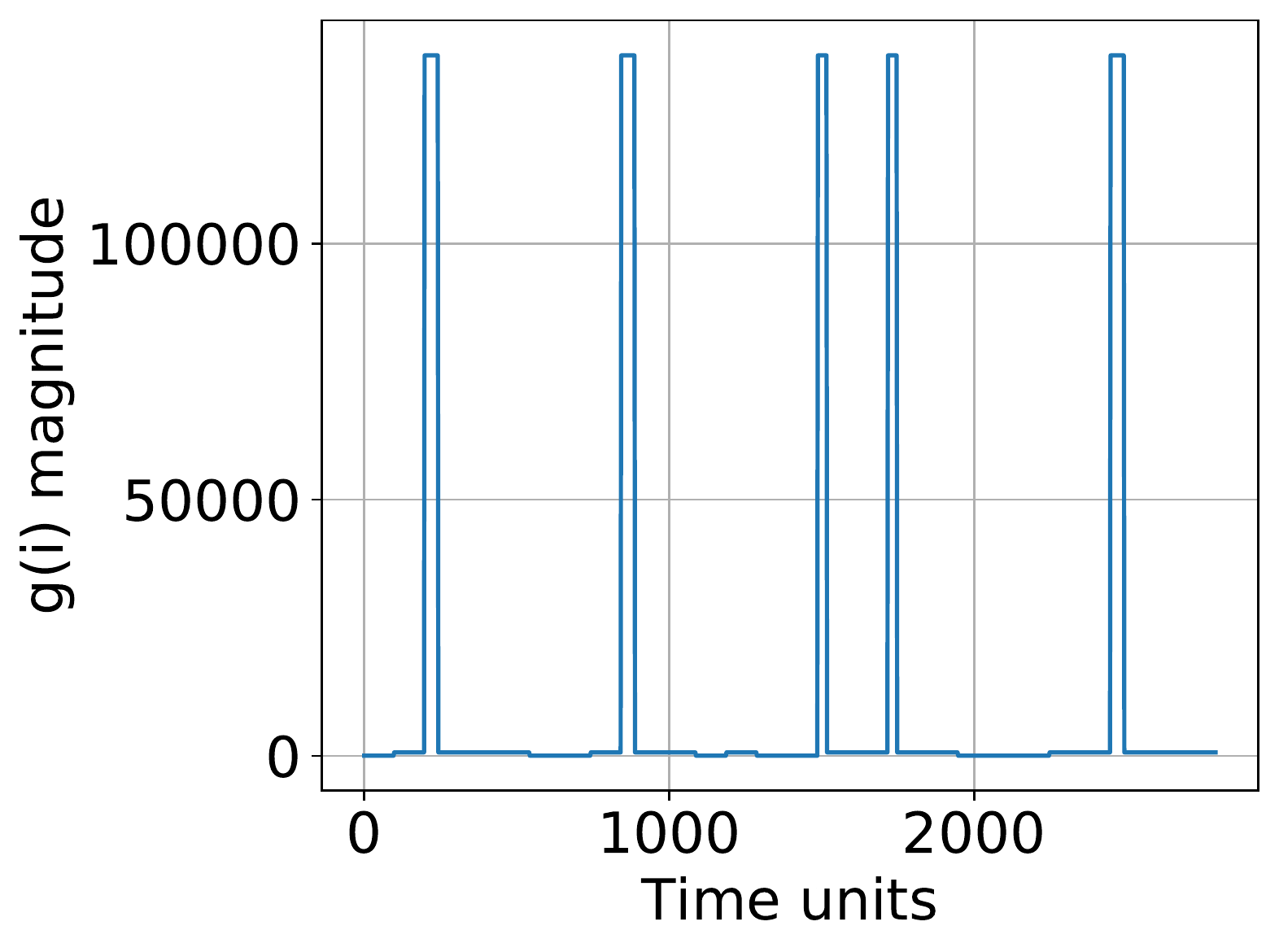} & 		
		\includegraphics[scale=0.22]{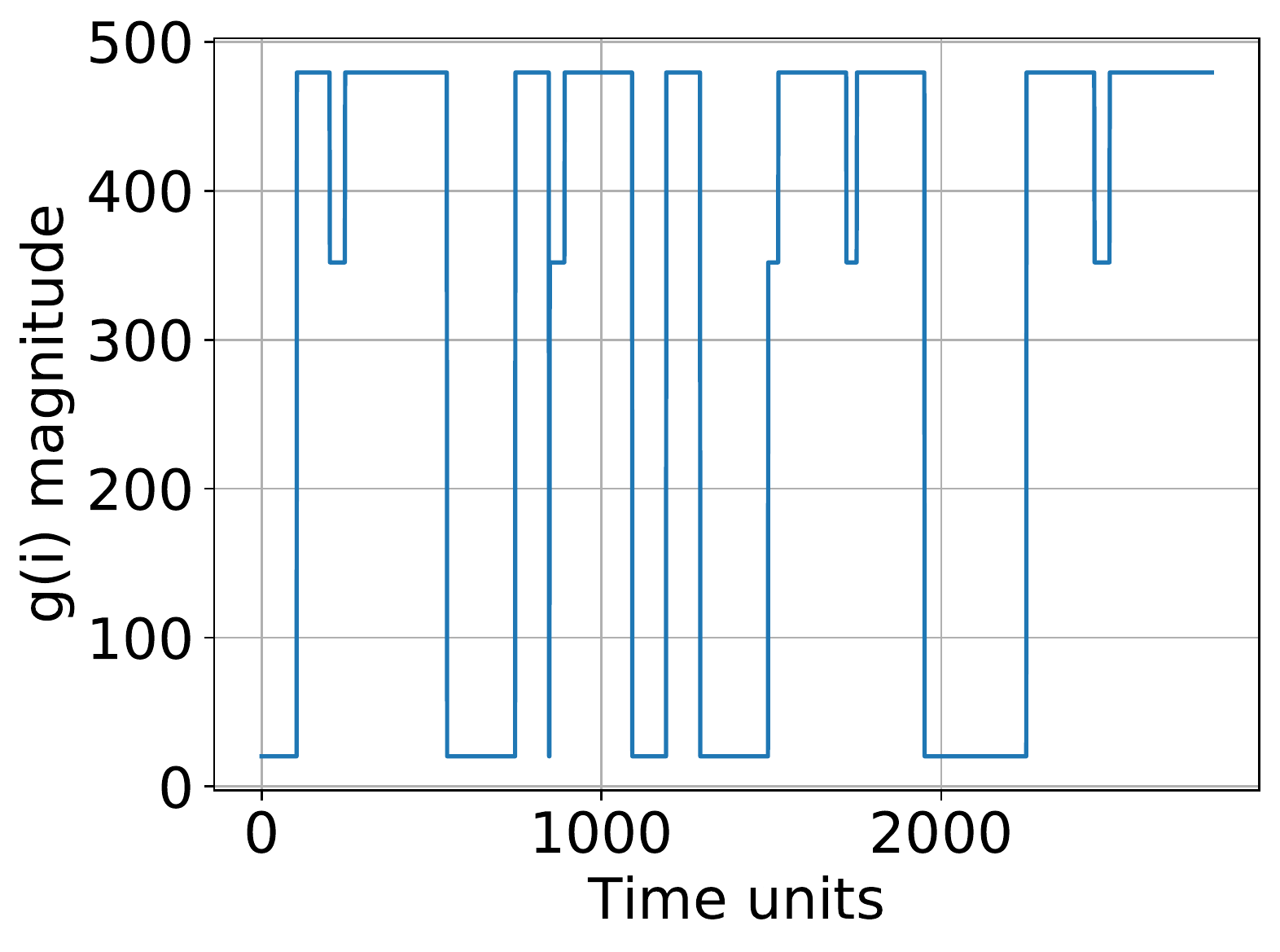}&
		\includegraphics[scale=0.22]{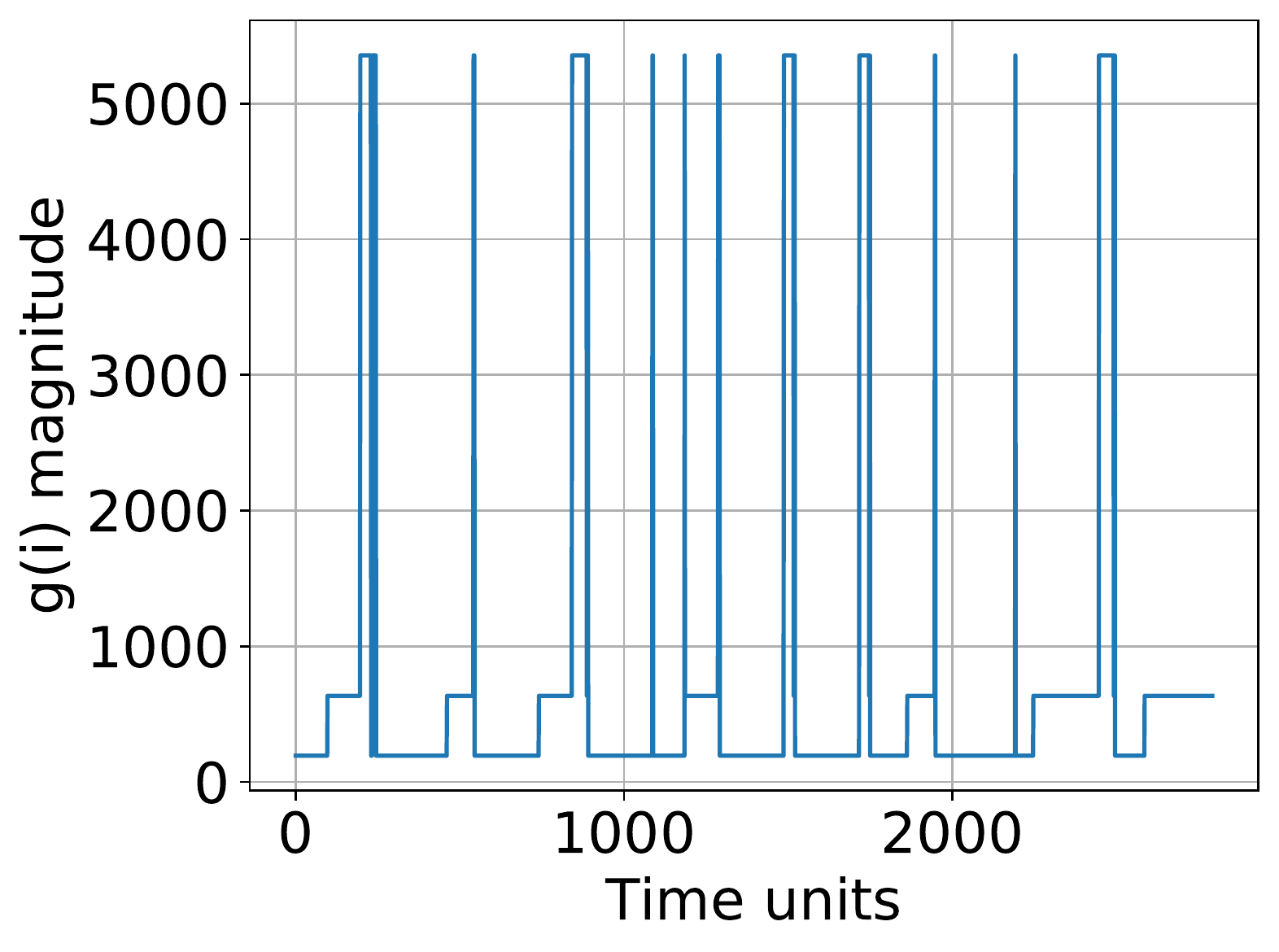}&
		\includegraphics[scale=0.22]{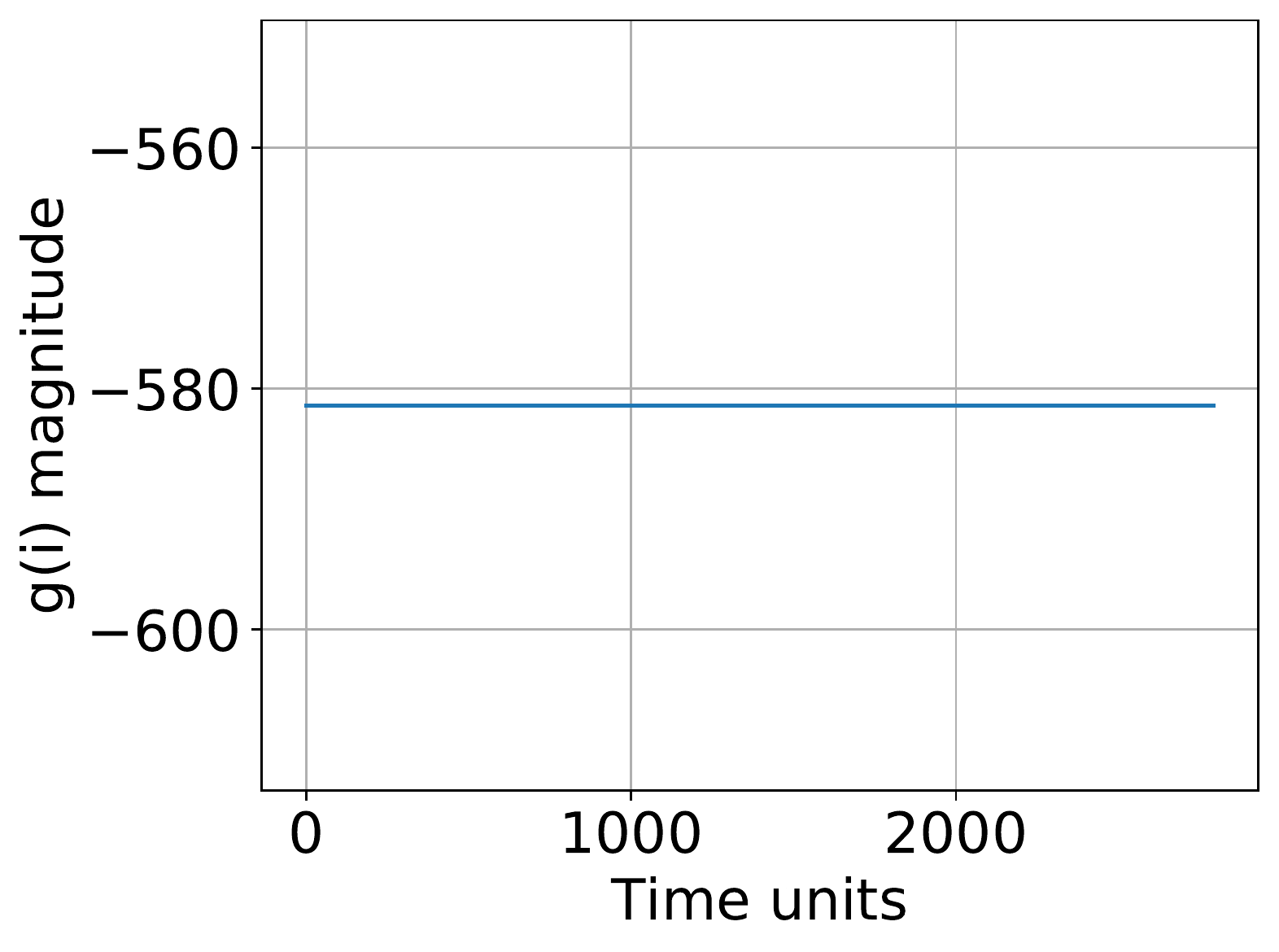}
		\\
		(a) 	AR-AsLG-HMM & (b)  AsLG-HMM & (c) LMSAR & (d) AR-MoG-HMM\\
		\includegraphics[scale=0.22]{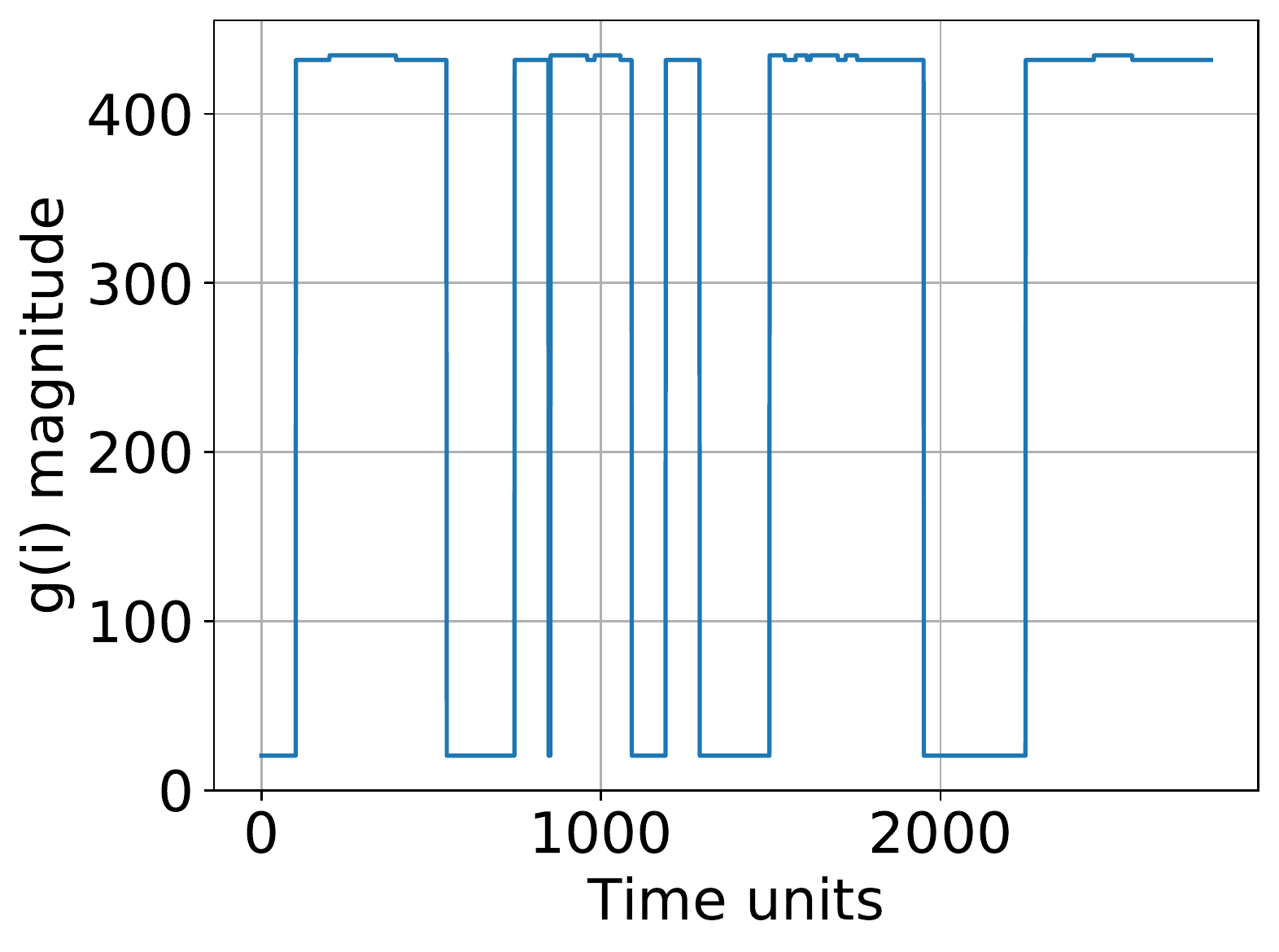}& 
		\includegraphics[scale=0.22]{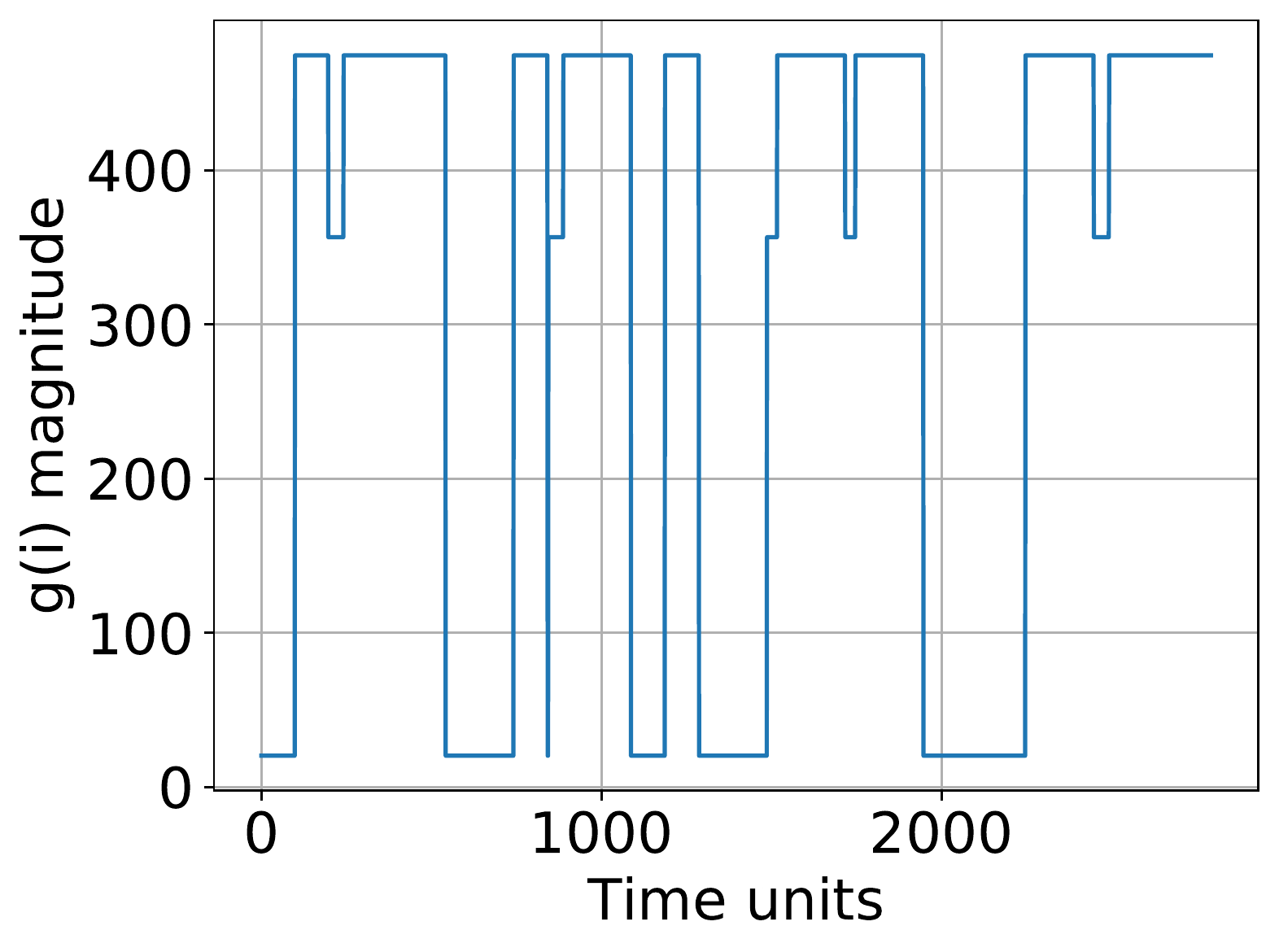}&
		\includegraphics[scale=0.22]{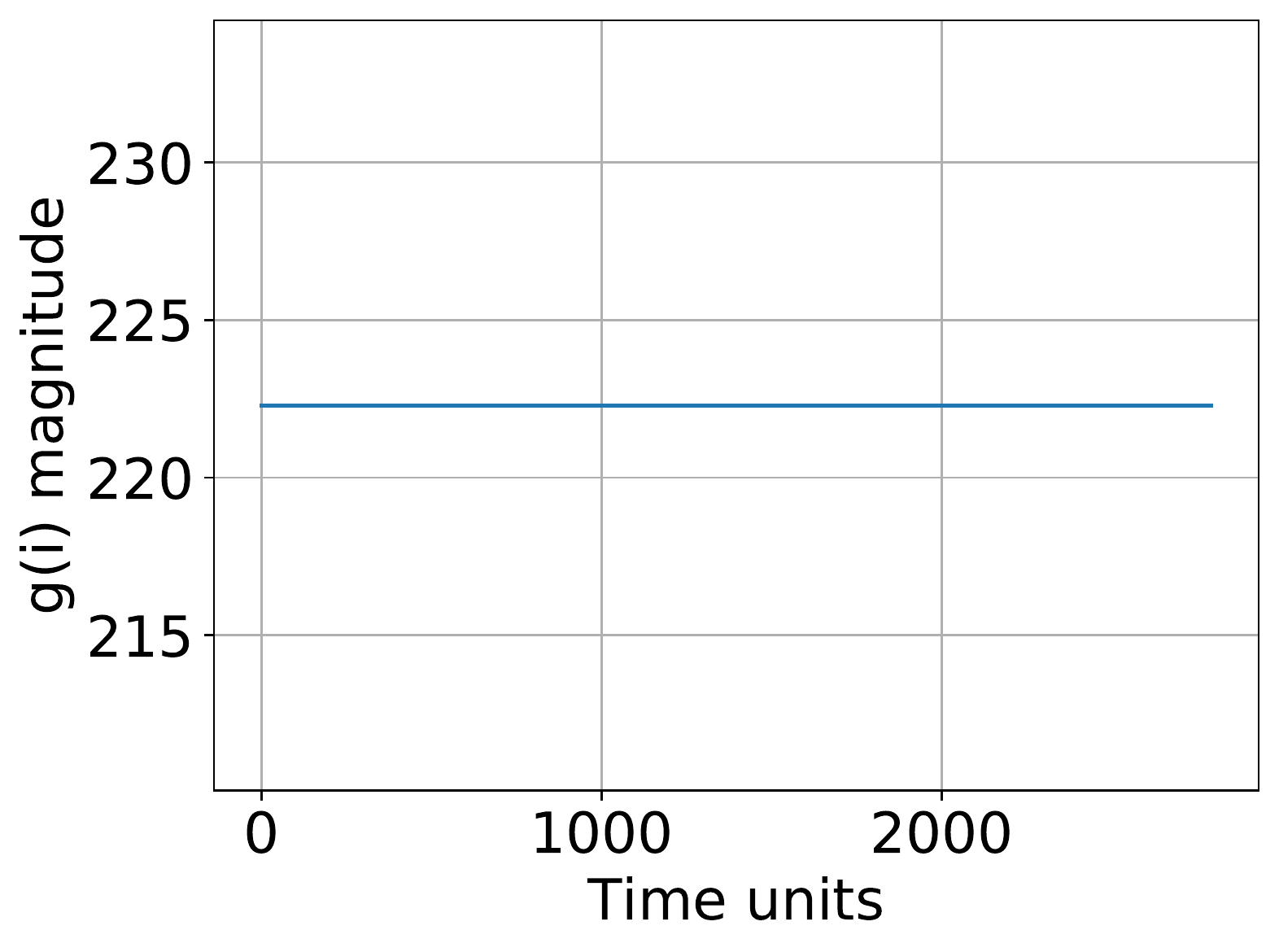}&
		\\
		(e) MoG-HMM & (f) Naïve-HMM&(h) BMM & \\
	\end{tabular}
	\caption{Viterbi paths for scenario 2 and sequence 4}
	\label{fig:s2_s4}	
\end{figure}

Fig.~\ref{fig:s1_s1} and  Fig.~\ref{fig:s1_s3} show the Viterbi paths obtained for sequence 1 and 3 for scenario 1. We are interested in these paths since they tell us the changes in the dynamics of the data for every time instance (smoothing). When new instances arrive, we can use the Viterbi path to determine the hidden state to which the new instance belongs (filtering) and make decisions or analyses of the observed process. In the case of the Viterbi paths obtained by asymmetric models and Naïve-HMM and MoG-HMM, we obtained desirable results, since they follow correctly the pattern of the hidden state sequences showed in Fig.~\ref{fig:sequencetest} in (a) and (c). The remaining models obtained not well-defined lectures of the evolution of hidden states proposed by  Fig.~\ref{fig:sequencetest}. In particular recall from Table~\ref{table:Scenario 1} that AR-MoG-HMM obtained the best results in BIC score, likelihood and standard deviation; however, the predictions are not as expected, this can be caused by the hypothesis of AR-MoG-HMM of constant unitary variance among the hidden states.

Fig.~\ref{fig:s2_s2} and  Fig.~\ref{fig:s1_s3}  show the predicted Viterbi paths for  sequence 2 and 4 for scenario 2. In this case we observe that only AR-AsLG-HMM and MoG-HMM obtained good Viterbi paths that follow the pattern described by Fig.~\ref{fig:sequencetest} (b) and (d). AsLG-HMM and naïve-HMM predicted correctly the transition between hidden states but are wrong in magnitude.  The remaining models obtained poor predictions results.

\begin{table}[H]
	\centering
	\begin{tabular}{lrr}
		Model           &   Training time 1 (s) &  Training time 2 (s)\\
		\hline
		AR-AsLG-HMM     & 5.32   & 6.56\\
		AsLG-HMM        & 3.59  & 5.75\\
		LMSAR           & \textbf{0.58}   & \textbf{0.16}\\
		AR-MoG-HMM      & 26.26   &  18.94\\
		MoG-HMM         & 28.64   & 3.45\\
		Naïve-HMM       & 3.50   &  2.33\\
		BMM				& 181.14 & 400.27\\
	\end{tabular}
	\caption{Scenario 1 and 2 learning times}
	\label{table:times1}
\end{table}

Table~\ref{table:times1} shows the required times for learning the models in each scenario. The number of iterations of the EM algorithm for LMSAR has to be set to one, since further iterations of the algorithm caused nan values. This causes short times of training. In the case of BMM, the EM algorithm after the structure optimization was iterated 1 time before nan values appeared in the models parameters. In the case of scenario 2, AR-MoG-HMM and MoG-HMM had to iterate the EM algorithm 8 times before the appearance of nan values appeared in the emission distribution parameters, this caused short training times in the scenario 2 for AR-MoG-HMM and MoG-HMM. The remaining models and cases could iterate the EM or SEM algorithm until convergence. For both scenarios, we see that the learning times for  BMM are the longest ones. The learning of structures is complex in time since many mutual informations and comparisons  must be computed. We can clearly see that this process scales with the number of variables. On the other hand,  we see that the shortest times with convergence were obtained by Näive-HMM due to the simple structures that the model uses. The proposed model using the forward greedy algorithm took a fair time to be trained taking in consideration the scores obtained and that it was iterated until convergence was met in EM and SEM algorithm.

We can observe from these experiments that AR-AsLG-HMM is capable of being simple enough to explain linear Gaussian autoregressive processes to prevent over-fit, but can be complex enough to detect relevant parameters that drive the hidden states. AsLG-HMM has this property as well, but as can be seen, the AR variables are pertinent. In terms of variance of the predictions, AR-AsLG-HMM also had decent and relevant results, which implies it is stable in the case of synthetic data.

\subsection{Real data}

\subsubsection{Air quality in Beijing}
 Here we use a dataset found in the UCI Machine Learning Repository named: "Beijing Multi-Site Air-Quality Data Data Set" \cite{Zh17}. The dataset consists of measurements of air quality in different monitoring stations in Beijing. We in particular take the measurements from the file "PRSA Data Aotizhongxin" which represent the name of the monitoring station Aotizhongxin. This dataset has hourly measurements from March 2013 until February 2017. The data contains missing data (3.37\% of the dataset for the selected variables). The missing data is filled using the mean of the values of the five previous hours. The hidden variable in this problem can be understood as the air quality. For this study we use the following variables:  sulfur dioxide ($\text{SO}_2$ in $\mu g/m^3$), nitrogen dioxide ($\text{NO}_2$ in $\mu g/m^3$), carbon monoxide ($\text{CO}$ in $\mu g/m^3$), ozone ($\text{O}_3$ in $\mu g/m^3$), coarse particulate matter ($\text{PM}_{10}$ in  $\mu g/m^3$) and  fine particulate matter ($\text{PM}_{2.5}$ in $\mu g/m^3$). Bayesian networks and HMM have been used before to determine air quality \cite{Vi17,Va19,Ya17,Su12}, showing  advantages in the generation of information and discovery of relationships between variables.

The Chinese air quality limits for hourly, daily and monthly measurements are expressed in the law GB 3095-2012. These limits are used to model the $g$ function for this problem. In particular, $\boldsymbol{\kappa} = \{500,200,10000,200,150,75\}$ and $\textbf{v} = \{1/500,1/200,1/10000,1/200,1/150,1/75\}$. The $g$ function in this case uses Eq.~(\ref{eq:label2}). If $g_2(i)>0$ it means that one or many variables are above the permissible limit and the air quality is pretty bad. Great negative values are desirable for $g_2$ to indicate good air quality. The aim is to learn models to determine the air quality when new observations arrive. We use the year 2013 to train the models. The number of hidden states is set to two to indicate those two states: good air quality and bad air quality. For each model we predict individually the air quality for year 2014, 2015 and 2016. We present the mean likelihoods (or the mean likelihood of the tested years), mean BICs, variance of likelihoods and number of parameters of each model.

\begin{table}[H]
	\centering
	\begin{tabular}{lrrrr}
		Model &  mean LL & mean BIC & Std & $\#$ \\
		\hline
AR-AsLG-HMM      &-180018.03 &360648.30 &81629.12 &71\\
AsLG-HMM         &-222147.85 &444813.07 &101657.73 &60\\
LMSAR 		     &-195975.10 &392571.05 &87441.29 &72\\
AR-MoG-HMM   	 &\textbf{-131606.95} &\textbf{264300.41 }&\textbf{60262.94} &126\\
MoG-HMM 		 &-217369.97 &435722.97 &99633.45 &114\\
Naïve-HMM   	 &-229340.56 &458939.80 &104667.49 &\textbf{30}\\
BMM   		     &-217491.34 &435965.70 &99672.93 &114\\
	\end{tabular}
	\caption{Scores for air quality}
	\label{table:Resultados_Aire}
\end{table}

\begin{figure}[h]
	\centering
	\begin{tabular}{cc}
		\includegraphics[scale=0.6]{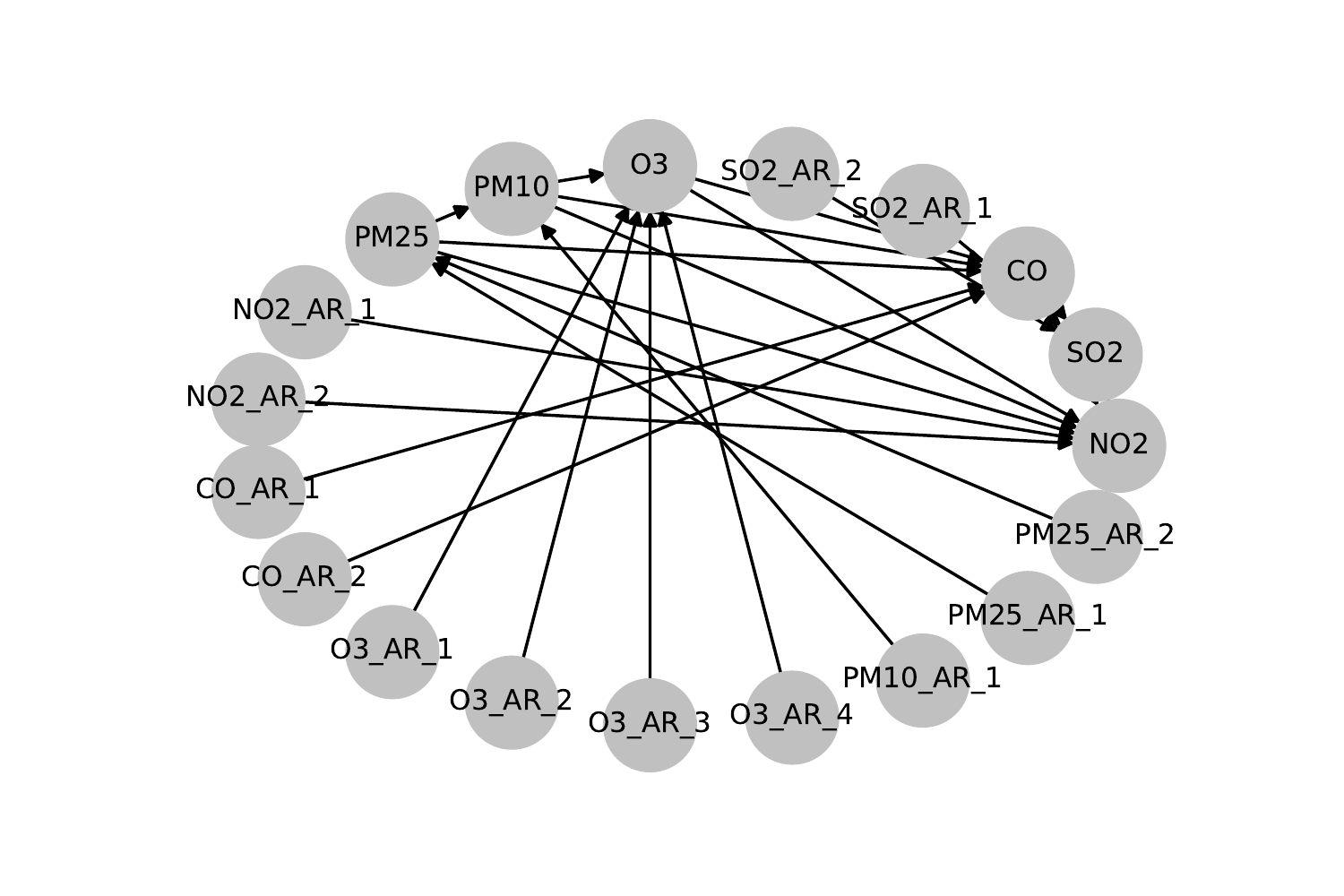} &
		\includegraphics[scale=0.6]{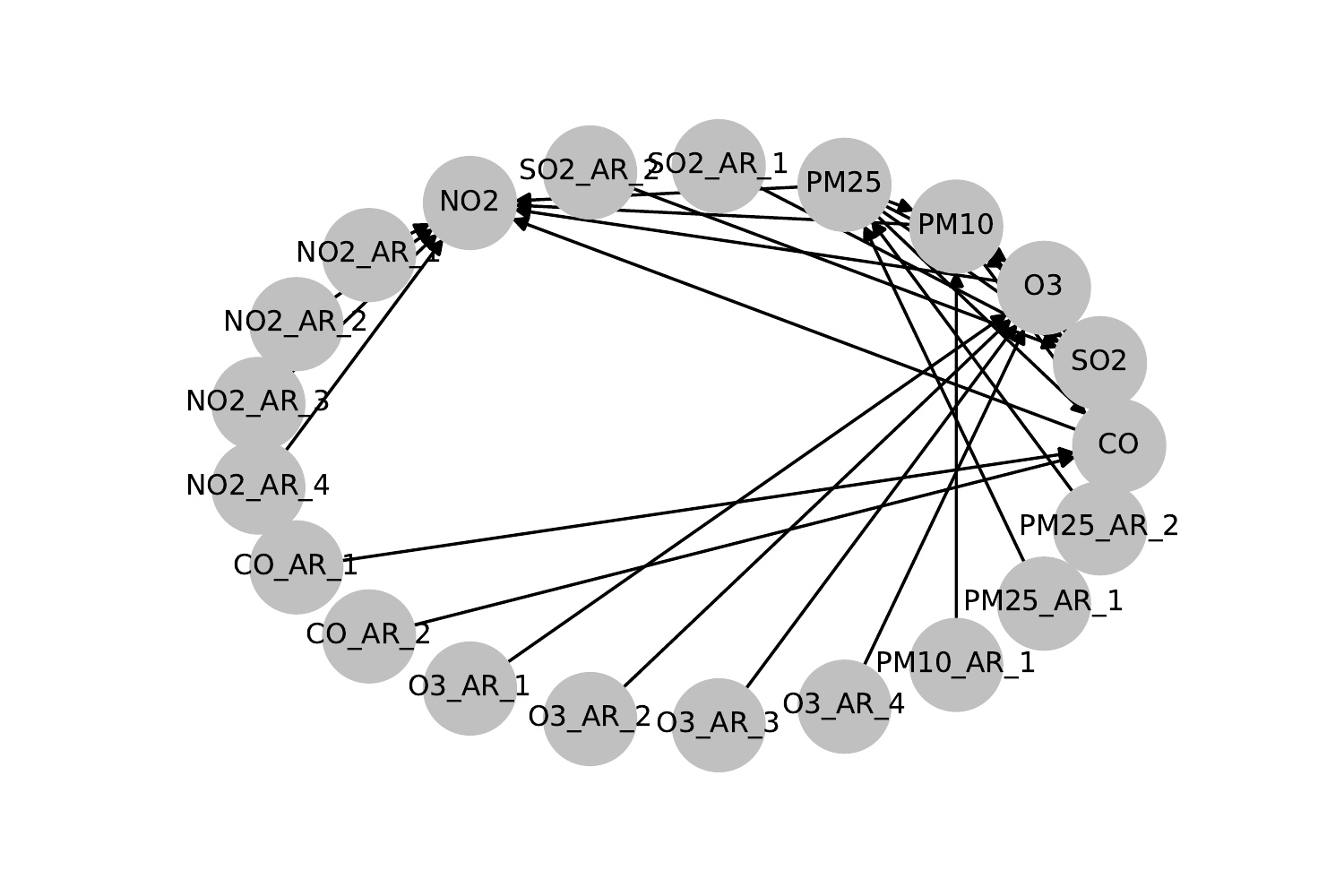}\\
		(a) $g(i) = -0.78$& (b) $g(i) = 0.7$ \\
	\end{tabular}
	\caption{Context-specific graphs learned by AR-AsLG-HMM. (a) shows a graph where the air quality is good and (b), where the air quality is bad.}
	\label{fig:grap_air}	
\end{figure}

\begin{figure}[H]
	\centering
	\begin{tabular}{ccc}
		\includegraphics[scale=0.25]{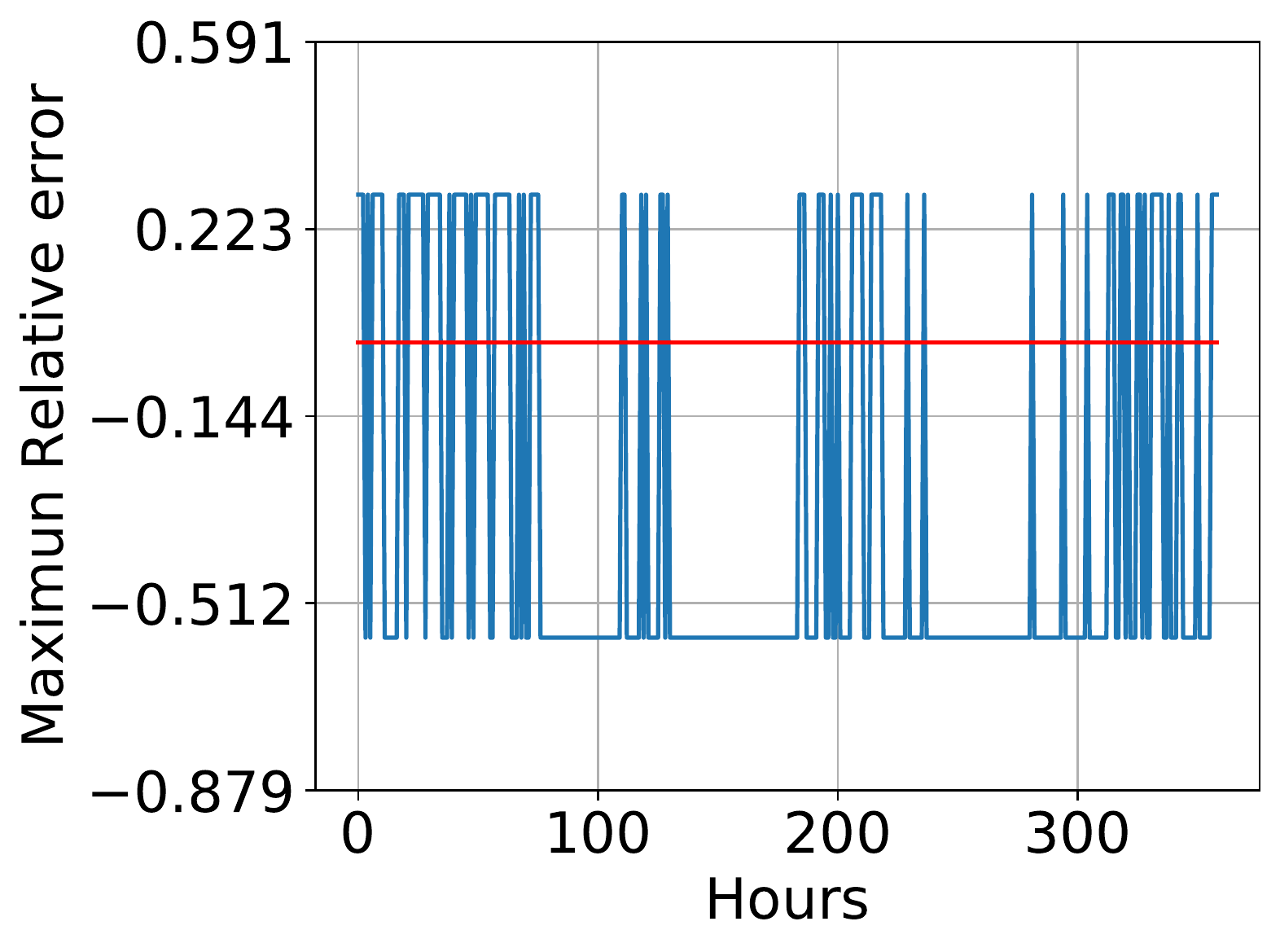} &
		\includegraphics[scale=0.25]{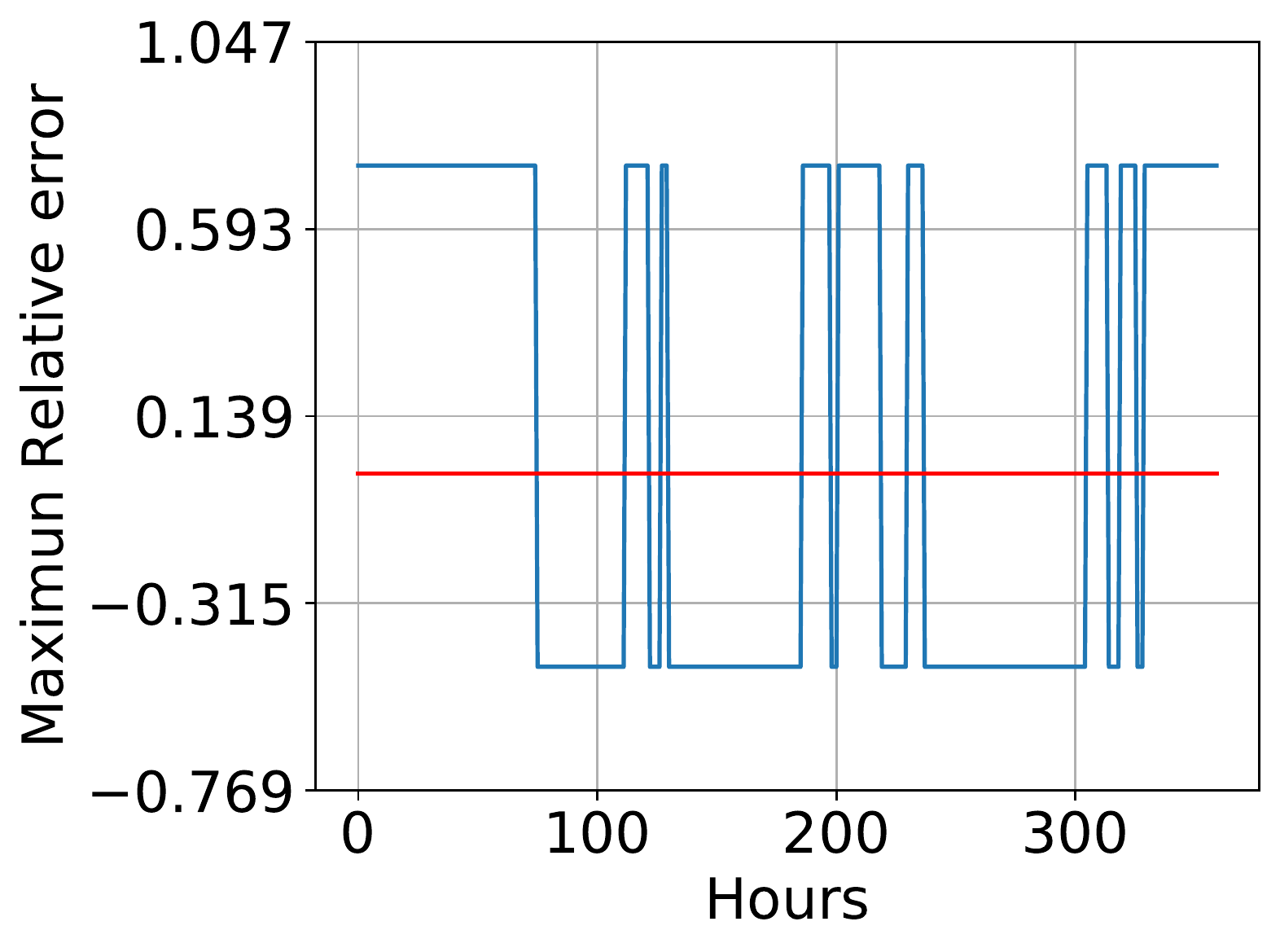}&
		\includegraphics[scale=0.25]{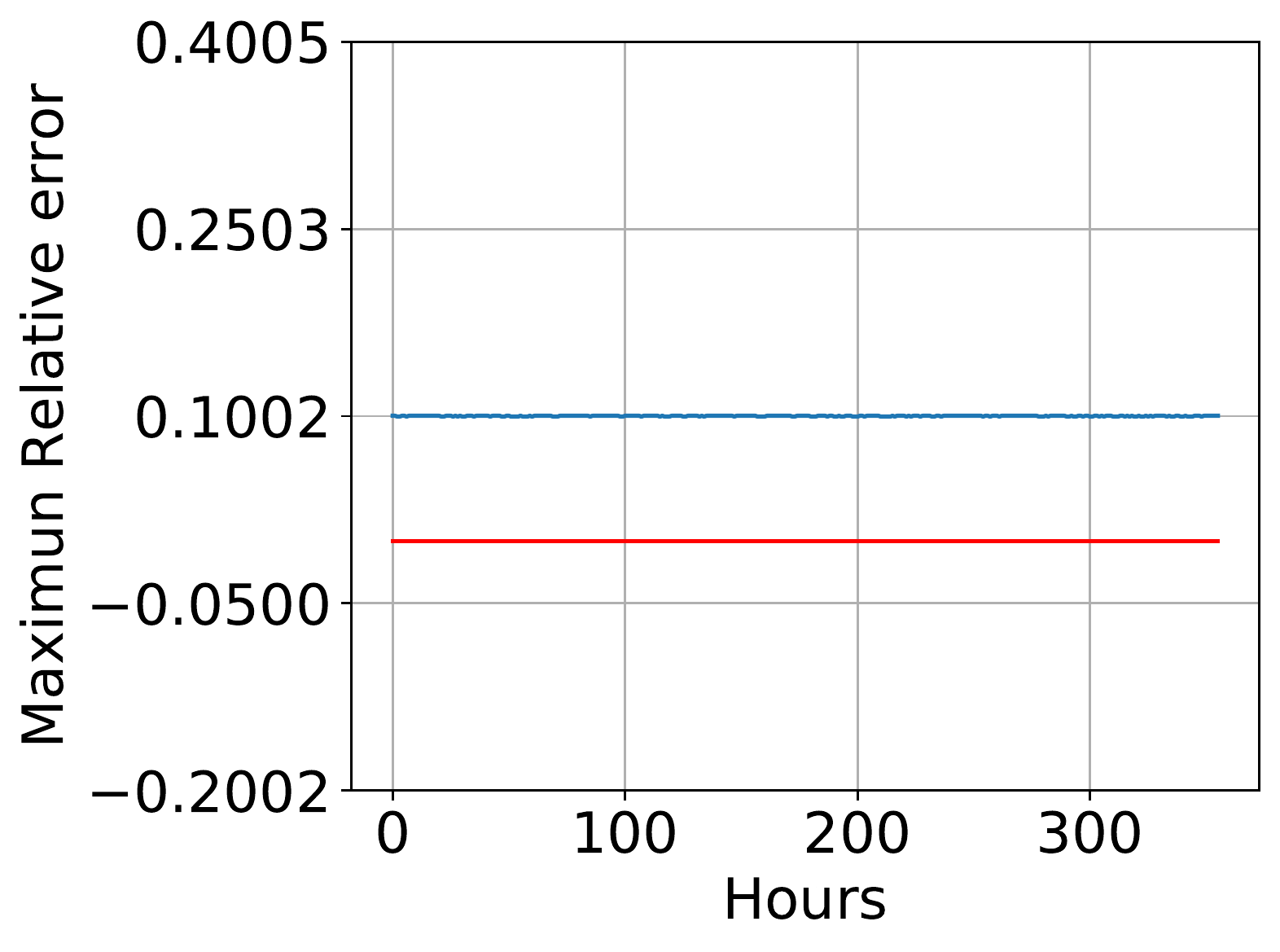}\\
		(a) AR-AsLG-HMM& (b) AsLG-HMM & (c) LMSAR\\
		\includegraphics[scale=0.25]{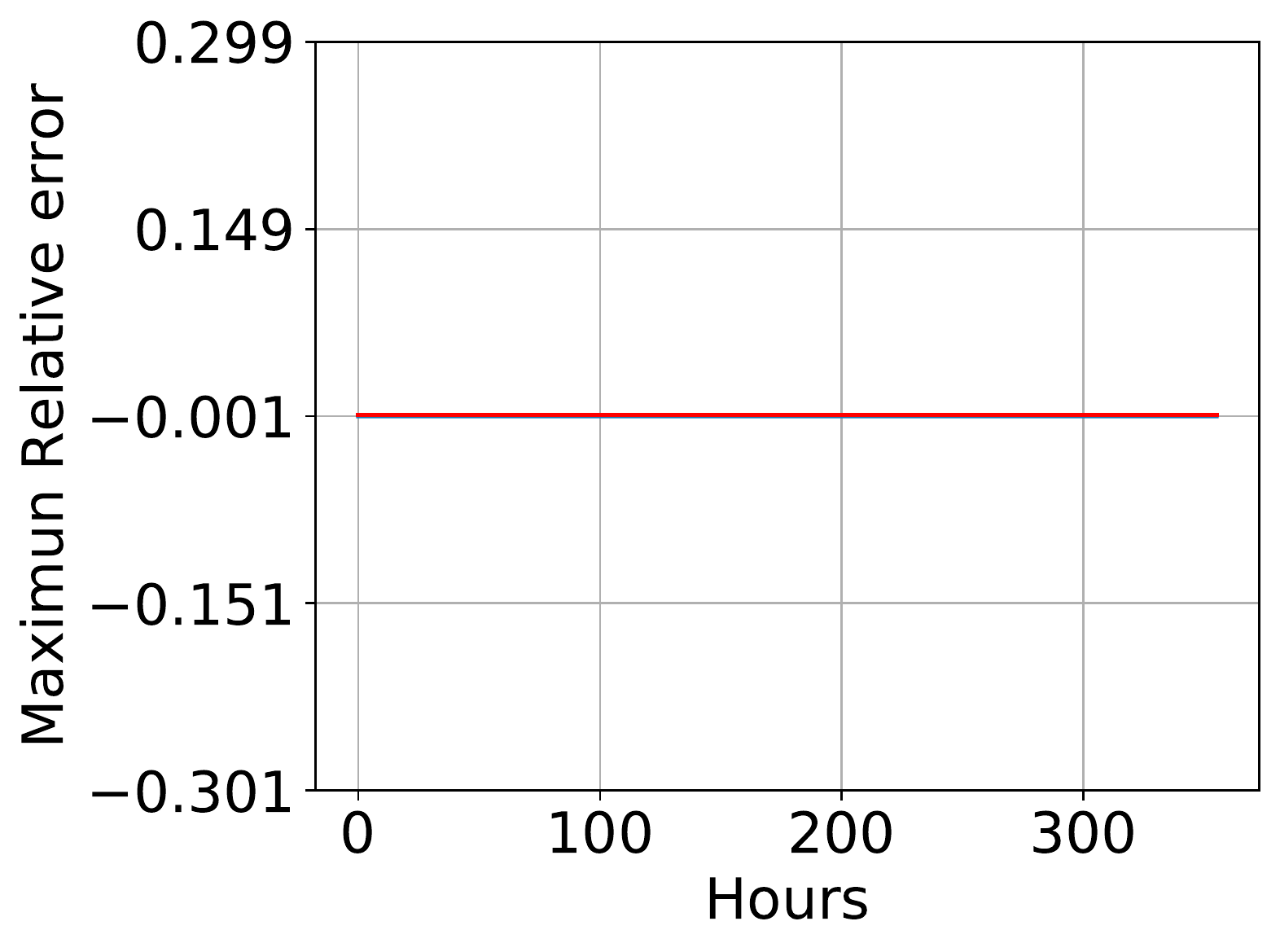} &
		\includegraphics[scale=0.25]{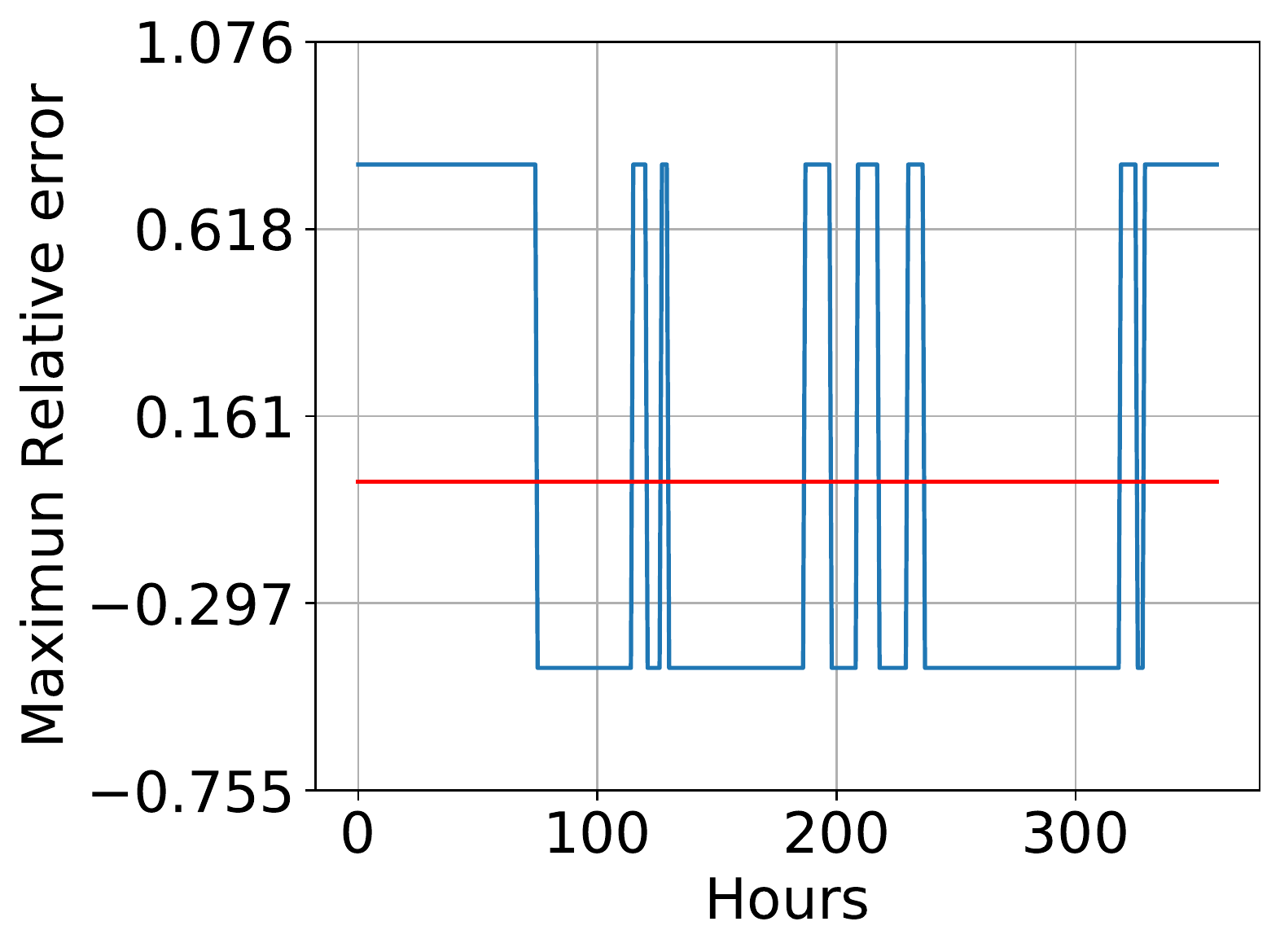}&
		\includegraphics[scale=0.25]{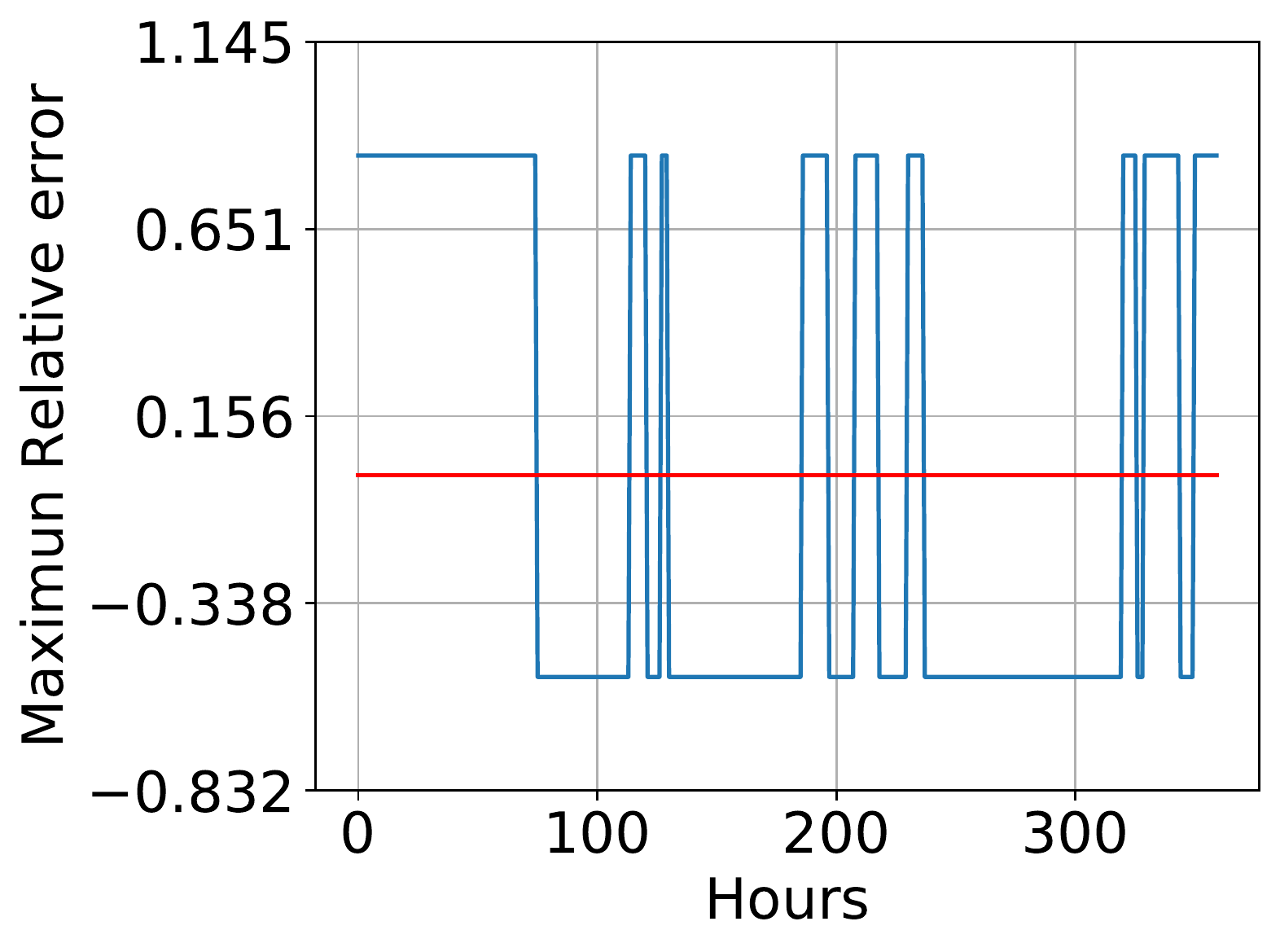}\\
		(d) AR-MoG-HMM&(e) MoG-HMM & (f) Naïve-HMM \\
		\includegraphics[scale=0.25]{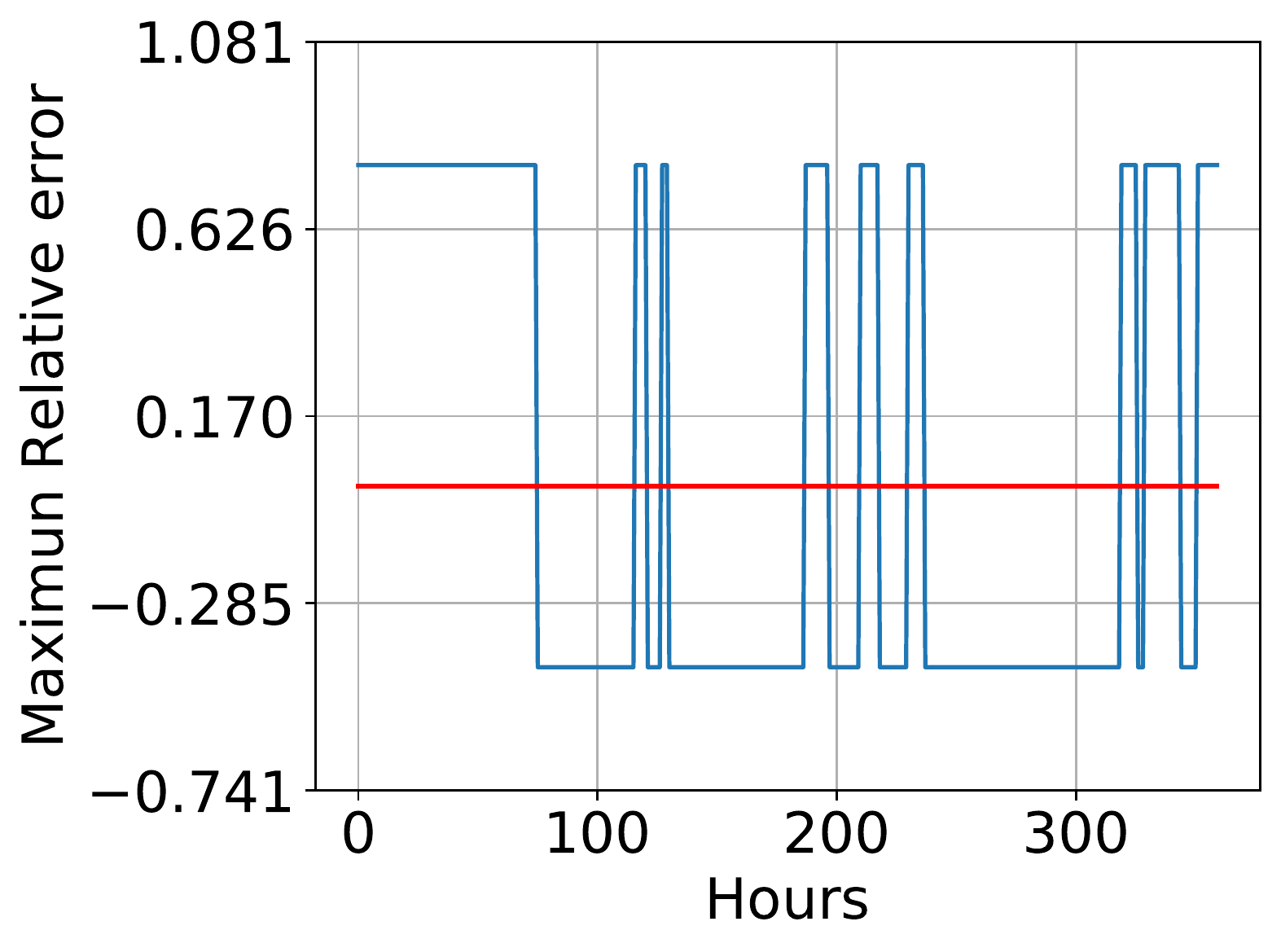}  &
		\includegraphics[scale=0.25]{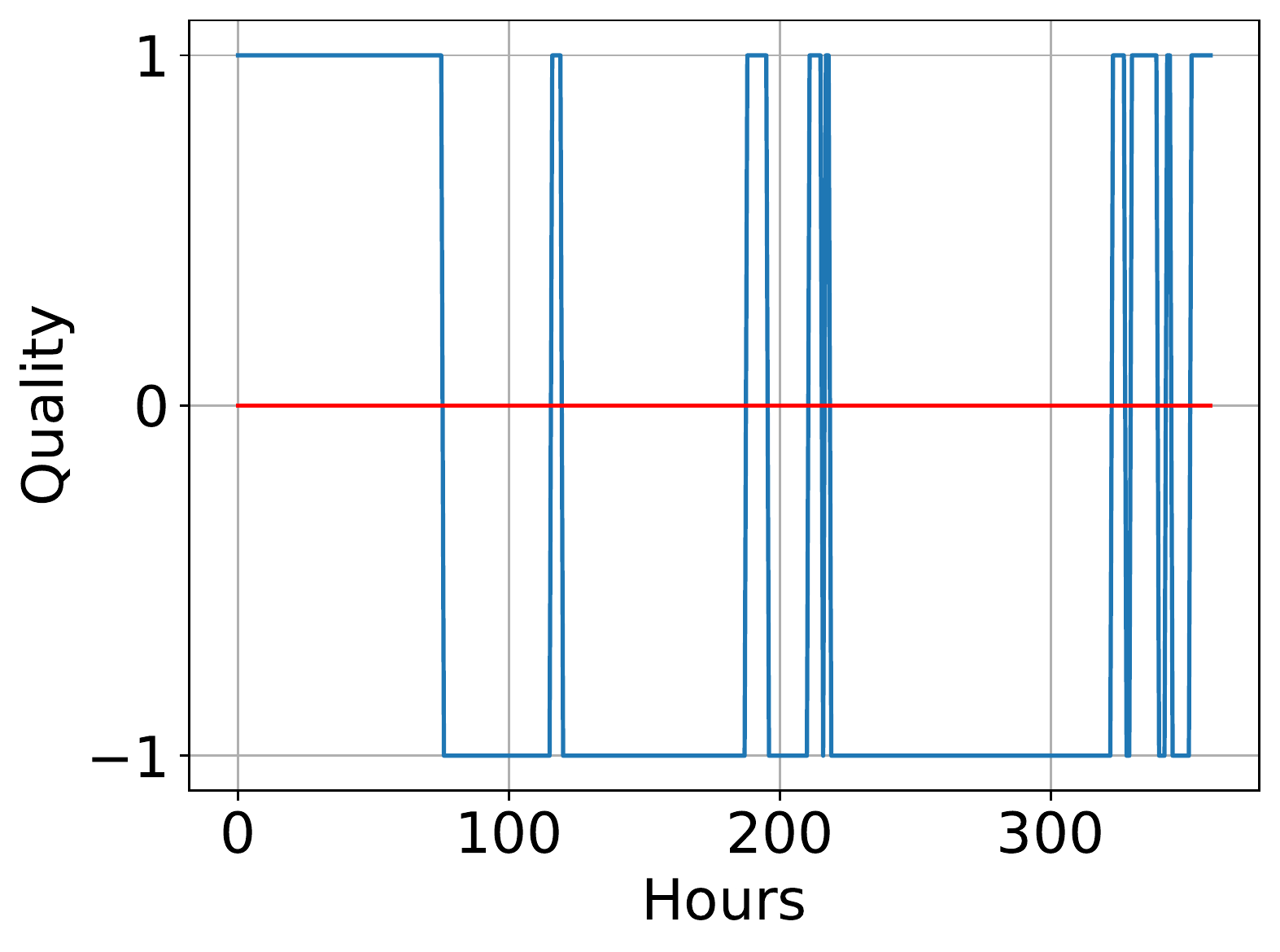} &\\
		(g) BMM & (h) Real&  \\
	\end{tabular}
	\caption{Viterbi paths for the air quality  example during the first week of 2016}
	\label{fig:air_quality}	
\end{figure}

Table~(\ref{table:Resultados_Aire}) shows the scores obtained from the testing years. We can observe that AR-AsLG-HMM and AR-MoG-HMM obtained the best results in the LL and BIC scores. In terms of stability, we see that also, AR-AsLG-HMM and AR-MoG-HMM have the lowest standard deviation, followed by LMSAR and MoG-HMM . In term of number of parameters, we observe that naïve-HMM and AsLG-HMM have the lowest number of parameters. Followed by these models, AR-AsLG-HMM and LMSAR obtained a fair number of parameters and finally, mixture models as expected had to use more parameters (up to four times the number of parameters used by Naïve-HMM). In the case of BMM, the learning of the structure could not be performed since nan values appeared in the re-training phase.

Fig.~\ref{fig:grap_air} shows two learned graphs. In the context-specific Bayesian networks, AR variables are denoted as $X_m\_\text{AR}\_r$, where $r$ is the number of lags for the variable $X_m$. In (a) we show a graph when the air quality is good and in (b) is bad.  In both graphs we can observe some interesting relationships similar to the ones founded in \cite{Vi17}. For example, in (a) we observe that $\text{CO}$ depends on $\text{PM}_{2.5}$ and  $\text{PM}_{10}$ and $\text{SO}_2$ and $\text{NO}_2$ are related to CO. These relationships come from the process of combustion of gas and charcoal. Also $\text{NO}_2$ is related to $\text{O}_3$ which indicates the photochemistry of $\text{NO}_2$ for the production of $\text{O}_3$. In (b) we see that these relationships remain. However, the dependences on previous values for each variable changes, which tells us the level of impact of the past on the pollution levels.

Fig.~\ref{fig:air_quality} shows the predicted air quality for the first two weeks of 2016 for each model using the Viterbi algorithm. Real readings are shown in (h), where we express 1 when any of the variables surpasses the law limits and -1 when all the variables are under the law limits. Clearly (h) does not tell us the severity of the pollution nor the closeness to an outlaw pollution level. We observe that AR-AsLG-HMM is capable of observing the periods of high pollution and low pollution but in a noisy way, on the other hand, the AsLG-HMM, MoG-HMM, Naïve-HMM and BMM make clearer predictions. On the other hand, AR-MoG-HMM and LMSAR obtained poor predictions in spite of their good scores in mean BIC and Std. Recall that in both models, the variance is assumed to be the same for all the hidden states, which for this dataset is not true. 
\subsubsection{Ball-bearings degradation}

Ball-bearings are used inside many mechanic tools as drills, rotors, etc. Ball-bearings represent critical components inside these machines. The failure  or degradation of these components can be translated to loses in time, money and assets for industries. Monitoring ball-bearings is crucial and relevant, and the use of HMM can give insight of the bearing degradation process and therefore help in the development of maintenance policies \cite{La18}. 

\begin{figure}[h]
	\centering
	\includegraphics[width=0.3\textwidth]{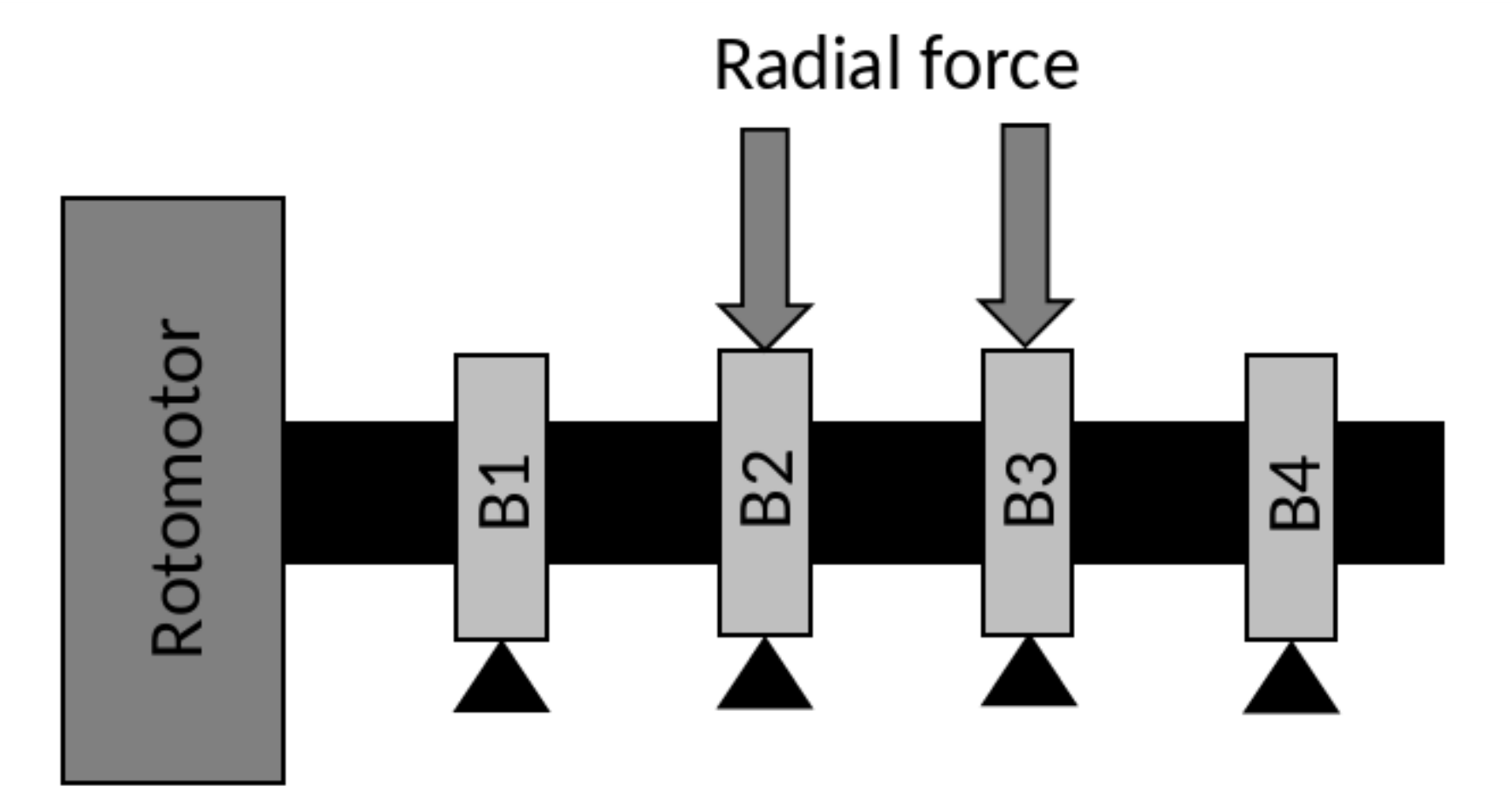}
	\caption{Graphical representation of the mechanical setup. A rotomotor spins a shaft at a rotational speed of 2000 RPM coupled with four Rexnord ZA-2115 double row ball-bearings with labels B1, B2, B3 and B4. A constant radial load of 2721.554 kg  is applied to B3. A signal record of 0.1 s is taken every twenty minutes. The sampling rate is 20 kHz.} \label{fig:setup}
\end{figure}

The benchmark used to validate the proposed model in this section comes from ball-bearing vibrational data  \cite{Qian17}. The run-to-failure tool machine setup is shown in Fig.~\ref{fig:setup}.   Four ball-bearings are tested in the setup. The signals are obtained with vibrational sensors. The desired vibrations are submerged in noise; therefore, filtration techniques are required. In this study, the signals are filtered as in \cite{Wang10}, where spectral kurtosis algorithms are used. From the filtered signal, we calculate its spectrum with the Fourier transform and the ball-bearing fundamental frequencies, namely, ball pass frequency outer (BPFO) related to the ball-bearings outer race, ball pass frequency inner (BPFI) related to the ball-bearings inner race, ball spin frequency (BSF) related to  the ball-bearings rollers and the fundamental train frequency (FTF) related to the ball-bearings cage.

The training signal consists of 2156 records, while the testing signal has 6324 records. We use the fundamental frequencies as variables of the models, hence four variables are used. We must recall that the dataset comes from a coupled mechanical system. Therefore, in the presence of a fail in any part of the system,  vibrations will be generated that will transmit across the whole system.  For all the ball-bearings, the magnitude of their frequencies grows abruptly at the end of the measures indicating a phase of ball-bearing failing somewhere in the mechanical setup. In the training dataset, Bearing 3 fails due to its inner race and Bearing 4 due to its rollers. In the testing dataset Bearing 3 fails due to its outer race. The hidden variable for this context can be understood as the ball-bearing health state. The number of hidden states was set depending on the scores obtained by naïve-HMM in the training data. We observed that with seven hidden states, the scores of the naïve-HMM were optimized.

For this problem, we constantly obtained underflow problems due to the small amplitudes of some frequencies. Therefore, all the dataset were multiplied by 1000. The $g$ function uses  $\textbf{v} = \{1/1000,1/1000,1/1000,1/1000\}$ and $\boldsymbol{\kappa} =\{0,0,0,0\}$. Therefore if we use Eq.~(\ref{eq:label}), $g_1$ adds the magnitude of all the relevant frequencies. If there is a degradation in any of the ball-bearing components, the relevant frequencies will have greater magnitudes and this will be perceived by $g_1$. Therefore, predicting the hidden state in the testing data can be seen as an approximation of the degradation of the ball-bearing. The idea here is to train models such that they can determine the degradation state of forthcoming ball-bearings. This can be accomplished with the Viterbi paths. In particular for this dataset, we are interested in Bearing 3, since it fails in both the training and testing dataset. Nevertheless, we will train the models for all the bearings and show the scores obtained in the testing dataset. Additionally we show the Viterbi paths of Bearing 3 to see the respective degradation. 

During the training time, the iterations of the LMSAR and BMM had to be tuned to prevent nan values in the parameters. Additionally, for the BMM, no structural optimization was performed, since it was unfeasible in time.

\begin{table}[H]
	\centering
\begin{tabular}{llrrr}
	B&Model &  LL & BIC  & $\#$ \\
	\hline
B1&AR-AsLG-HMM   &\textbf{-32372.06}  &\textbf{65881.49}  &130\\
&AsLG-HMM        &-32618.64 &66295.91  &121\\
&LMSAR 		     &-43531.13 &88812.05  &200\\
&AR-MoG-HMM   	 &-34481.73 &71903.11  &336\\
&MoG-HMM 		 &-39843.90 &81892.53  &252\\
&Naïve-HMM   	 &-35512.49 &72004.87  &\textbf{112}\\
&BMM   		     &-35723.94 &73652.60  &252\\
	\hline
B2&AR-AsLG-HMM   &-34576.76 &70220.90  &122\\
&AsLG-HMM        &-36392.74 &73835.36  &120\\
&LMSAR 		     &\textbf{-29191.95} &\textbf{60133.68}  &200\\
&AR-MoG-HMM   	 &-31965.66 &66870.96  &336\\
&MoG-HMM 		 &-38355.91 &78916.55  &252\\
&Naïve-HMM   	 &-37160.21 &75300.29  &\textbf{112}\\
&BMM   		     &-38967.22 &80139.17  &252\\
\hline
B3&AR-AsLG-HMM   &-103952.04 &209120.18  &139\\
&AsLG-HMM        &-107973.53 &217075.67  &129\\
&LMSAR 		     &\textbf{-44120.05}  &\textbf{89744.93}   &172\\
&AR-MoG-HMM   	 &-49412.30  &101274.30  &280\\
&MoG-HMM 		 &-114615.57 &231435.86  &252\\
&Naïve-HMM   	 &-108900.10 &218780.08  &\textbf{112}\\
&BMM   		     &-132201.55 &266607.84  &252\\
	\hline
B4&AR-AsLG-HMM   &\textbf{-36486.69} &\textbf{74329.46}  &155\\
&AsLG-HMM        &-42404.01 &86024.12  &139\\
&LMSAR 		     &-40498.35 &82746.48  &200\\
&AR-MoG-HMM   	 &-38730.60 &80400.84  &336\\
&MoG-HMM 		 &-41528.35 &85261.43  &252\\
&Naïve-HMM   	 &-42205.88 &85391.65  &\textbf{112}\\
&BMM   		     &-44761.39 &91727.52  &252\\
\end{tabular}
	\caption{Model scores for ball-bearing data}
	\label{table:Bearing}
\end{table}

Table~\ref{table:Bearing} shows the results obtained by the models for each ball-bearing. We observe that for all the ball-bearings, the best results were obtained by AR-AsLG-HMM and LMSAR. Whereas, the worst results were obtained by BMM. However in B3, AR-AsLG-HMM, AsLG-HMM and naïve-HMM obtained results far from the ones obtained by AR-MoG-HMM and LMSAR. Nevertheless, as it will be observed in the viterbi paths, the predictions obtained by these models are poor and again the hypothesis of constant variance among hidden states may be the cause of the good fit but the poor prediction capabilities. Nevertheless, in terms of number of parameters, we see that naïve-HMM, AsLG-HMM and AR-AsLG-HMM used the least amount of parameters for all the ball-bearings. The remaining models used two or three times the amount of parameters used by naïve-HMM. 

\begin{figure}[h]
	\centering
	\begin{tabular}{ccc}
		\includegraphics[scale=0.26]{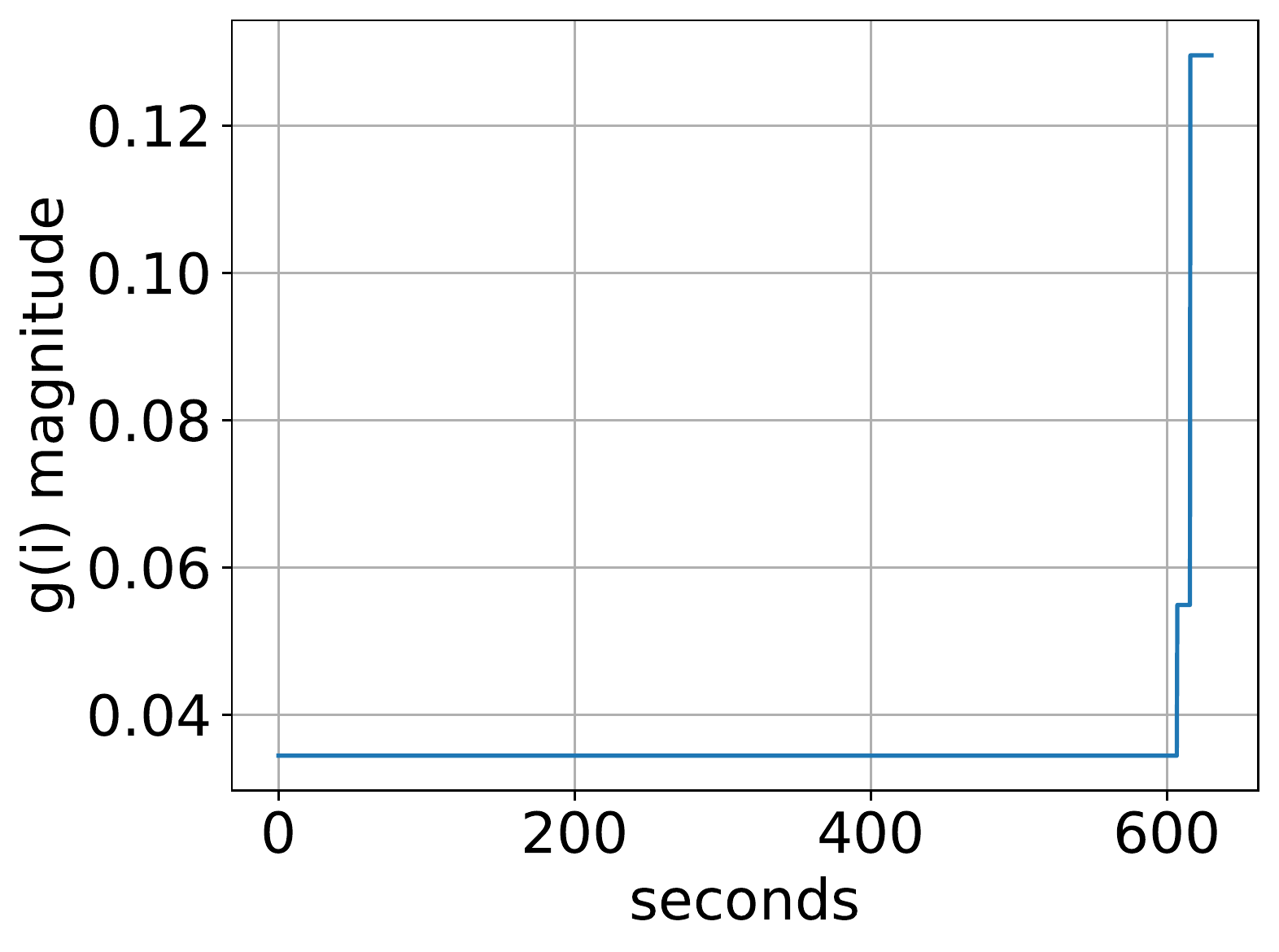} &
		\includegraphics[scale=0.26]{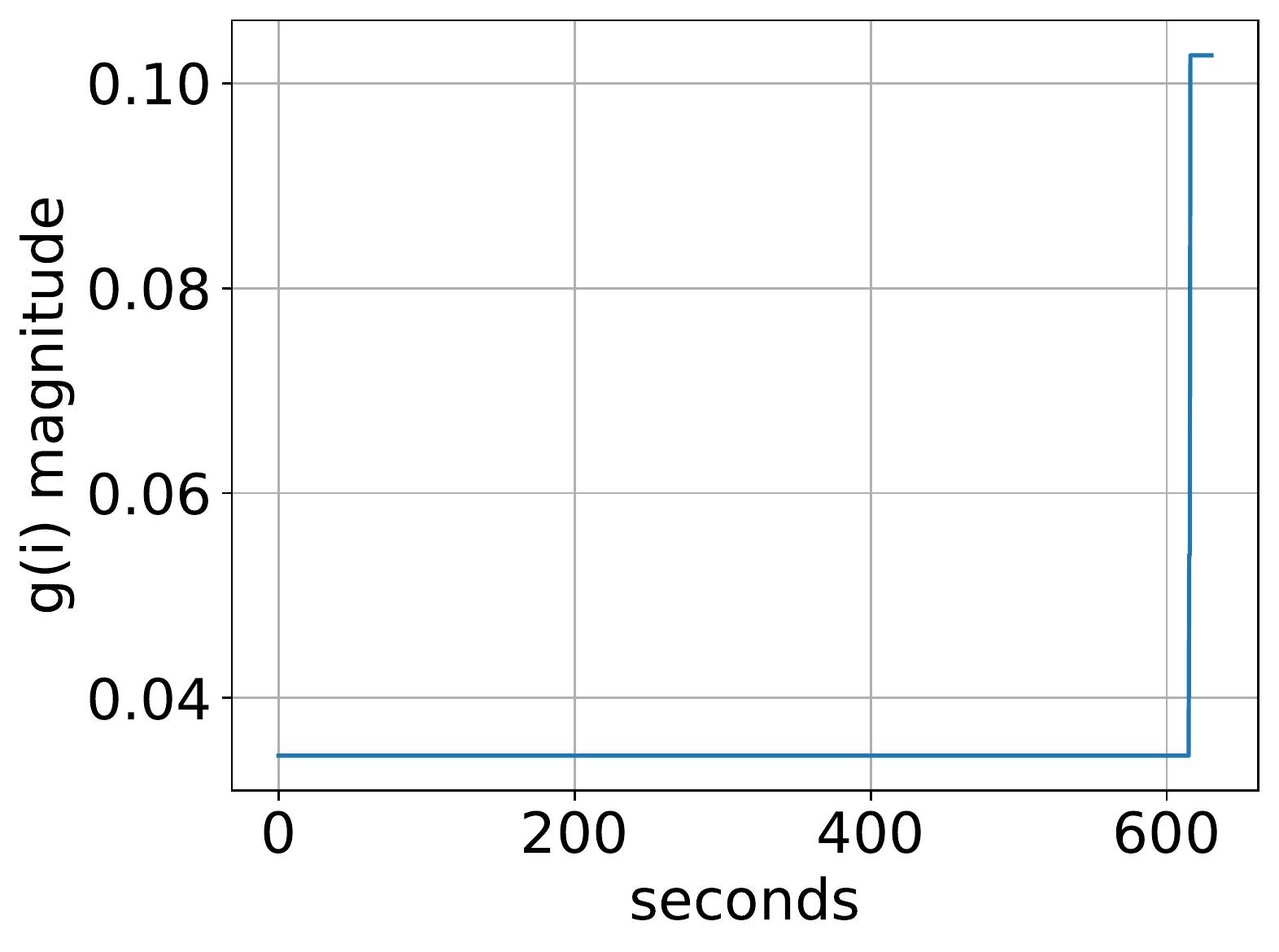}&
		\includegraphics[scale=0.26]{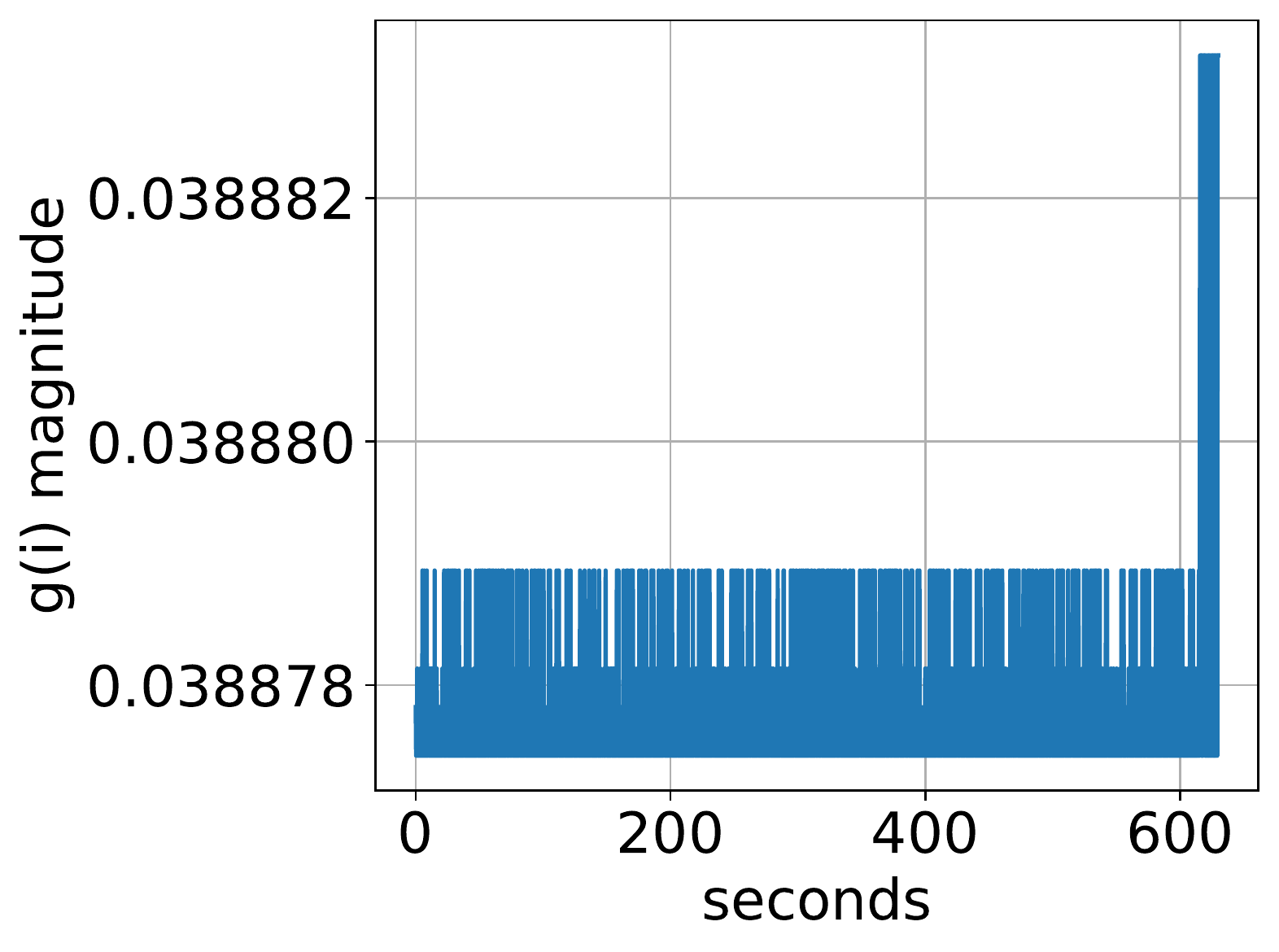} \\
		(a) AR-AsLG-HMM& (b) AsLG-HMM & (c) LMSAR \\
		\includegraphics[scale=0.26]{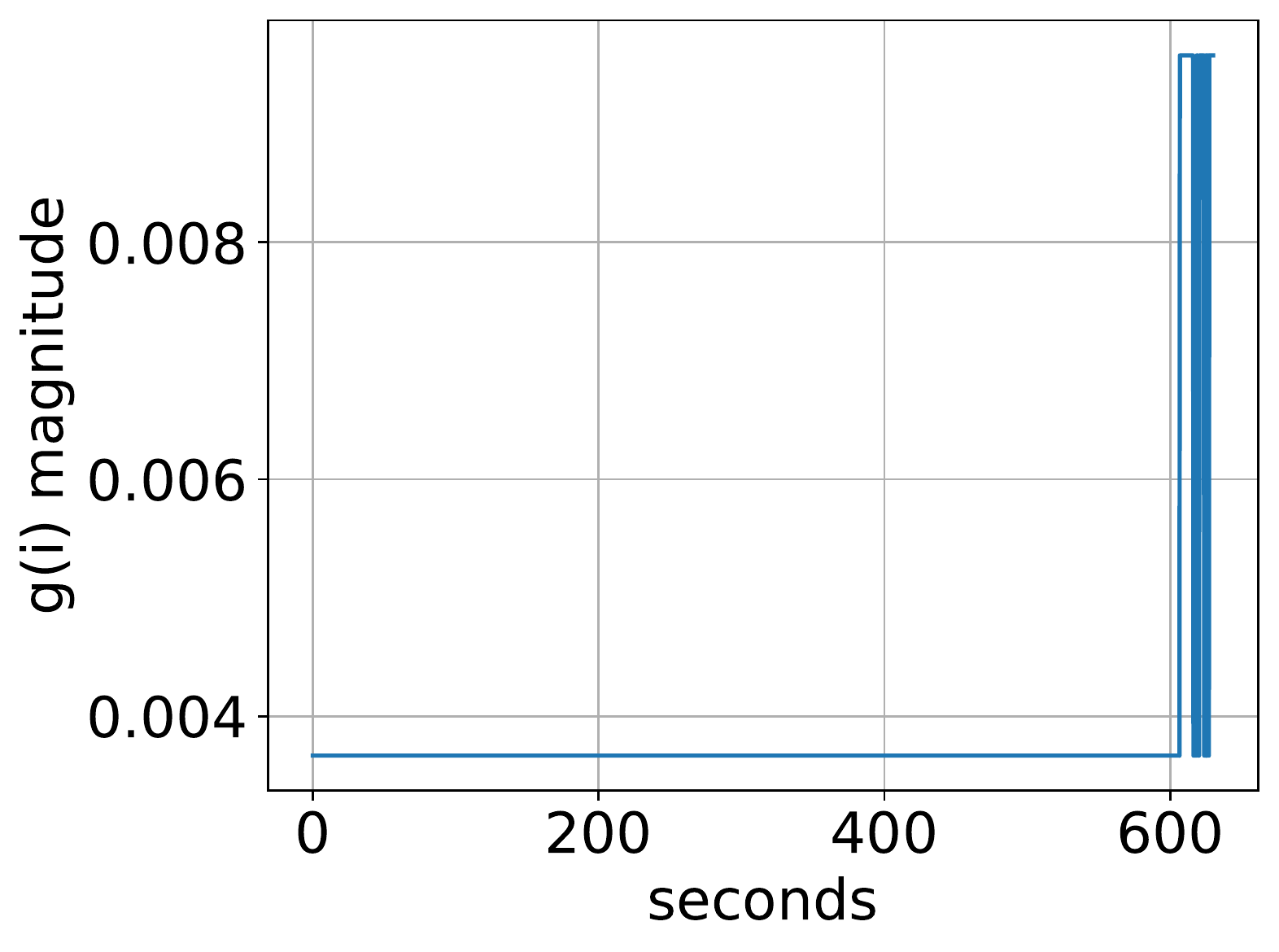} &
		\includegraphics[scale=0.26]{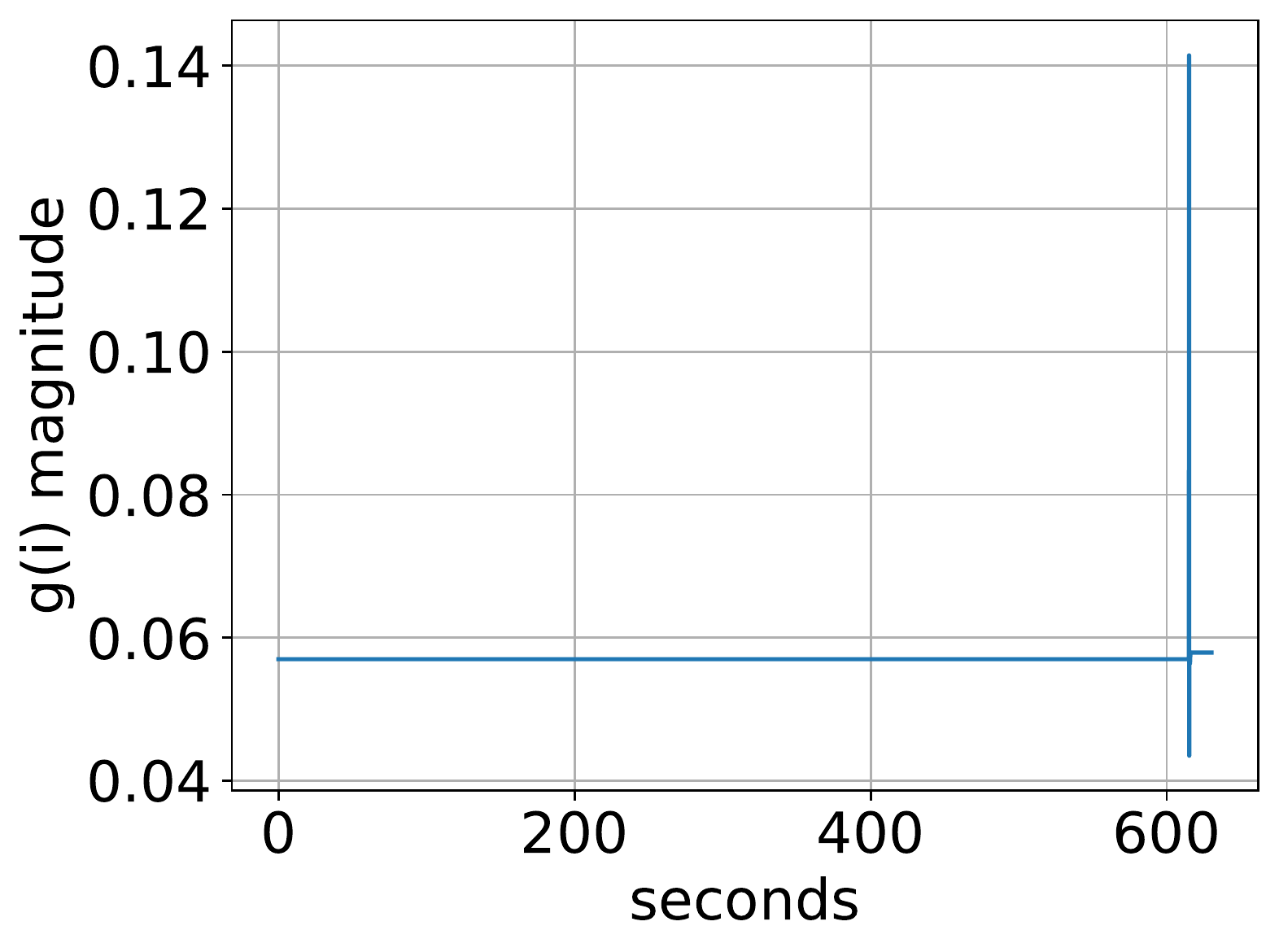}&
		\includegraphics[scale=0.26]{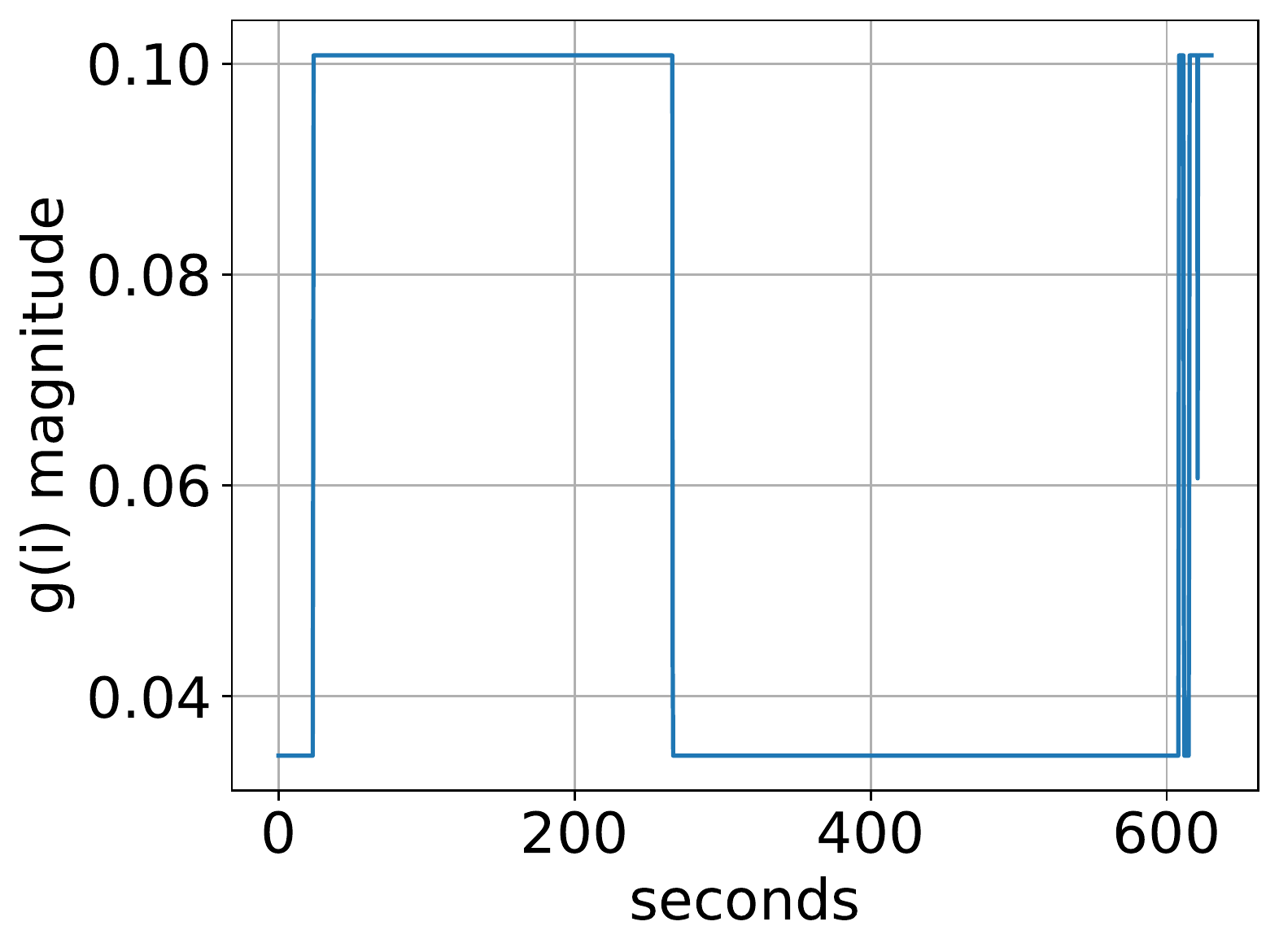}\\
		(d) AR-MoG-HMM&(e) MoG-HMM & (f) Naïve-HMM\\
		\includegraphics[scale=0.26]{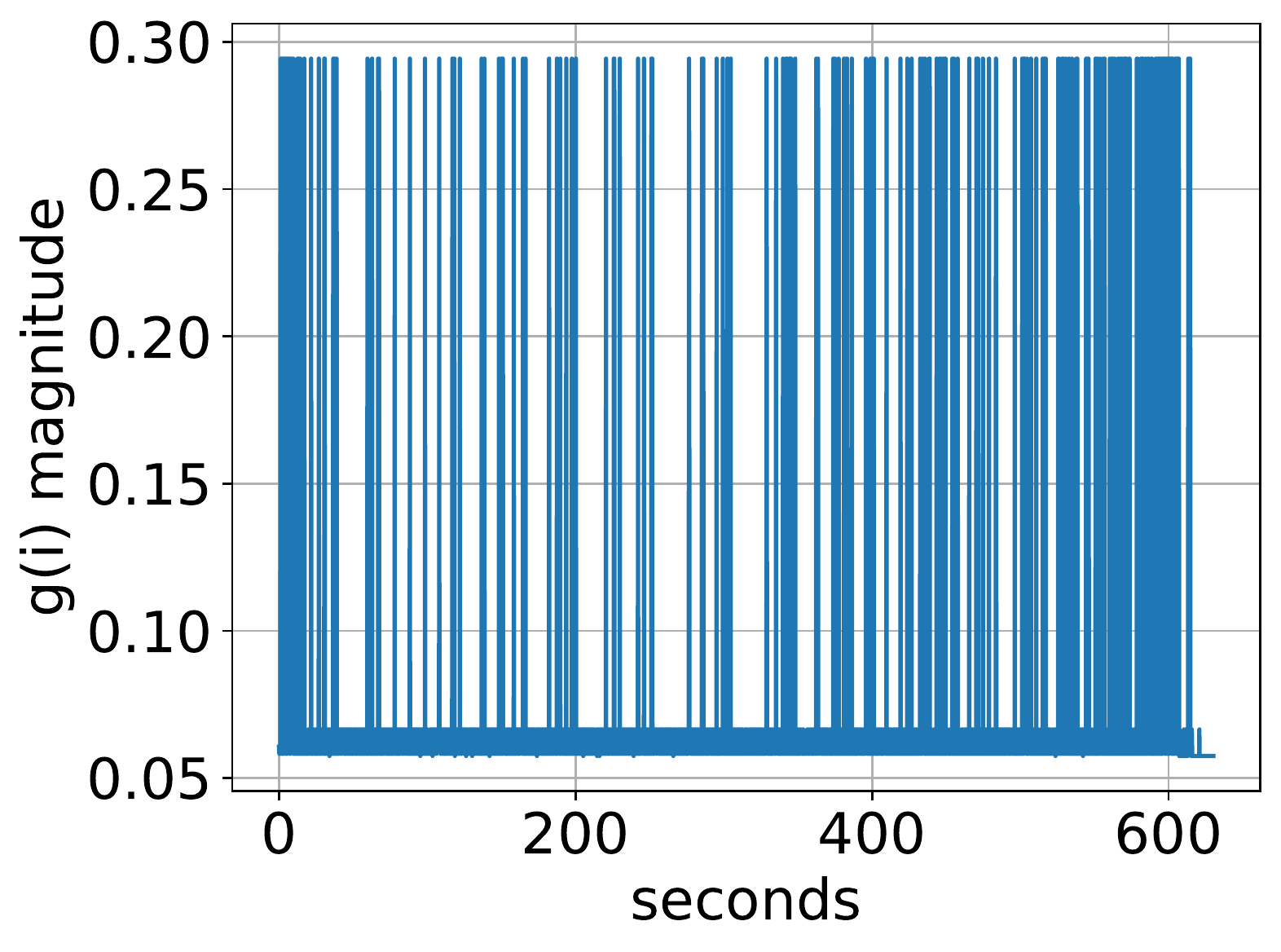}  &&\\
		(g) BMM & & \\
	\end{tabular}
	\caption{Sequence of hidden states by each model for B3}
	\label{fig:Bearing3}	
\end{figure}

Fig.~\ref{fig:Bearing3} shows the paths for the testing B3. All the models exhibit low $g_1(i)$ values  during all the life of the ball-bearing and a jump to higher values of $g_1(i)$ at the end of the life of the ball-bearing as expected. High values of $g(i)$ represent high amplitudes in the fundamental frequencies of the ball-bearing and therefore shows evidence of degradation and a near failure. As we can see, the proposed model is capable of determining this behaviour from incoming ball-bearing data. On the other hand, LMSAR and AR-MoG-HMM in spite of the high BIC score obtained, we observe that obtained predictions may be noisy and not so clear as the predicted Viterbi path obtained by AR-AsLG-HMM. 

\begin{figure}[H]
	\centering
	\begin{tabular}{cc}
		\includegraphics[scale=0.6]{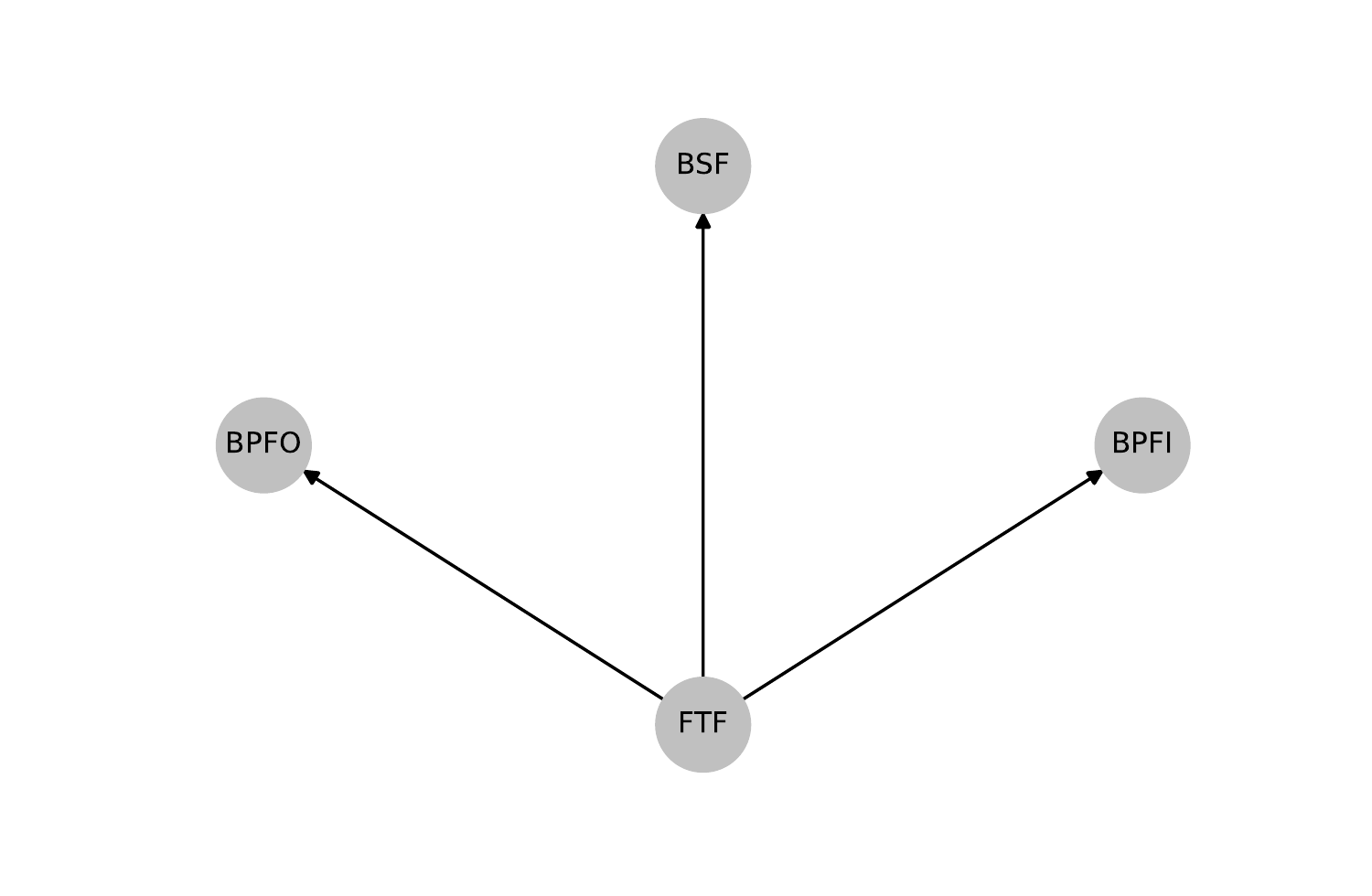} &
		\includegraphics[scale=0.6]{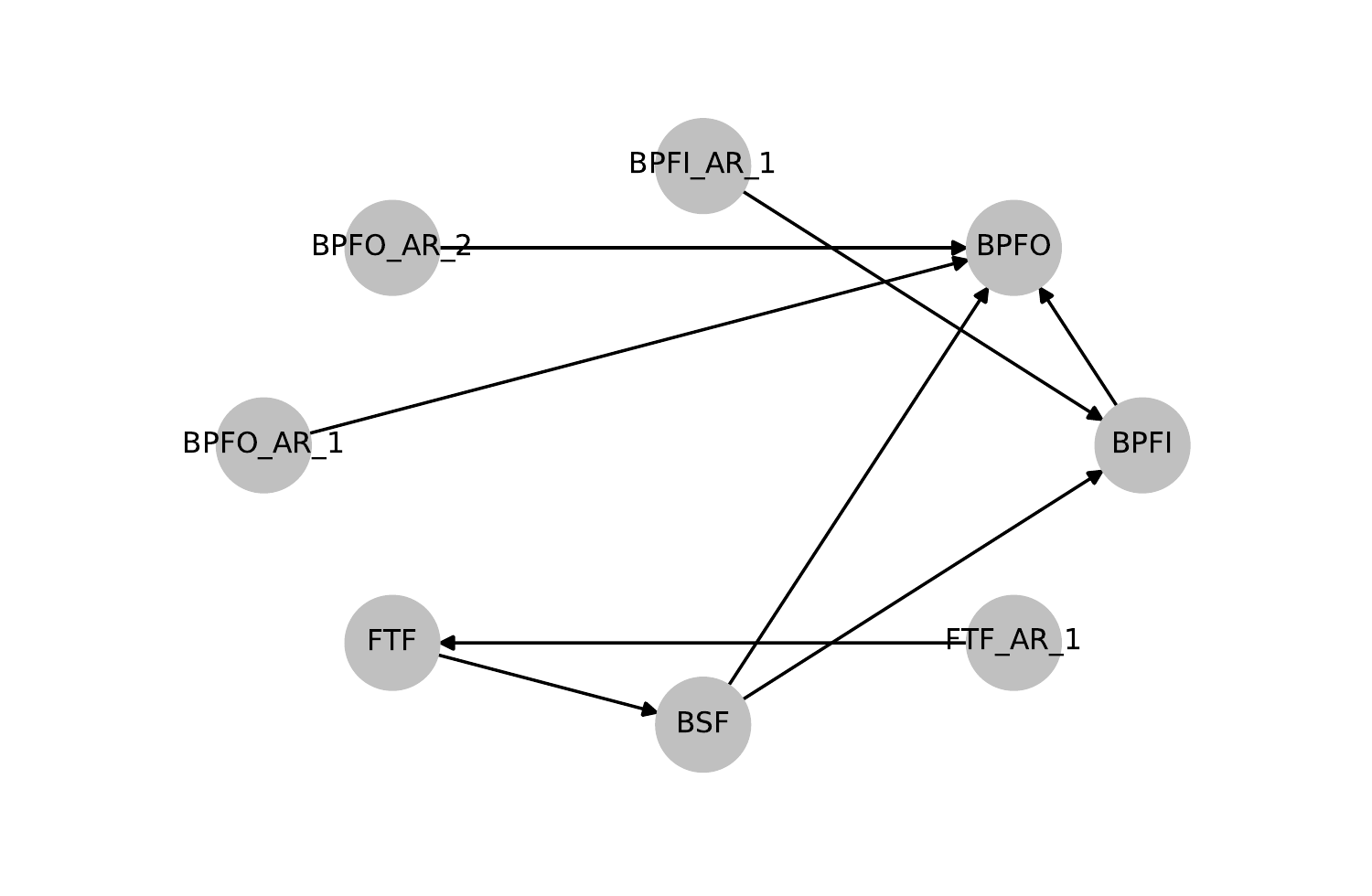}\\
		(a) $g(i) = 0.03$& (b) $g(i) = 0.13$ \\
	\end{tabular}
	\caption{Context-specific graphs learned by AR-AsLG-HMM. (a) shows a graph where the bearings health is good and (b), where the  bearings health is bad.}
	\label{fig:grap_bear}	
\end{figure}

A relevant part of this model is the generation of context-specific Bayesian networks. In Fig.~\ref{fig:grap_bear} we observe two context-specific Bayesian networks. (a) represents a good health state. In this graph we observe that the cage frequencies (FTF) determines the remaining variables. This implies that knowing the behaviour of the cage, determines the behaviour of the ball-bearing rollers and races. (b) represents a bad health state and shows a more complex structure.  In this context-specific Bayesian network AR values are relevant and are taken into consideration. We again see the dominance of the ball-bearings cage (FTF) to determine the dynamical process of the model, but some frequencies are not directly dependant to this variable e.g., the outer race frequencies (BPFO) depends on the inner race frequencies (BPFI) and the rollers frequencies (BSF) and these depend directly on the cage frequency (FTF). In summary, these graphs are capable of explaining the ball-bearings dynamical process depending on its health.

\section{Conclusions}
In this paper, we extended the development of asymmetric hidden Markov models allowing us to determine and learn the optimal number of time lags depending on the value of the hidden state via the SEM algorithm. Also we introduce a greedy-forward heuristic used to find the best structure for the model. We also theoretically adapted the forward-backward, Viterbi and EM algorithms to our proposed log-likelihood function. Additionally, we showed that every iteration of the EM algorithm improves the log-likelihood of the model. We introduced a numerical labelling function, which can be helpful in determining the nature of the learned hidden Markov models and to identify changes in the magnitude of the hidden variable.

We used synthetic and real data to validate the proposed model. We compared ourselves with many other models. In general, the AR-AsLG-HMM obtained fair results  for synthetic data and for real data. We also showed the use of the learned context-specific Bayesian networks to extract information about the nature of the problem being modelled. Additionally, the number of parameters learned by  AR-AsLG-HMM were usually in an intermediate point between the simplest model (naïve-HMM) and the mixture models which is helpful to prevent data overfit and interpret the nature of the model.  

In future work, we would like to combine the idea of asymmetric autoregressive models with other types of HMM such as HSMMs or HHMMs. Finally, we want to apply the proposed model to online environments and observe its behaviour to detect and treat concept drifts.

\section*{Acknowledgment}
This study was supported by the Spanish Centre for the Development of Industrial Technology (CDTI) through the IDI-20180156 LearnIIoT and supported by the Spanish Ministry of Economy and Competitiveness through the TIN2016-79684-P project. We would like to thank Etxe-Tar for advice regarding the machine hardware and Ikergune for support related to filtering the datasets to perform the corresponding experiments in the case of ball-bearing degradation case.

\section*{Appendix}
\subsection*{Parameters used in the synthetic data}

\begin{table}[H]
	\centering
	\begin{tabular}{ccc}
		State & Parameter & Value \\
		\hline
		1  &$f^t_{11}$  & $1$ \\
		&$f^t_{12}$  & $2$ \\
		&$f^t_{13}$  & $3$ \\
		&$\sigma_{11}$ & $1$ \\
		&$\sigma_{12}$ & $1$ \\
		&$\sigma_{13}$ & $1$ \\
		\hline
		2  &$f^t_{21}$  & $2$ \\
		&$f^t_{22}$  & $1+2x_3^t$ \\
		&$f^t_{23}$  &  $4$   \\
		&$\sigma_{21}$ & $3$  \\
		&$\sigma_{22}$ & $5$  \\
		&$\sigma_{23}$ & $4$  \\
		\hline
		3  &$f^t_{31}$  &  $1+0.1x_1^{t-1}$     \\
		&$f^t_{32}$  &  $5x^t_1+4x^t_3+0.7x_2^{t-1}$     \\
		&$f^t_{33}$  &  $4+2x^t_1+0.99x_3^{t-1}$     \\
		&$\sigma_{31}$ &  $2$    \\
		&$\sigma_{32}$ &  $3$  \\
		&$\sigma_{33}$ &  $1$  \\
		\hline
	\end{tabular}
	\caption{Scenario 1 parameters}
	\label{table:parameters}
\end{table}

\begin{table}[H]
	\centering
	\begin{tabular}{ccc}
		State & Parameter & Value \\
		\hline
		1	&$f^t_{11}$  & $1.5$ \\
		&$f^t_{12}$  & $2.5$ \\
		&$f^t_{13}$  & $4.5.$ \\
		&$f^t_{14}$  & $3.5$ \\
		&$f^t_{15}$  & $6.5$ \\
		&$f^t_{16}$  & $1.5$ \\
		&$\sigma_{11}$ & $1.5$ \\
		&$\sigma_{12}$ & $2.0$ \\
		&$\sigma_{13}$ & $3.0$ \\
		&$\sigma_{14}$ & $8.0$ \\
		&$\sigma_{15}$ & $6.0$ \\
		&$\sigma_{16}$ & $0.5$ \\
		\hline
		2 	&$f^t_{21}$  & $1.5+0.2x^{t-1}_1$ \\
		&$f^t_{22}$  & $2.5x_3^t$ \\
		&$f^t_{23}$  & $3.0x_5^t+0.99x_3^{t-1}$ \\
		&$f^t_{24}$  &  $1.5+ 9.5x_1^t$\\
		&$f^t_{25}$  & $6.5+0.99x^{t-1}_5$ \\
		&$f^t_{26}$  & $0.5+ 6.8x_5^t$ \\
		&$\sigma_{21}$ & $3.5$ \\
		&$\sigma_{22}$ & $2$ \\
		&$\sigma_{23}$ & $4$ \\
		&$\sigma_{24}$ & $2$ \\
		&$\sigma_{25}$ & $5.5$ \\
		&$\sigma_{26}$ & $2$ \\
		\hline
		3   &$f^t_{31}$  & $1.5+0.999x_1^{t-1}$ \\
		&$f^t_{32}$  & $2.5+5.0x_1^t+8.0x_4^t+0.888x_2^{t-1}+0.111x_2^{t-2}$ \\
		&$f^t_{33}$  & $4.5+1.5x_1^t+0.999x_3^{t-1}$ \\
		&$f^t_{34}$  & $3.5+1.5x_1^t+2.0x_3^t+0.1x_4^{t-1} $ \\
		&$f^t_{35}$  & $5.0x_3^t$ \\
		&$f^t_{36}$  & $1+3.5x_3^t-4.5x_5^t+0.8x_6^{t-1}$ \\
		&$\sigma_{31}$ & $3$ \\
		&$\sigma_{32}$ & $3.5$ \\
		&$\sigma_{33}$ & $4.0$ \\
		&$\sigma_{34}$ & $6.5$ \\
		&$\sigma_{35}$ & $5.5$ \\	
		&$\sigma_{36}$ & $7.0$ \\
		\hline
	\end{tabular}
	\caption{Scenario 2 parameters}
	\label{table:parameters2}
\end{table}

\subsection*{Proofs}
\textbf{Lemma~{1}}:
	From the maximization step, we have $\mathcal{Q}^{p^*}(\boldsymbol{\lambda}|\boldsymbol{\lambda}^{(s)})\ = \max_{\boldsymbol{\lambda}}\mathcal{Q}^{p^*}(\boldsymbol{\lambda}|\boldsymbol{\lambda}^{(s)})\geq \mathcal{Q}^{p^ *}(\boldsymbol{\lambda}^{(s)}|\boldsymbol{\lambda}^{(s)})$.
\begin{flushright}
$\blacksquare$
\end{flushright}
	
\textbf{Lemma~{2}}:
 Notice that:
	\begin{equation}\label{eq:cross}
	\mathcal{H}^{p^*}(\boldsymbol{\lambda}|\boldsymbol{\lambda}')-\mathcal{H}^{p^*}(\boldsymbol{\lambda}'|\boldsymbol{\lambda}') 	=\sum_{\boldsymbol{R}(\boldsymbol{Q}^{p^*:T})} P(\boldsymbol{q}^{p^*:T}|\boldsymbol{x}^{0:T},\boldsymbol{\lambda}')\ln\frac{ P(\boldsymbol{q}^{p^*:T}|\boldsymbol{x}^{0:T},\boldsymbol{\lambda})}{P(\boldsymbol{q}^{p^*:T}|\boldsymbol{x}^{0:T},\boldsymbol{\lambda}')}.
	\end{equation}
	Observe that if $ P(\boldsymbol{q}^{p^*:T}|\boldsymbol{x}^{0:T},\boldsymbol{\lambda})=  P(\boldsymbol{q}^{p^*:T}|\boldsymbol{x}^{0:T},\boldsymbol{\lambda}')$, then the logarithm in  Eq.~(\ref{eq:cross}) is zero and $\mathcal{H}^{p^*}(\boldsymbol{\lambda}|\boldsymbol{\lambda}')=\mathcal{H}^{p^*}(\boldsymbol{\lambda}'|\boldsymbol{\lambda}')$. However, if we apply Jensen's inequality for concave functions to Eq.~(\ref{eq:cross}), we have that:
	\begin{equation*}
	\begin{aligned}
	\mathcal{H}^{p^*}(\boldsymbol{\lambda}|\boldsymbol{\lambda}')-\mathcal{H}^{p^*}(\boldsymbol{\lambda}'|\boldsymbol{\lambda}')
	\leq& \ln\sum_{\boldsymbol{R}(\boldsymbol{Q}^{p^*:T})} P(\boldsymbol{q}^{p^*:T}|\boldsymbol{x}^{0:T}, \boldsymbol{\lambda}')\frac{ P(\boldsymbol{q}^{p^*:T}|\boldsymbol{x}^{0:T},\boldsymbol{\lambda})}{P(\boldsymbol{q}^{p^*:T}|\boldsymbol{x}^{0:T},\boldsymbol{\lambda}')}\\
	=& \ln\sum_{\boldsymbol{R}(\boldsymbol{Q}^{p^*:T})} P(\boldsymbol{q}^{p^*:T}|\boldsymbol{x}^{0:T},\boldsymbol{\lambda}).\\
	\end{aligned}
	\end{equation*}
	However $\ln\sum_{\boldsymbol{R}(\boldsymbol{Q}^{p^*:T})} P(\boldsymbol{q}^{p^*:T}|\boldsymbol{x}^{0:T},\boldsymbol{\lambda})\leq 0$, and  therefore $\mathcal{H}^{p^*}(\boldsymbol{\lambda}|\boldsymbol{\lambda}')\leq\mathcal{H}^{p^*}(\boldsymbol{\lambda}'|\boldsymbol{\lambda}')$ as desired.\begin{flushright}
	$\blacksquare$	
\end{flushright}

\textbf{Theorem 1}:
To prove (a), we have the following identity: $LL(\boldsymbol{\lambda}^{(s+1)})) = \mathcal{Q}^{p^*}(\boldsymbol{\lambda}^{(s+1)}|\boldsymbol{\lambda}^{(s)})-\mathcal{H}^{p^*}(\boldsymbol{\lambda}^{(s+1)}|\boldsymbol{\lambda}^{(s)})$. Note that:
	\begin{equation*}\label{eq:diffs}
	\begin{split}
	LL(\boldsymbol{\lambda}^{(s+1)})-LL(\boldsymbol{\lambda}^{(s)})  &=\mathcal{Q}^{p^*}(\boldsymbol{\lambda}^{(s+1)}|\boldsymbol{\lambda}^{(s)})-\mathcal{Q}^{p^*}(\boldsymbol{\lambda}^{(s)}|\boldsymbol{\lambda}^{(s)})\\
	&+\mathcal{H}^{p^*}(\boldsymbol{\lambda}^{(s)}|\boldsymbol{\lambda}^{(s)}) -\mathcal{H}^{p^*}(\boldsymbol{\lambda}^{(s+1)}|\boldsymbol{\lambda}^{(s)}).
	\end{split}
	\end{equation*}
	From Lemma~1, we have that $\mathcal{Q}^{p^*}(\boldsymbol{\lambda}^{(s+1)}|\boldsymbol{\lambda}^{(s)})-\mathcal{Q}^{p^*}(\boldsymbol{\lambda}^{(s)}|\boldsymbol{\lambda}^{(s)})\geq 0$ and from Lemma~2, we have that $\mathcal{H}^{p^*}(\boldsymbol{\lambda}^{(s)}|\boldsymbol{\lambda}^{(s)}) -\mathcal{H}^{p^*}(\boldsymbol{\lambda}^{(s+1)}|\boldsymbol{\lambda}^{(s)})\geq 0$. Therefore $LL(\boldsymbol{\lambda}^{(s+1)})-LL(\boldsymbol{\lambda}^{(s)})\geq 0$ and the desired results are obtained.
	
	To prove (b), we know from (a) that the sequence $\{{LL}(\boldsymbol{\lambda}^{(s)})\}_{s\in{Z}^+}$ does not decrease and is also upper bounded by zero. Therefore, $\{LL(\boldsymbol{\lambda}^{(s)})\}_{s}$ converges to a certain real finite number $LL^*$  with $LL^*\leq 0$. 
\begin{flushright}
	$\blacksquare$	
\end{flushright}

\textbf{Lemma 3}:

	For the forward variable, we have that for $t=p^*+1,...,T$ and  $i=1,...,N$:
	\begin{equation}\label{eq:forar}
	\begin{aligned}
	\alpha^t_{p^*}&(i)  \\
	=& \sum_{j=1}^N P(Q^t=i,Q^{t-1}=j,\boldsymbol{x}^{{p^*}:t}|\boldsymbol{x}^{0:{p^*}-1},\boldsymbol{\lambda}) \\
	=& \sum_{j=1}^N P(\boldsymbol{x}^t|Q^t= i,\boldsymbol{x}^{{t-p^*}:t-1},\boldsymbol{\lambda})\\
	&\qquad \quad \times P(Q^t=i,Q^{t-1}=j,\boldsymbol{x}^{{p^*}:t-1}|\boldsymbol{x}^{0:{p^*}-1},\boldsymbol{\lambda}) \\
	=&  \sum_{j=1}^N b_i^{p^*}(\boldsymbol{x}^t)P(Q^t=i|Q^{t-1}=j,\boldsymbol{\lambda})\\
	&\qquad \quad \times P(Q^{t-1}=j,\boldsymbol{x}^{{p^*}:t-1}|\boldsymbol{x}^{0:{p^*}-1}, \boldsymbol{\lambda}) \\
	=&  \sum_{j=1}^N b_i^{p^*}(\boldsymbol{x}^t)a_{ji}\alpha_{p^*}^{t-1}(j)  \\
	\end{aligned}
	\end{equation}
	In the second equality of Eq.~(\ref{eq:forar}), we used the fact that $\boldsymbol{X}^t$ is D-separated from $Q^{t-1}$ given $Q^t$ and $\boldsymbol{X}^{t-p^*:t-1}$. In the third equality, we used the fact that $Q^t$ is D-separated from $\boldsymbol{X}^{0:t-1}$ given $Q^{t-1}$. D-separation implies conditional independence in Bayesian networks. As we can see, the forward variable can be computed iteratively as in the traditional HMM. Additionally, the forward variable is initialized with $\alpha^{p^*}_{p^*}(i) = \pi_ib_i^{p^*}(\boldsymbol{x}^{p^*})$, $i=1,...,N$. 
	
	In the case of the backward variable, we have that for  $t=T-1,...,p^* $ and  $i=1,...,N$ :
	\begin{equation}\label{eq:backar}
	\begin{aligned}
	\beta^t_{p^*}&(i) \\
	=& \sum_{j=1}^{N} P(\boldsymbol{x}^{t+1:T},Q^{t+1}=j|Q^t=i,\boldsymbol{x}^{0:t},\boldsymbol{\lambda}) \\
	=&  \sum_{j=1}^{N} P(\boldsymbol{x}^{t+2:T}|Q^{t+1}=j,\boldsymbol{x}^{0:t+1},\boldsymbol{\lambda})\\
	& \qquad \quad \times P(\boldsymbol{x}^{t+1},Q^{t+1}=j|Q^t=i,\boldsymbol{x}^{0:t},\boldsymbol{\lambda}) \\
	=&  \sum_{j=1}^{N} \beta_{p^*}^{t+1}(j)P(\boldsymbol{x}^{t+1}|Q^{t+1}=j,\boldsymbol{x}^{t+1-p^*:t}, \boldsymbol{\lambda})\\
	& \qquad \quad \times P(Q^{t+1}=j|Q^t=i,\boldsymbol{\lambda}) \\
	=&  \sum_{j=1}^{N} \beta_{p^*}^{t+1}(j)b_j^{p^*}(\boldsymbol{x}^{t+1})a_{ij}
	\end{aligned}
	\end{equation}
	
	In the second equality of Eq.~(\ref{eq:backar}), we again applied D-separation, specifically $\boldsymbol{X}^{t+2:T}$ is D-separated from $Q^t$ given $Q^{t+1}$ and $\boldsymbol{X}^{0:t+1}$. In the third equality, we used that $\boldsymbol{X}^{t+1}$ is D-separated from $Q^t$ given $Q^{t+1}$ and $\boldsymbol{X}^{0:t}$. Additionally, $Q^{t+1}$ is D-separated from $\boldsymbol{X}^{0:t}$ given $Q^t$; also,  $\boldsymbol{X}^{t+1}$ is D-separated of $\boldsymbol{X}^{0:t-p^*}$ given $\boldsymbol{X}^{t+1-p^*:t}$ because each $\boldsymbol{X}^t$ is dependent on maximum $p^*$ lags. As we can see, the backward variable can be computed iteratively as in the traditional HMM. Finally, the backward variable is initialized with $\beta^T_{p^*}(i)= 1$, $i=1,...,N$.
\begin{flushright}
	$\blacksquare$	
\end{flushright}

\textbf{Theorem 2}:
	We use Lagrange multipliers with the restrictions $\sum_{i=1}^{N} \pi_i = 1$ and $\sum_{j=1}^Na_{ij}=1$ for $i=1,...,N$. The corresponding Lagrangian function is:
	\begin{equation}\label{eq:lagrange}
	\begin{split}
	\mathcal{L}(\boldsymbol{\lambda},&\tau_0,\tau_1,...,\tau_N)\\ = &\mathcal{Q}^{p^{*}}(\boldsymbol{\lambda}|\boldsymbol{\lambda}')+ \tau_0(1-\sum_{i=1}^{N} \pi_i) + \sum_{i=1}^N\tau_i(1-\sum_{j=1}^Na_{ij})
	\end{split}
	\end{equation}
	
	If we compute the derivative of $\mathcal{L}$ in Eq.~(\ref{eq:lagrange}) with respect to $\pi_i$ and equalize to zero, we obtain from $\mathcal{Q}^{p^*}(\boldsymbol{\lambda}|\boldsymbol{\lambda}')$: 
	$$
	\frac{\partial \mathcal{L}}{\partial \pi_i} = \frac{\gamma^{p^*}(i)}{\pi_i} - \tau_0 =0.$$ 
	Then, $\pi_i = \frac{\gamma^{p^*}(i)}{\tau_0}$, and $\sum_{i=1}^{N}\pi_i = \frac{\sum_{i=1}^N\gamma^{p^*}(i)}{\tau_0}$. Hence,  $\tau_0 = 1$. Therefore, the updating formula for $\pi_i$, for $i=1,...,N$, is:
	\begin{equation}\label{eq:uppinew}
	\pi_i^* = \gamma^{p^*}(i).
	\end{equation}  
	Similarly, with respect to $a_{ij}$: $$\frac{\partial \mathcal{L}}{\partial a_{ij}} = \sum_{t=p^*}^{T-1}\frac{\xi^t(i,j)}{a_{ij}} - \tau_i =0.$$ 
	Then, $a_{ij} = \frac{\sum_{t=p^*}^{T-1}\xi^t(i,j)}{\tau_i}$, and $\sum_{j=1}^{N}a_{ij}  = \frac{\sum_{t=p^*}^{T-1}\sum_{j=1}^N\xi^t(i,j)}{\tau_i}$. Additionally,
	$\tau_i = \sum_{t=p^*}^{T-1}\gamma^t(i)$. Therefore, the updating formula for $a_{ij}$ for $i,j=1,...,N$ is:
	\begin{equation}\label{eq:updateaij}
	a_{ij}^* = \frac{\sum_{t=p^*}^{T-1}\xi^t(i,j)}{\sum_{t=p^*}^{T-1}\gamma^t(i)}.
	\end{equation}  
	Now, we compute the derivative of $\mathcal{L}$ in Eq.~(\ref{eq:lagrange}) with respect to  parameters $\eta_{imp}$, $\beta_{imk}$ and  $\sigma_{im}^2$ of the Gaussian. For the derivative with respect to $\beta_{im0}$, we obtain:
	\begin{equation}\label{eq:Hder1}
	\frac{\partial \mathcal{L} }{\partial \beta_{im0}}  = \sum_{t=p^*}^T\gamma^t(i)\frac{\partial}{\partial \beta_{im0}}\ln(\mathcal{N}(x^t_m|f^t_{im}, \sigma_{im}^2)).
	\end{equation}
	Thus,
	\begin{equation*}\label{eq:Hder2}
	0=\sum_{t=p^*}^{T}\frac{\gamma^{t}(i)}{\sigma_{im}^2}(f^t_{im}- x^t_m).\\
	\end{equation*}
	Then,
	\begin{equation}\label{eq:Hder}
	\sum_{t=p^*}^{T}\gamma^{t}(i)x^t_m = \sum_{t=p^*}^{T}\gamma^{t}(i)f^t_{im}.
	\end{equation}
	Now, if $\mathcal{L}$ in Eq.~(\ref{eq:lagrange}) is derived with respect to $\beta_{imk}$, with $k=1,...,k_{im}$, and with respect to $\eta_{imr}$ with $r=1,...,p_{im}$ as in Eq.~(\ref{eq:Hder1}), we obtain the following equations:
	\begin{equation}\label{eq:siseq}
	\begin{aligned}
	\sum_{t=p^*}^{T}&\gamma^{t}(i)x^t_mu^t_{im1} = \sum_{t=p^*}^{T}\gamma^{t}(i)u^t_{im1}f^t_{im}\\
	\sum_{t=p^*}^{T}&\gamma^{t}(i)x^t_mu^t_{im2} = \sum_{t=p^*}^{T}\gamma^{t}(i)u^t_{im2}f^t_{im} \\
	&\vdots  \qquad\qquad\qquad   \vdots \qquad\qquad\qquad   \vdots \\
	\sum_{t=p^*}^{T}&\gamma^{t}(i)x^t_mu^t_{imk_{im}} = \sum_{t=p^*}^{T}\gamma^{t}(i)u^t_{imk_{im}}f^t_{im} \\
	\sum_{t=p^*}^{T}&\gamma^{t}(i)x^t_mx^{t-1}_{m} = \sum_{t=p^*}^{T}\gamma^{t}(i)x^{t-1}_{m}f^t_{im} \\
	&\vdots  \qquad\qquad\qquad   \vdots \qquad\qquad\qquad   \vdots  \\
	\sum_{t=p^*}^{T}&\gamma^{t}(i)x^t_mx^{t-p_{im}}_{m} = \sum_{t=p^*}^{T}\gamma^{t}(i)x^{t-p_{im}}_{m}f^t_{im} \\
	\end{aligned}
	\end{equation}
	
	The solution to this linear system of equations gives $\{\beta^*_{im0},..., \beta^*_{imk_{im}}\}$ and  $\{\eta^*_{im1},...,\eta^*_{imp_{im}}\}$, for each variable $X_m$, $m=1,2,...,M$ and hidden state $i\in R(Q)$. Once these parameters are known, the mean $\hat{f}_{im}^t:=\beta^*_{im0}+ \beta^*_{im1} u_{im1}^{t}+\cdots+\beta^*_{imk_{im}}u_{imk_{im}}^{t}+\eta^*_{im1}x^{t-1}_m+\cdots+\eta^*_{imp_{im}}x^{t-p_{im}}_m$ can be computed. To update $\sigma_{im}^2$, we compute the derivative of $\mathcal{L}$ in Eq.~(\ref{eq:lagrange}) with respect to $\sigma_{im}^2$ and equalize to zero:
	
	\begin{equation*}\label{eq:upsigma1}
	\begin{split}
	\frac{\partial \mathcal{L}}{\partial \sigma_{im}^2} &= \sum_{t=p^*}^T\gamma^t(i)\frac{\partial}{\partial \sigma_{im}^2}\ln(\mathcal{N}\big(x^t_m| \hat{f}_{im}^t,\sigma_{im}^2\big))
	\end{split}
	\end{equation*}
	Thus
	\begin{equation*}\label{eq:upsigma15}
	\begin{split} 0=\sum_{t=p^*}^T\gamma^t(i)\big(\frac{(x_m^t-\hat{f}_{im}^t)^2}{\sigma_{im}^4} -\frac{1}{\sigma_{im}^2}\big) \\
	\end{split}
	\end{equation*}
	Hence,
	\begin{equation}\label{eq:upsigma2}
	\begin{split}
	(\sigma_{im}^2)^*  = \frac{\sum_{t=p^*}^T\gamma^t(i)(x_m^t-\hat{f}_{im}^t)^2}{\sum_{t=p^*}^T\gamma^t(i)}.
	\end{split}
	\end{equation}
\begin{flushright}
	$\blacksquare$	
\end{flushright}

\textbf{Lemma 4}:

	Observe that:
	\begin{equation}\label{eq:vitar}
	\begin{aligned}
	\delta^t_{p^*}(i) &= \max_{\boldsymbol{q}^{p^*:t-1}}\{P(\boldsymbol{x}^t|\boldsymbol{x}^{t-p^*:t-1},Q^t=i,\boldsymbol{\lambda})\\
	& \qquad \times P(\boldsymbol{x}^{p^*:t-1},\boldsymbol{q}^{p^*:t-1},Q^t=i|\boldsymbol{x}^{0:p^*-1},\boldsymbol{\lambda})  \}\\
	&= \max_{\boldsymbol{q}^{p^*:t-1}}\{b^{p^*}_i(\boldsymbol{x^t})P(Q^t=i|q^{t-1},\boldsymbol{\lambda})\\
	& \qquad \times P(\boldsymbol{x}^{p^*:t-1},\boldsymbol{q}^{p^*:t-1}|\boldsymbol{x}^{0:p^*-1},\boldsymbol{\lambda})\} \\
	&= \max_{\boldsymbol{q}^{p^*:t-1}}\{b^{p^*}_i(\boldsymbol{x^t})a_{q^{t-1}i}\delta_{p^*}^{t-1}(q^{t-1})\} \\
	&= \max_{j=1,...,N}\{\delta_{p^*}^{t-1}(j)a_{ji}\}b^{p^*}_i(\boldsymbol{x^t}) \\
	\end{aligned}
	\end{equation}
	The first equality of Eq.~(\ref{eq:vitar}), we noticed that $\boldsymbol{X}^t$ is D-separated of $Q^{p^*:t-1}$ given $Q^t$ and $\boldsymbol{X}^{t-p^*:t-1}$. In the second equality, we also take advantage  of $Q^t$ being D-separated from $\boldsymbol{X}^{0:t-1}$ and $\boldsymbol{Q}^{p^*:t-2}$ given $Q^{t-1}$. In the third equality, we have used the dynamic
	programming principle \cite{Fo73, Om69} in $\delta^{t-1}_{p^*}$.  As we can see, $\delta_{p^*}^t(i)$ can be computed iteratively as in its traditional version. The Viterbi algorithm is initialized with $\delta^{p^*}_{p^*}(i) = \pi_ib^{p^*}_i(\boldsymbol{x}^{p^*})$ for $i=1,...,N$.
\begin{flushright}
	$\blacksquare$	
\end{flushright}

\bibliographystyle{ieeetr}
\bibliography{hmmar}

%
\vskip 0pt plus -1fil
\end{document}